\pdfoutput=1
\documentclass{article} %
\usepackage{nips14submit_e,times}
\usepackage{url}
\usepackage{graphicx}
\usepackage[font={small}]{caption}
\usepackage{subcaption}
\usepackage{amsmath,amssymb}
\usepackage{floatrow}
\usepackage{array}
\usepackage{booktabs}
\usepackage{makecell}
\usepackage{setspace}
\usepackage{bbm}				%
\usepackage{morefloats} 				%
\usepackage{natbib}
\usepackage{fancyvrb}   %
\usepackage{geometry}

\newcommand{\figref}[1]{\ref{Figure #1}}

    \definecolor{orange}{cmyk}{0,0.4,0.8,0.2}
    \definecolor{darkorange}{rgb}{.71,0.21,0.01}
    \definecolor{darkgreen}{rgb}{.12,.54,.11}
    \definecolor{myteal}{rgb}{.26, .44, .56}
    \definecolor{gray}{gray}{0.45}
    \definecolor{lightgray}{gray}{.95}
    \definecolor{mediumgray}{gray}{.8}
    \definecolor{inputbackground}{rgb}{.95, .95, .85}
    \definecolor{outputbackground}{rgb}{.95, .95, .95}
    \definecolor{traceback}{rgb}{1, .95, .95}
    \definecolor{red}{rgb}{.6,0,0}
    \definecolor{green}{rgb}{0,.65,0}
    \definecolor{brown}{rgb}{0.6,0.6,0}
    \definecolor{blue}{rgb}{0,.145,.698}
    \definecolor{purple}{rgb}{.698,.145,.698}
    \definecolor{cyan}{rgb}{0,.698,.698}
    \definecolor{lightgray}{gray}{0.5}

    \definecolor{darkgray}{gray}{0.25}
    \definecolor{lightred}{rgb}{1.0,0.39,0.28}
    \definecolor{lightgreen}{rgb}{0.48,0.99,0.0}
    \definecolor{lightblue}{rgb}{0.53,0.81,0.92}
    \definecolor{lightpurple}{rgb}{0.87,0.63,0.87}
    \definecolor{lightcyan}{rgb}{0.5,1.0,0.83}

    \DefineVerbatimEnvironment{Highlighting}{Verbatim}{commandchars=\\\{\}}
    \newenvironment{Shaded}{}{}

    \newcommand{\StringTok}[1]{\textcolor[rgb]{0.25,0.44,0.63}{{#1}}}

    \newcommand{\NormalTok}[1]{{#1}}

    \title{visual\_turing\_test}

\makeatletter
\def\PY@reset{\let\PY@it=\relax \let\PY@bf=\relax%
    \let\PY@ul=\relax \let\PY@tc=\relax%
    \let\PY@bc=\relax \let\PY@ff=\relax}
\def\PY@tok#1{\csname PY@tok@#1\endcsname}
\def\PY@toks#1+{\ifx\relax#1\empty\else%
    \PY@tok{#1}\expandafter\PY@toks\fi}
\def\PY@do#1{\PY@bc{\PY@tc{\PY@ul{%
    \PY@it{\PY@bf{\PY@ff{#1}}}}}}}
\def\PY#1#2{\PY@reset\PY@toks#1+\relax+\PY@do{#2}}

\expandafter\def\csname PY@tok@gd\endcsname{\def\PY@tc##1{\textcolor[rgb]{0.63,0.00,0.00}{##1}}}
\expandafter\def\csname PY@tok@gu\endcsname{\let\PY@bf=\textbf\def\PY@tc##1{\textcolor[rgb]{0.50,0.00,0.50}{##1}}}
\expandafter\def\csname PY@tok@gt\endcsname{\def\PY@tc##1{\textcolor[rgb]{0.00,0.27,0.87}{##1}}}
\expandafter\def\csname PY@tok@gs\endcsname{\let\PY@bf=\textbf}
\expandafter\def\csname PY@tok@gr\endcsname{\def\PY@tc##1{\textcolor[rgb]{1.00,0.00,0.00}{##1}}}
\expandafter\def\csname PY@tok@cm\endcsname{\let\PY@it=\textit\def\PY@tc##1{\textcolor[rgb]{0.25,0.50,0.50}{##1}}}
\expandafter\def\csname PY@tok@vg\endcsname{\def\PY@tc##1{\textcolor[rgb]{0.10,0.09,0.49}{##1}}}
\expandafter\def\csname PY@tok@vi\endcsname{\def\PY@tc##1{\textcolor[rgb]{0.10,0.09,0.49}{##1}}}
\expandafter\def\csname PY@tok@mh\endcsname{\def\PY@tc##1{\textcolor[rgb]{0.40,0.40,0.40}{##1}}}
\expandafter\def\csname PY@tok@cs\endcsname{\let\PY@it=\textit\def\PY@tc##1{\textcolor[rgb]{0.25,0.50,0.50}{##1}}}
\expandafter\def\csname PY@tok@ge\endcsname{\let\PY@it=\textit}
\expandafter\def\csname PY@tok@vc\endcsname{\def\PY@tc##1{\textcolor[rgb]{0.10,0.09,0.49}{##1}}}
\expandafter\def\csname PY@tok@il\endcsname{\def\PY@tc##1{\textcolor[rgb]{0.40,0.40,0.40}{##1}}}
\expandafter\def\csname PY@tok@go\endcsname{\def\PY@tc##1{\textcolor[rgb]{0.53,0.53,0.53}{##1}}}
\expandafter\def\csname PY@tok@cp\endcsname{\def\PY@tc##1{\textcolor[rgb]{0.74,0.48,0.00}{##1}}}
\expandafter\def\csname PY@tok@gi\endcsname{\def\PY@tc##1{\textcolor[rgb]{0.00,0.63,0.00}{##1}}}
\expandafter\def\csname PY@tok@gh\endcsname{\let\PY@bf=\textbf\def\PY@tc##1{\textcolor[rgb]{0.00,0.00,0.50}{##1}}}
\expandafter\def\csname PY@tok@ni\endcsname{\let\PY@bf=\textbf\def\PY@tc##1{\textcolor[rgb]{0.60,0.60,0.60}{##1}}}
\expandafter\def\csname PY@tok@nl\endcsname{\def\PY@tc##1{\textcolor[rgb]{0.63,0.63,0.00}{##1}}}
\expandafter\def\csname PY@tok@nn\endcsname{\let\PY@bf=\textbf\def\PY@tc##1{\textcolor[rgb]{0.00,0.00,1.00}{##1}}}
\expandafter\def\csname PY@tok@no\endcsname{\def\PY@tc##1{\textcolor[rgb]{0.53,0.00,0.00}{##1}}}
\expandafter\def\csname PY@tok@na\endcsname{\def\PY@tc##1{\textcolor[rgb]{0.49,0.56,0.16}{##1}}}
\expandafter\def\csname PY@tok@nb\endcsname{\def\PY@tc##1{\textcolor[rgb]{0.00,0.50,0.00}{##1}}}
\expandafter\def\csname PY@tok@nc\endcsname{\let\PY@bf=\textbf\def\PY@tc##1{\textcolor[rgb]{0.00,0.00,1.00}{##1}}}
\expandafter\def\csname PY@tok@nd\endcsname{\def\PY@tc##1{\textcolor[rgb]{0.67,0.13,1.00}{##1}}}
\expandafter\def\csname PY@tok@ne\endcsname{\let\PY@bf=\textbf\def\PY@tc##1{\textcolor[rgb]{0.82,0.25,0.23}{##1}}}
\expandafter\def\csname PY@tok@nf\endcsname{\def\PY@tc##1{\textcolor[rgb]{0.00,0.00,1.00}{##1}}}
\expandafter\def\csname PY@tok@si\endcsname{\let\PY@bf=\textbf\def\PY@tc##1{\textcolor[rgb]{0.73,0.40,0.53}{##1}}}
\expandafter\def\csname PY@tok@s2\endcsname{\def\PY@tc##1{\textcolor[rgb]{0.73,0.13,0.13}{##1}}}
\expandafter\def\csname PY@tok@nt\endcsname{\let\PY@bf=\textbf\def\PY@tc##1{\textcolor[rgb]{0.00,0.50,0.00}{##1}}}
\expandafter\def\csname PY@tok@nv\endcsname{\def\PY@tc##1{\textcolor[rgb]{0.10,0.09,0.49}{##1}}}
\expandafter\def\csname PY@tok@s1\endcsname{\def\PY@tc##1{\textcolor[rgb]{0.73,0.13,0.13}{##1}}}
\expandafter\def\csname PY@tok@ch\endcsname{\let\PY@it=\textit\def\PY@tc##1{\textcolor[rgb]{0.25,0.50,0.50}{##1}}}
\expandafter\def\csname PY@tok@m\endcsname{\def\PY@tc##1{\textcolor[rgb]{0.40,0.40,0.40}{##1}}}
\expandafter\def\csname PY@tok@gp\endcsname{\let\PY@bf=\textbf\def\PY@tc##1{\textcolor[rgb]{0.00,0.00,0.50}{##1}}}
\expandafter\def\csname PY@tok@sh\endcsname{\def\PY@tc##1{\textcolor[rgb]{0.73,0.13,0.13}{##1}}}
\expandafter\def\csname PY@tok@ow\endcsname{\let\PY@bf=\textbf\def\PY@tc##1{\textcolor[rgb]{0.67,0.13,1.00}{##1}}}
\expandafter\def\csname PY@tok@sx\endcsname{\def\PY@tc##1{\textcolor[rgb]{0.00,0.50,0.00}{##1}}}
\expandafter\def\csname PY@tok@bp\endcsname{\def\PY@tc##1{\textcolor[rgb]{0.00,0.50,0.00}{##1}}}
\expandafter\def\csname PY@tok@c1\endcsname{\let\PY@it=\textit\def\PY@tc##1{\textcolor[rgb]{0.25,0.50,0.50}{##1}}}
\expandafter\def\csname PY@tok@o\endcsname{\def\PY@tc##1{\textcolor[rgb]{0.40,0.40,0.40}{##1}}}
\expandafter\def\csname PY@tok@kc\endcsname{\let\PY@bf=\textbf\def\PY@tc##1{\textcolor[rgb]{0.00,0.50,0.00}{##1}}}
\expandafter\def\csname PY@tok@c\endcsname{\let\PY@it=\textit\def\PY@tc##1{\textcolor[rgb]{0.25,0.50,0.50}{##1}}}
\expandafter\def\csname PY@tok@mf\endcsname{\def\PY@tc##1{\textcolor[rgb]{0.40,0.40,0.40}{##1}}}
\expandafter\def\csname PY@tok@err\endcsname{\def\PY@bc##1{\setlength{\fboxsep}{0pt}\fcolorbox[rgb]{1.00,0.00,0.00}{1,1,1}{\strut ##1}}}
\expandafter\def\csname PY@tok@mb\endcsname{\def\PY@tc##1{\textcolor[rgb]{0.40,0.40,0.40}{##1}}}
\expandafter\def\csname PY@tok@ss\endcsname{\def\PY@tc##1{\textcolor[rgb]{0.10,0.09,0.49}{##1}}}
\expandafter\def\csname PY@tok@sr\endcsname{\def\PY@tc##1{\textcolor[rgb]{0.73,0.40,0.53}{##1}}}
\expandafter\def\csname PY@tok@mo\endcsname{\def\PY@tc##1{\textcolor[rgb]{0.40,0.40,0.40}{##1}}}
\expandafter\def\csname PY@tok@kd\endcsname{\let\PY@bf=\textbf\def\PY@tc##1{\textcolor[rgb]{0.00,0.50,0.00}{##1}}}
\expandafter\def\csname PY@tok@mi\endcsname{\def\PY@tc##1{\textcolor[rgb]{0.40,0.40,0.40}{##1}}}
\expandafter\def\csname PY@tok@kn\endcsname{\let\PY@bf=\textbf\def\PY@tc##1{\textcolor[rgb]{0.00,0.50,0.00}{##1}}}
\expandafter\def\csname PY@tok@cpf\endcsname{\let\PY@it=\textit\def\PY@tc##1{\textcolor[rgb]{0.25,0.50,0.50}{##1}}}
\expandafter\def\csname PY@tok@kr\endcsname{\let\PY@bf=\textbf\def\PY@tc##1{\textcolor[rgb]{0.00,0.50,0.00}{##1}}}
\expandafter\def\csname PY@tok@s\endcsname{\def\PY@tc##1{\textcolor[rgb]{0.73,0.13,0.13}{##1}}}
\expandafter\def\csname PY@tok@kp\endcsname{\def\PY@tc##1{\textcolor[rgb]{0.00,0.50,0.00}{##1}}}
\expandafter\def\csname PY@tok@w\endcsname{\def\PY@tc##1{\textcolor[rgb]{0.73,0.73,0.73}{##1}}}
\expandafter\def\csname PY@tok@kt\endcsname{\def\PY@tc##1{\textcolor[rgb]{0.69,0.00,0.25}{##1}}}
\expandafter\def\csname PY@tok@sc\endcsname{\def\PY@tc##1{\textcolor[rgb]{0.73,0.13,0.13}{##1}}}
\expandafter\def\csname PY@tok@sb\endcsname{\def\PY@tc##1{\textcolor[rgb]{0.73,0.13,0.13}{##1}}}
\expandafter\def\csname PY@tok@k\endcsname{\let\PY@bf=\textbf\def\PY@tc##1{\textcolor[rgb]{0.00,0.50,0.00}{##1}}}
\expandafter\def\csname PY@tok@se\endcsname{\let\PY@bf=\textbf\def\PY@tc##1{\textcolor[rgb]{0.73,0.40,0.13}{##1}}}
\expandafter\def\csname PY@tok@sd\endcsname{\let\PY@it=\textit\def\PY@tc##1{\textcolor[rgb]{0.73,0.13,0.13}{##1}}}

\def\PYZbs{\char`\\}
\def\PYZus{\char`\_}
\def\PYZob{\char`\{}
\def\PYZcb{\char`\}}

\def\PYZgt{\char`\>}
\def\PYZsh{\char`\#}
\def\PYZpc{\char`\%}

\def\PYZhy{\char`\-}
\def\PYZsq{\char`\'}
\def\PYZdq{\char`\"}

\makeatother

    \definecolor{incolor}{rgb}{0.0, 0.0, 0.5}
    \definecolor{outcolor}{rgb}{0.545, 0.0, 0.0}

\graphicspath{{./fig/}{./fig/plots/}}

\newcommand{\footlabel}[2]{%
    \addtocounter{footnote}{1}%
    \footnotetext[\thefootnote]{%
        \addtocounter{footnote}{-1}%
        \refstepcounter{footnote}\label{#1}%
        #2%
    }%
    $^{\ref{#1}}$%
}

\newcommand{\footref}[1]{%
    $^{\ref{#1}}$%
}

 \makeatletter
 \DeclareRobustCommand\onedot{\futurelet\@let@token\@onedot}
 \def\@onedot{\ifx\@let@token.\else.\null\fi\xspace}

 \makeatother

\DeclareRobustCommand{\figref}[1]{Figure~\ref{#1}}

\newfloatcommand{capbtabbox}{table}[][\FBwidth]

\title{Tutorial on Answering Questions about Images with Deep Learning}

\author{
Mateusz Malinowski \hspace{50pt} Mario Fritz
\\
Max Planck Institute for Informatics \\
Saarbr{\"u}cken, Germany \\
\texttt{\{mmalinow,mfritz\}@mpi-inf.mpg.de} 
}

\nipsfinalcopy %

\begin{document}

\maketitle
\setlength{\textfloatsep}{10pt plus 1.0pt minus 2.0pt}

\begin{abstract}
Together with the development of more accurate methods in Computer Vision and Natural Language Understanding, holistic architectures that  answer on questions about the content of real-world images have emerged.
In this tutorial, we build  a neural-based approach to answer questions about images. We base our tutorial on two datasets: (mostly on) DAQUAR, and (a bit on) VQA. With small tweaks the models that we present here can achieve a  competitive performance on both datasets, in fact, they are among the best methods that use a combination of LSTM with a global, full frame CNN representation of an image.  We hope that after reading this tutorial, the reader will be able to use  Deep Learning frameworks, such as Keras and introduced Kraino, to build various architectures that will lead to a further performance improvement on this challenging task.
\end{abstract}
	
\section{Preface}
In this tutorial\footnote{This tutorial was presented for the first time during the  2nd Summer School on Integrating Vision and Language: Deep Learning.} 
we build a few architectures that can answer questions about images. The architectures are based on our two papers on this topic: \citet{malinowski2015ask} and \citet{malinowski2016ask}; and more broadly, on our project towards a Visual Turing Test\footnote{\url{http://mpii.de/visual_turing_test}}. 
In particular, an encoder-decoder perspective of \citet{malinowski2016ask} allows us to effectively experiment with various design choices. For the sake of simplicity, we only consider a classification-based approach to answer questions about images, although an approach that generate answers word-by-word is also studied in the community \citep{malinowski2015ask}. In the tutorial, we mainly focus on the DAQUAR dataset \citep{malinowski14nips}, but a few possible directions to apply learnt techniques to VQA \citep{antol2015vqa} are also pointed. First, we will get familiar with the task of answering questions about images, and a dataset that implements the task (due to a small size, we mainly use DAQUAR as it better serves an educational purpose that we aim at this tutorial). Next, we build a few blind models that answer questions about images without actually seeing such images. Such models already exhibit a reasonable performance as they can effectively learn various biases that exist in a dataset, which we also interpret as learning a common sense knowledge \citep{malinowski2015ask, malinowski2016ask}. Subsequently, we build a few language+vision models that answer questions based on both a textual and a visual inputs. Finally, we leave the tutorial with a few possible research directions. 

\paragraph{Technical aspects} The tutorial is originally written using Python Notebook, which the reader is welcome to download\footlabel{tutorial:tutorial_url}{\url{https://github.com/mateuszmalinowski/visual_turing_test-tutorial/blob/master/visual_turing_test.ipynb}} and use through the tutorial. Instructions necessary to run the Notebook version of this tutorial are provided in the following: \url{https://github.com/mateuszmalinowski/visual_turing_test-tutorial}. In this tutorial, we heavily use a Python code, and therefore it is expected the reader either already knows this language, or can quickly learn it. However, we made an effort to make this tutorial approachable to a wider audience. We use Kraino\footref{tutorial:tutorial_url} that is a framework prepared for this tutorial in order to simplify the development of the question answering architectures. Under the hood, it uses Theano\footnote{\url{http://deeplearning.net/software/theano/}} \citep{Bastien-Theano-2012} and Keras\footnote{\url{https://keras.io}} \citep{chollet2015} -- two frameworks to build Deep Learning models.
We also use various CNNs representations extracted from images that can be downloaded as explained at the beginning of our Notebook tutorial\footref{tutorial:tutorial_url}.
We also highlight exercises that a curious reader may attempt to solve.

\section{Dataset}\label{datasets}
    \begin{figure}[htbp]
\centering
\includegraphics[width=0.75\linewidth]{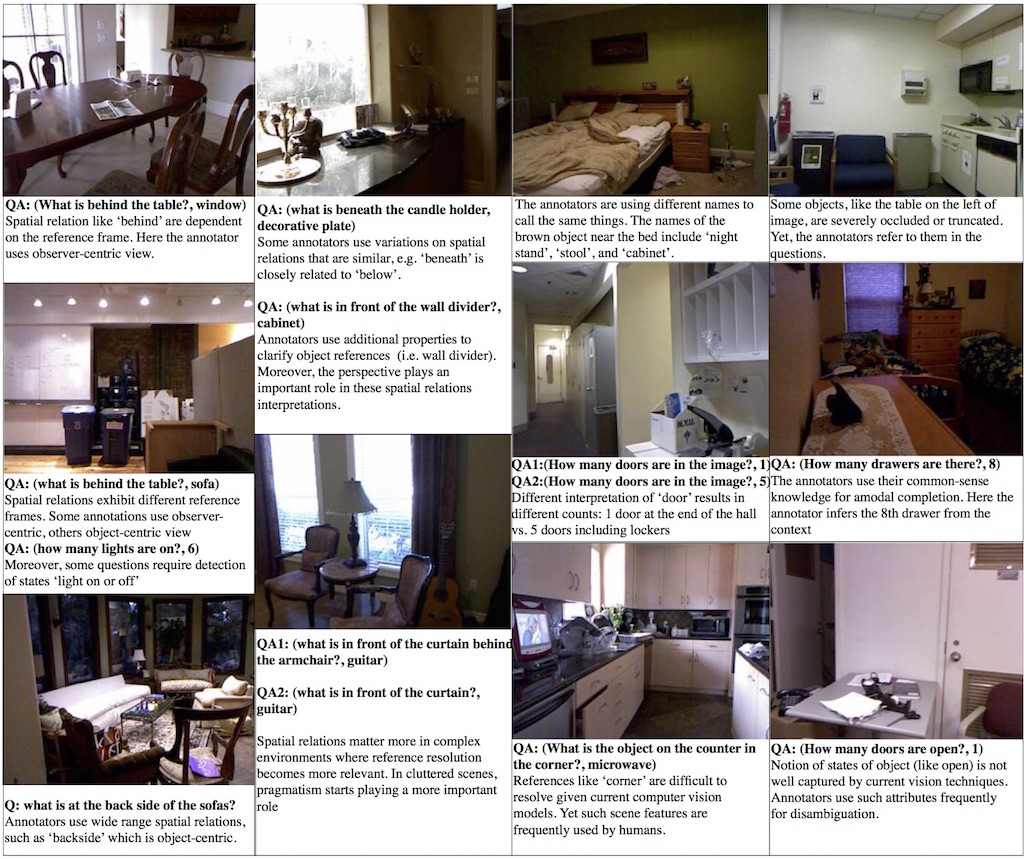}
\caption{Challenges present in the DAQUAR dataset.}
\label{tutorial:challenges}
\end{figure}
This section introduces the DAQUAR dataset \citep{malinowski14nips} from a programming perspective.
Let us first list a few DAQUAR entries to become familiar with the format.

    \begin{Verbatim}[commandchars=\\\{\}]
{\color{incolor}In [{\color{incolor}1}]:} \PY{o}{!} head \PYZhy{}15 data/daquar/qa.894.raw.train.format\PYZus{}triple
\end{Verbatim}

    \begin{Verbatim}[commandchars=\\\{\}]
what is on the right side of the black telephone and on the left side of the red chair ?
desk
image3
what is in front of the white door on the left side of the desk ?
telephone
image3
what is on the desk ?
book, scissor, papers, tape\_dispenser
image3
what is the largest brown objects ?
carton
image3
what color is the chair in front of the white wall ?
red
image3
\end{Verbatim}
Note that the format is: question, answer (could be many answer words), and the image.
Let us have a look at \figref{tutorial:challenges}. The figure
lists images with associated question-answer pairs. It also comments on
challenges associated with  question-answer-image triplets. We see that to
answer properly on the wide range of questions, an answerer not only needs to understand the
scene visually or to just understand the question, but also, arguably, has to
resort to the common sense knowledge, or even know the preferences of the
person asking the question, e.g. what `behind' exactly
means in `What is behind the table?'. Hence, architectures that answer questions about images have to face many challenges.
Ambiguities make it also difficult to judge the provided answers. We revisit this issue in a later section.
Meantime, a curious reader may try to answer the following question.

\begin{verbatim}
	Can you spot ambiguities that are present in the first column of the figure?
	Think of a spatial relationship between an observer, object of interest, and the world. 
\end{verbatim}

The following code returns a dictionary of three views on the
DAQUAR dataset. For now, we look only into the `text'
view. dp{[}`text'{]} returns a function from a dataset split
into the dataset's textual view. Executing the following code makes it more clear.
    \begin{Verbatim}[commandchars=\\\{\}]
{\color{incolor}In [{\color{incolor} }]:} \PY{c+c1}{\PYZsh{}TODO: Execute the following procedure (Shift+Enter in the Notebook)}
        \PY{k+kn}{from} \PY{n+nn}{kraino.utils} \PY{k+kn}{import} \PY{n}{data\PYZus{}provider}
        
        \PY{n}{dp} \PY{o}{=} \PY{n}{data\PYZus{}provider}\PY{o}{.}\PY{n}{select}\PY{p}{[}\PY{l+s+s1}{\PYZsq{}}\PY{l+s+s1}{daquar\PYZhy{}triples}\PY{l+s+s1}{\PYZsq{}}\PY{p}{]}
        \PY{n}{train\PYZus{}text\PYZus{}representation} \PY{o}{=} \PY{n}{dp}\PY{p}{[}\PY{l+s+s1}{\PYZsq{}}\PY{l+s+s1}{text}\PY{l+s+s1}{\PYZsq{}}\PY{p}{]}\PY{p}{(}\PY{n}{train\PYZus{}or\PYZus{}test}\PY{o}{=}\PY{l+s+s1}{\PYZsq{}}\PY{l+s+s1}{train}\PY{l+s+s1}{\PYZsq{}}\PY{p}{)}
\end{Verbatim}

\noindent This view specifies how questions are ended (`?'), answers are
ended (`.'), answer words are delimited (DAQUAR sometimes has a set of
answer words as an answer, for instance `knife, fork' may be an answer
answer), but most important, it has questions (key `x'), answers (key
`y'), and names of the corresponding images (key `img\_name').

    \begin{Verbatim}[commandchars=\\\{\}]
{\color{incolor}In [{\color{incolor} }]:} \PY{c+c1}{\PYZsh{} let us check some entries of the text\PYZsq{}s representation}
        \PY{n}{n\PYZus{}elements} \PY{o}{=} \PY{l+m+mi}{10}
        \PY{k}{print}\PY{p}{(}\PY{l+s+s1}{\PYZsq{}}\PY{l+s+s1}{== Questions:}\PY{l+s+s1}{\PYZsq{}}\PY{p}{)}
        \PY{n}{print\PYZus{}list}\PY{p}{(}\PY{n}{train\PYZus{}text\PYZus{}representation}\PY{p}{[}\PY{l+s+s1}{\PYZsq{}}\PY{l+s+s1}{x}\PY{l+s+s1}{\PYZsq{}}\PY{p}{]}\PY{p}{[}\PY{p}{:}\PY{n}{n\PYZus{}elements}\PY{p}{]}\PY{p}{)}
        \PY{k}{print}
        \PY{k}{print}\PY{p}{(}\PY{l+s+s1}{\PYZsq{}}\PY{l+s+s1}{== Answers:}\PY{l+s+s1}{\PYZsq{}}\PY{p}{)}
        \PY{n}{print\PYZus{}list}\PY{p}{(}\PY{n}{train\PYZus{}text\PYZus{}representation}\PY{p}{[}\PY{l+s+s1}{\PYZsq{}}\PY{l+s+s1}{y}\PY{l+s+s1}{\PYZsq{}}\PY{p}{]}\PY{p}{[}\PY{p}{:}\PY{n}{n\PYZus{}elements}\PY{p}{]}\PY{p}{)}
        \PY{k}{print}
        \PY{k}{print}\PY{p}{(}\PY{l+s+s1}{\PYZsq{}}\PY{l+s+s1}{== Image Names:}\PY{l+s+s1}{\PYZsq{}}\PY{p}{)}
        \PY{n}{print\PYZus{}list}\PY{p}{(}\PY{n}{train\PYZus{}text\PYZus{}representation}\PY{p}{[}\PY{l+s+s1}{\PYZsq{}}\PY{l+s+s1}{img\PYZus{}name}\PY{l+s+s1}{\PYZsq{}}\PY{p}{]}\PY{p}{[}\PY{p}{:}\PY{n}{n\PYZus{}elements}\PY{p}{]}\PY{p}{)}
\end{Verbatim}
\paragraph{Summary}
DAQUAR consists of question-answer-image triplets. Question-answer pairs for different folds are accessible from executing the following code.
\begin{Shaded}
\begin{Highlighting}[]
\NormalTok{data_provider.select[}\StringTok{'text'}\NormalTok{]}
\end{Highlighting}
\end{Shaded}
Finally, as we see in \figref{tutorial:challenges}, DAQUAR poses many challenges in designing good architectures, or evaluation metrics.

\section{Textual Features}\label{textual-features}
\begin{figure}[htbp]
\centering
\includegraphics[width=0.75\linewidth]{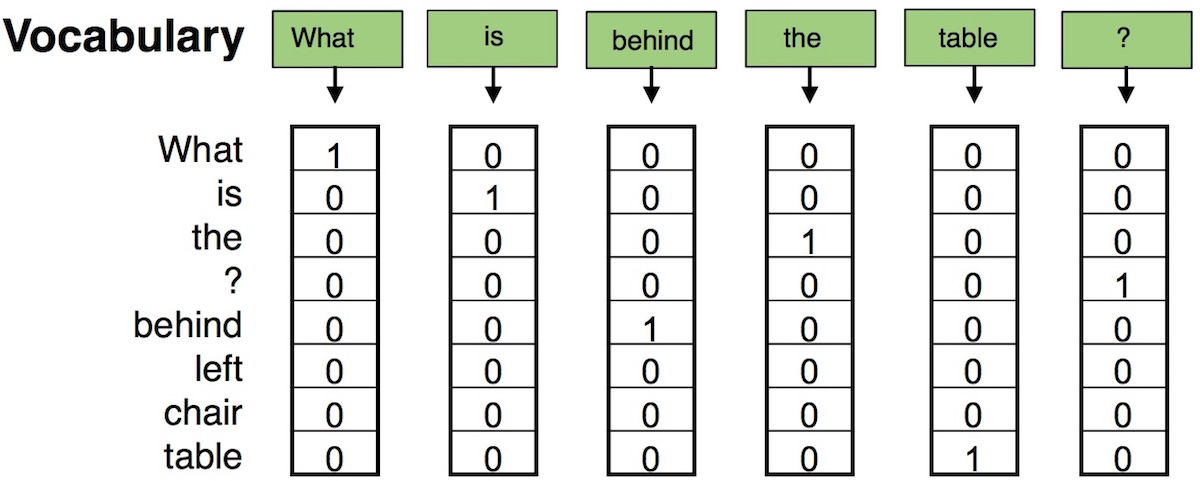}
\caption{One hot representations of the textual words in the question.}
\label{tutorial:one_hot}
\end{figure}

We have an access to a textual representation of questions. This is however not very helpful since neural networks expect a numerical
input, and hence we cannot really work with the raw text. We need to transform
the textual input into some numerical value or a vector of values. One
particularly successful representation is called one-hot vector and it
is a binary vector with exactly one non-zero entry. This entry points to
the corresponding word in the vocabulary. See the illustration shown in \figref{tutorial:one_hot}.

\noindent The reader can pause here a bit to answer the following questions.
    \begin{verbatim}
		Can you sum up the one-hot vectors for the `What table is behind the table?'.
		How would you interpret the resulting vector?
		Why is it a good idea to work with one-hot vector represetantions of the text?
\end{verbatim}

\noindent As we see from the illustrative example above, we first need to build a
suitable vocabulary from our raw textual training data, and next
transform them into one-hot representations. The following code can do this.

    \begin{Verbatim}[commandchars=\\\{\}]
{\color{incolor}In [{\color{incolor} }]:} \PY{k+kn}{from} \PY{n+nn}{toolz} \PY{k+kn}{import} \PY{n}{frequencies}
        \PY{n}{train\PYZus{}raw\PYZus{}x} \PY{o}{=} \PY{n}{train\PYZus{}text\PYZus{}representation}\PY{p}{[}\PY{l+s+s1}{\PYZsq{}}\PY{l+s+s1}{x}\PY{l+s+s1}{\PYZsq{}}\PY{p}{]}
        \PY{c+c1}{\PYZsh{} we start from building the frequencies table}
        \PY{n}{wordcount\PYZus{}x} \PY{o}{=} \PY{n}{frequencies}\PY{p}{(}\PY{l+s+s1}{\PYZsq{}}\PY{l+s+s1}{ }\PY{l+s+s1}{\PYZsq{}}\PY{o}{.}\PY{n}{join}\PY{p}{(}\PY{n}{train\PYZus{}raw\PYZus{}x}\PY{p}{)}\PY{o}{.}\PY{n}{split}\PY{p}{(}\PY{l+s+s1}{\PYZsq{}}\PY{l+s+s1}{ }\PY{l+s+s1}{\PYZsq{}}\PY{p}{)}\PY{p}{)}
        \PY{c+c1}{\PYZsh{} print the most and least frequent words}
        \PY{n}{n\PYZus{}show} \PY{o}{=} \PY{l+m+mi}{5}
        \PY{k}{print}\PY{p}{(}\PY{n+nb}{sorted}\PY{p}{(}\PY{n}{wordcount\PYZus{}x}\PY{o}{.}\PY{n}{items}\PY{p}{(}\PY{p}{)}\PY{p}{,} \PY{n}{key}\PY{o}{=}\PY{k}{lambda} \PY{n}{x}\PY{p}{:} \PY{n}{x}\PY{p}{[}\PY{l+m+mi}{1}\PY{p}{]}\PY{p}{,} \PY{n}{reverse}\PY{o}{=}\PY{n+nb+bp}{True}\PY{p}{)}\PY{p}{[}\PY{p}{:}\PY{n}{n\PYZus{}show}\PY{p}{]}\PY{p}{)}
        \PY{k}{print}\PY{p}{(}\PY{n+nb}{sorted}\PY{p}{(}\PY{n}{wordcount\PYZus{}x}\PY{o}{.}\PY{n}{items}\PY{p}{(}\PY{p}{)}\PY{p}{,} \PY{n}{key}\PY{o}{=}\PY{k}{lambda} \PY{n}{x}\PY{p}{:} \PY{n}{x}\PY{p}{[}\PY{l+m+mi}{1}\PY{p}{]}\PY{p}{)}\PY{p}{[}\PY{p}{:}\PY{n}{n\PYZus{}show}\PY{p}{]}\PY{p}{)}
\end{Verbatim}

\noindent In many parts of this tutorial, we use Kraino, which was developed for the purpose of this tutorial to simplify the development of various `question answering' models through prototyping.
    \begin{Verbatim}[commandchars=\\\{\}]
        \PY{k+kn}{from} \PY{n+nn}{kraino.utils.input\PYZus{}output\PYZus{}space} \PY{k+kn}{import} \PY{n}{build\PYZus{}vocabulary}
        
        \PY{c+c1}{\PYZsh{} This function takes wordcounts, }
        \PY{c+c1}{\PYZsh{} and returns word2index \PYZhy{} mapping from words into indices, }
        \PY{c+c1}{\PYZsh{} and index2word \PYZhy{} mapping from indices to words.}
        \PY{n}{word2index\PYZus{}x}\PY{p}{,} \PY{n}{index2word\PYZus{}x} \PY{o}{=} \PY{n}{build\PYZus{}vocabulary}\PY{p}{(}
            \PY{n}{this\PYZus{}wordcount}\PY{o}{=}\PY{n}{wordcount\PYZus{}x}\PY{p}{,}
            \PY{n}{truncate\PYZus{}to\PYZus{}most\PYZus{}frequent}\PY{o}{=}\PY{l+m+mi}{0}\PY{p}{)}
        \PY{n}{word2index\PYZus{}x}
\end{Verbatim}

\noindent In addition, we use a few special, extra symbols that do not occur in the
training dataset. Most important are $<pad>$ and $<unk>$. We  use
the former to pad sequences in order to have the same number of temporal
elements; we use the latter for words (at test time) that do not
exist in the training set.
Armed with the vocabulary, we can build one-hot representations of the
training data. However, this is not necessary and may even be wasteful.
Our one-hot representation of the input text does not explicitly build
long and sparse vectors, but instead it operates on indices. The example from \figref{tutorial:one_hot}
would be encoded as {[}0,1,4,2,7,3{]}.

Due to the sparsity existing in the one-hot representation, we can more efficiently operate on indices instead of performing full linear transformations by matrix-vector multiplications. This is reflected in the following claim.
\paragraph{Claim:}
Let $x$ be a binary vector with exactly one value $1$ at the position
$index$, that is $x[index]=1$. Then \[W[:,index] = Wx\] where $W[:,b]$
denotes a vector built from a column $b$ of $W$. This shows that matrix-vector multiplication can be replaced by retrieving a right vector of parameters according to the index.

\begin{verbatim}
Can you show that the claim is valid?
\end{verbatim}
We can encode textual questions into one-hot vector representations by executing the following code.
    \begin{Verbatim}[commandchars=\\\{\}]
{\color{incolor}In [{\color{incolor} }]:} \PY{k+kn}{from} \PY{n+nn}{kraino.utils.input\PYZus{}output\PYZus{}space} \PY{k+kn}{import} \PY{n}{encode\PYZus{}questions\PYZus{}index}
        \PY{n}{one\PYZus{}hot\PYZus{}x} \PY{o}{=} \PY{n}{encode\PYZus{}questions\PYZus{}index}\PY{p}{(}\PY{n}{train\PYZus{}raw\PYZus{}x}\PY{p}{,} \PY{n}{word2index\PYZus{}x}\PY{p}{)}
        \PY{k}{print}\PY{p}{(}\PY{n}{train\PYZus{}raw\PYZus{}x}\PY{p}{[}\PY{p}{:}\PY{l+m+mi}{3}\PY{p}{]}\PY{p}{)}
        \PY{k}{print}\PY{p}{(}\PY{n}{one\PYZus{}hot\PYZus{}x}\PY{p}{[}\PY{p}{:}\PY{l+m+mi}{3}\PY{p}{]}\PY{p}{)}
\end{Verbatim}

\noindent As we can see, the sequences have different number of elements. We can pad the
sequences to have the same length by setting up $MAXLEN$.

    \begin{Verbatim}[commandchars=\\\{\}]
        \PY{k+kn}{from} \PY{n+nn}{keras.preprocessing} \PY{k+kn}{import} \PY{n}{sequence}
        \PY{n}{MAXLEN}\PY{o}{=}\PY{l+m+mi}{30}
        \PY{n}{train\PYZus{}x} \PY{o}{=} \PY{n}{sequence}\PY{o}{.}\PY{n}{pad\PYZus{}sequences}\PY{p}{(}\PY{n}{one\PYZus{}hot\PYZus{}x}\PY{p}{,} \PY{n}{maxlen}\PY{o}{=}\PY{n}{MAXLEN}\PY{p}{)}
        \PY{n}{train\PYZus{}x}\PY{p}{[}\PY{p}{:}\PY{l+m+mi}{3}\PY{p}{]}
\end{Verbatim}
We do the same with the answers.
    \begin{Verbatim}[commandchars=\\\{\}]
{\color{incolor}In [{\color{incolor} }]:} \PY{c+c1}{\PYZsh{} for simplicity, we consider only first answer words;}
{\color{incolor}In [{\color{incolor} }]:} \PY{c+c1}{\PYZsh{} that is, if answer is \PYZsq{}knife,fork\PYZsq{} we encode only \PYZsq{}knife\PYZsq{}}
        \PY{n}{MAX\PYZus{}ANSWER\PYZus{}TIME\PYZus{}STEPS}\PY{o}{=}\PY{l+m+mi}{1}
        
        \PY{k+kn}{from} \PY{n+nn}{kraino.utils.input\PYZus{}output\PYZus{}space} \PY{k+kn}{import} \PY{n}{encode\PYZus{}answers\PYZus{}one\PYZus{}hot}
        \PY{n}{train\PYZus{}raw\PYZus{}y} \PY{o}{=} \PY{n}{train\PYZus{}text\PYZus{}representation}\PY{p}{[}\PY{l+s+s1}{\PYZsq{}}\PY{l+s+s1}{y}\PY{l+s+s1}{\PYZsq{}}\PY{p}{]}
        \PY{n}{wordcount\PYZus{}y} \PY{o}{=} \PY{n}{frequencies}\PY{p}{(}\PY{l+s+s1}{\PYZsq{}}\PY{l+s+s1}{ }\PY{l+s+s1}{\PYZsq{}}\PY{o}{.}\PY{n}{join}\PY{p}{(}\PY{n}{train\PYZus{}raw\PYZus{}y}\PY{p}{)}\PY{o}{.}\PY{n}{split}\PY{p}{(}\PY{l+s+s1}{\PYZsq{}}\PY{l+s+s1}{ }\PY{l+s+s1}{\PYZsq{}}\PY{p}{)}\PY{p}{)}
        \PY{n}{word2index\PYZus{}y}\PY{p}{,} \PY{n}{index2word\PYZus{}y} \PY{o}{=} \PY{n}{build\PYZus{}vocabulary}\PY{p}{(}\PY{n}{this\PYZus{}wordcount}\PY{o}{=}\PY{n}{wordcount\PYZus{}y}\PY{p}{)}
        \PY{n}{train\PYZus{}y}\PY{p}{,} \PY{n}{\PYZus{}} \PY{o}{=} \PY{n}{encode\PYZus{}answers\PYZus{}one\PYZus{}hot}\PY{p}{(}
            \PY{n}{train\PYZus{}raw\PYZus{}y}\PY{p}{,} 
            \PY{n}{word2index\PYZus{}y}\PY{p}{,} 
            \PY{n}{answer\PYZus{}words\PYZus{}delimiter}\PY{o}{=}\PY{n}{train\PYZus{}text\PYZus{}representation}\PY{p}{[}\PY{l+s+s1}{\PYZsq{}}\PY{l+s+s1}{answer\PYZus{}words\PYZus{}delimiter}\PY{l+s+s1}{\PYZsq{}}\PY{p}{]}\PY{p}{,}
            \PY{n}{is\PYZus{}only\PYZus{}first\PYZus{}answer\PYZus{}word}\PY{o}{=}\PY{n+nb+bp}{True}\PY{p}{,}
            \PY{n}{max\PYZus{}answer\PYZus{}time\PYZus{}steps}\PY{o}{=}\PY{n}{MAX\PYZus{}ANSWER\PYZus{}TIME\PYZus{}STEPS}\PY{p}{)}
        \PY{k}{print}\PY{p}{(}\PY{n}{train\PYZus{}x}\PY{o}{.}\PY{n}{shape}\PY{p}{)}
        \PY{k}{print}\PY{p}{(}\PY{n}{train\PYZus{}y}\PY{o}{.}\PY{n}{shape}\PY{p}{)}
\end{Verbatim}
At the last step, we  encode test questions. We need them later to see how
well our models generalize to new question-answer-image triplets.
Remember, however, that we should use the vocabulary we generated from the training samples.

\begin{verbatim}
Why should we use the training vocabulary to encode test questions?
\end{verbatim}

    \begin{Verbatim}[commandchars=\\\{\}]
{\color{incolor}In [{\color{incolor} }]:} \PY{n}{test\PYZus{}text\PYZus{}representation} \PY{o}{=} \PY{n}{dp}\PY{p}{[}\PY{l+s+s1}{\PYZsq{}}\PY{l+s+s1}{text}\PY{l+s+s1}{\PYZsq{}}\PY{p}{]}\PY{p}{(}\PY{n}{train\PYZus{}or\PYZus{}test}\PY{o}{=}\PY{l+s+s1}{\PYZsq{}}\PY{l+s+s1}{test}\PY{l+s+s1}{\PYZsq{}}\PY{p}{)}
        \PY{n}{test\PYZus{}raw\PYZus{}x} \PY{o}{=} \PY{n}{test\PYZus{}text\PYZus{}representation}\PY{p}{[}\PY{l+s+s1}{\PYZsq{}}\PY{l+s+s1}{x}\PY{l+s+s1}{\PYZsq{}}\PY{p}{]}
        \PY{n}{test\PYZus{}one\PYZus{}hot\PYZus{}x} \PY{o}{=} \PY{n}{encode\PYZus{}questions\PYZus{}index}\PY{p}{(}\PY{n}{test\PYZus{}raw\PYZus{}x}\PY{p}{,} \PY{n}{word2index\PYZus{}x}\PY{p}{)}
        \PY{n}{test\PYZus{}x} \PY{o}{=} \PY{n}{sequence}\PY{o}{.}\PY{n}{pad\PYZus{}sequences}\PY{p}{(}\PY{n}{test\PYZus{}one\PYZus{}hot\PYZus{}x}\PY{p}{,} \PY{n}{maxlen}\PY{o}{=}\PY{n}{MAXLEN}\PY{p}{)}
        \PY{n}{print\PYZus{}list}\PY{p}{(}\PY{n}{test\PYZus{}raw\PYZus{}x}\PY{p}{[}\PY{p}{:}\PY{l+m+mi}{3}\PY{p}{]}\PY{p}{)}
        \PY{n}{test\PYZus{}x}\PY{p}{[}\PY{p}{:}\PY{l+m+mi}{3}\PY{p}{]}
\end{Verbatim}
With the encoded question-answer pairs we finish this section. However,
before delving into details of building and training new models, let us have a look
at the summary to see bigger picture.

\paragraph{Summary}
We started from raw questions from the training set. We use them to build a
vocabulary. Next, we encode questions into sequences of one-hot vectors
based on the vocabulary. Finally, we use the same vocabulary to encode
questions from the test set. If a word is absent, we use an extra token $<unk>$
to denote this fact, so that we encode the $<unk>$ token, not the word.

\section{Language Only Models}\label{language-only}
\subsubsection{Training}\label{training---overall-picture}
As you may already know, we train models by weights updates. Let $x$ and
$y$ be training samples (an input, and an output), and $\ell(x,y)$ be an objective
function. The formula for weights updates is:
\[w := w - \alpha \nabla \ell(x,y; w)\] with $\alpha$ that we call the
learning rate, and $\nabla$ that is a gradient wrt. the weights $w$. The learning rate is a
hyper-parameter that must be set in advance. The rule shown above is
called the SGD update, but other variants are also possible. In fact, we
 use its variant called ADAM \citep{kingma2014adam}.

We cast the question answering problem into a classification framework,
so that we classify an input $x$ into some class that represents an answer
word. Therefore, we use, commonly used in the classification, logistic regression
as the objective:
\[\ell(x,y;w):=\sum_{y' \in \mathcal{C}} \mathbbm{1}\{y'=y\}\log p(y'\;|\;x,w)\]
where $\mathcal{C}$ is a set of all classes, and $p(y\;|\;x,w)$ is the
  softmax: $e^{w^y\phi(x)} / \sum_{z}e^{w^z\phi(x)}$. Here
$\phi(x)$ denotes an output of a model (more precisely, it is often a response of a
neural network to the input, just before softmax of the neural network is applied).
Note, however, that another variant of providing answers, called the answer generation, is also possible \citep{malinowski2015ask}.
For training, we need to execute the following code.
\begin{Shaded}
\begin{Highlighting}[]
\NormalTok{training(gradient_of_the_model, optimizer=}\StringTok{'Adam'}\NormalTok{)}
\end{Highlighting}
\end{Shaded}
\paragraph{Summary} Given a model, and an optimization procedure (SGD,
Adam, etc.) all we need is to compute gradient of the model
$\nabla \ell(x,y;w)$ wrt. to its parameters $w$, and next plug it to the
optimization procedure.
\subsubsection{Theano}\label{theano}

    Since computing gradients $\nabla \ell(x,y;w)$ may quickly become 
tedious, especially for more complex models, we search for tools that
could automatize this process. Imagine that you build a model $M$ and
you get its gradient $\nabla M$ by just executing the tool, something like the following piece of code.

\begin{Shaded}
\begin{Highlighting}[]
\NormalTok{nabla_M = compute_gradient_symbolically(M,x,y)}
\end{Highlighting}
\end{Shaded}
This would definitely speed up prototyping.
Theano \citep{Bastien-Theano-2012} is such a tool
that is specifically tailored to work with deep learning models. For a
broader understanding of Theano, you can check a suitable tutorial\footnote{For instance, \url{http://deeplearning.net/tutorial/}.}.

The following coding example defines ReLU, a popular
activation function defined as $ReLU(x) = \max(x,0)$, as well as derive its derivative using Theano.
Note however that, with this example, we obviously only scratch the surface.

    \begin{Verbatim}[commandchars=\\\{\}]
{\color{incolor}In [{\color{incolor} }]:} \PY{k+kn}{import} \PY{n+nn}{theano}
        \PY{k+kn}{import} \PY{n+nn}{theano.tensor} \PY{k+kn}{as} \PY{n+nn}{T}
        
        \PY{c+c1}{\PYZsh{} Theano uses symbolic calculations,}
		\PY{c+c1}{\PYZsh{} so we need to first create symbolic variables}
        \PY{n}{theano\PYZus{}x} \PY{o}{=} \PY{n}{T}\PY{o}{.}\PY{n}{scalar}\PY{p}{(}\PY{p}{)}
        \PY{c+c1}{\PYZsh{} we define a relationship between a symbolic input and a symbolic output}
        \PY{n}{theano\PYZus{}y} \PY{o}{=} \PY{n}{T}\PY{o}{.}\PY{n}{maximum}\PY{p}{(}\PY{l+m+mi}{0}\PY{p}{,}\PY{n}{theano\PYZus{}x}\PY{p}{)}
        \PY{c+c1}{\PYZsh{} now it\PYZsq{}s time for a symbolic gradient wrt. to symbolic variable x}
        \PY{n}{theano\PYZus{}nabla\PYZus{}y} \PY{o}{=} \PY{n}{T}\PY{o}{.}\PY{n}{grad}\PY{p}{(}\PY{n}{theano\PYZus{}y}\PY{p}{,} \PY{n}{theano\PYZus{}x}\PY{p}{)}
        
        \PY{c+c1}{\PYZsh{} we can see that both variables are symbolic, they don\PYZsq{}t have numerical values}
        \PY{k}{print}\PY{p}{(}\PY{n}{theano\PYZus{}x}\PY{p}{)}
        \PY{k}{print}\PY{p}{(}\PY{n}{theano\PYZus{}y}\PY{p}{)}
        \PY{k}{print}\PY{p}{(}\PY{n}{theano\PYZus{}nabla\PYZus{}y}\PY{p}{)}
        
        \PY{c+c1}{\PYZsh{} theano.function compiles the symbolic representation of the network}
        \PY{n}{theano\PYZus{}f\PYZus{}x} \PY{o}{=} \PY{n}{theano}\PY{o}{.}\PY{n}{function}\PY{p}{(}\PY{p}{[}\PY{n}{theano\PYZus{}x}\PY{p}{]}\PY{p}{,} \PY{n}{theano\PYZus{}y}\PY{p}{)}
        \PY{k}{print}\PY{p}{(}\PY{n}{theano\PYZus{}f\PYZus{}x}\PY{p}{(}\PY{l+m+mi}{3}\PY{p}{)}\PY{p}{)}
        \PY{k}{print}\PY{p}{(}\PY{n}{theano\PYZus{}f\PYZus{}x}\PY{p}{(}\PY{o}{\PYZhy{}}\PY{l+m+mi}{3}\PY{p}{)}\PY{p}{)}
        \PY{c+c1}{\PYZsh{} and now for gradients}
        
        \PY{n}{nabla\PYZus{}f\PYZus{}x} \PY{o}{=} \PY{n}{theano}\PY{o}{.}\PY{n}{function}\PY{p}{(}\PY{p}{[}\PY{n}{theano\PYZus{}x}\PY{p}{]}\PY{p}{,} \PY{n}{theano\PYZus{}nabla\PYZus{}y}\PY{p}{)}
        \PY{k}{print}\PY{p}{(}\PY{n}{nabla\PYZus{}f\PYZus{}x}\PY{p}{(}\PY{l+m+mi}{3}\PY{p}{)}\PY{p}{)}
        \PY{k}{print}\PY{p}{(}\PY{n}{nabla\PYZus{}f\PYZus{}x}\PY{p}{(}\PY{o}{\PYZhy{}}\PY{l+m+mi}{3}\PY{p}{)}\PY{p}{)}
\end{Verbatim}

\begin{verbatim}
Can you derive a derivative of ReLU on your own? Consider two cases. 
\end{verbatim}
It should also be mentioned that ReLU is a non-differentiable function at the point $0$, and therefore, technically, we 
compute its sub-gradient -- this is however
still fine for Theano.

\paragraph{Summary}
To compute gradient symbolically, we can use Theano. This  speeds up prototyping, and hence developing new question answering models.

\subsubsection{Keras}\label{keras}

Keras \citep{chollet2015} builds upon Theano, and significantly
simplifies creating new deep learning models as well as training such models,
effectively speeding up the prototyping even further. Keras also
abstracts away from some technical burden such as a symbolic variable
creation. Many examples of using Keras can be found by following the links: \url{https://keras.io/getting-started/sequential-model-guide/},
and \url{https://keras.io/getting-started/functional-api-guide/}. Note that, in the tutorial we use an older sequential model.
Please also pay attention to the version of the Keras, since not all versions are compatible with this tutorial.

\subsubsection{Models}\label{models}

    For the purpose of the Visual Turing Test, and this tutorial, we have
compiled a light framework that builds on top of Keras, and simplify
building and training `question answering' machines. With the tradition of
using  Greek names, we call it Kraino. Note that some parts of the
Kraino, such as a data provider, were already covered in this tutorial.

In the following, we will go through BOW and LSTM approaches to answer
questions about images, but, surprisingly, without the images. It turns
out that a substantial fraction of questions can be answered without an
access to an image, but rather by resorting to a common sense (or
statistics of the dataset). For instance, `what can be placed at the
table?', or `How many eyes this human have?'. Answers like `chair' and `2' are
quite likely to be good answers.

\subsubsection{BOW}\label{bow}

\begin{figure}[htbp]
\centering
\includegraphics[width=0.75\linewidth]{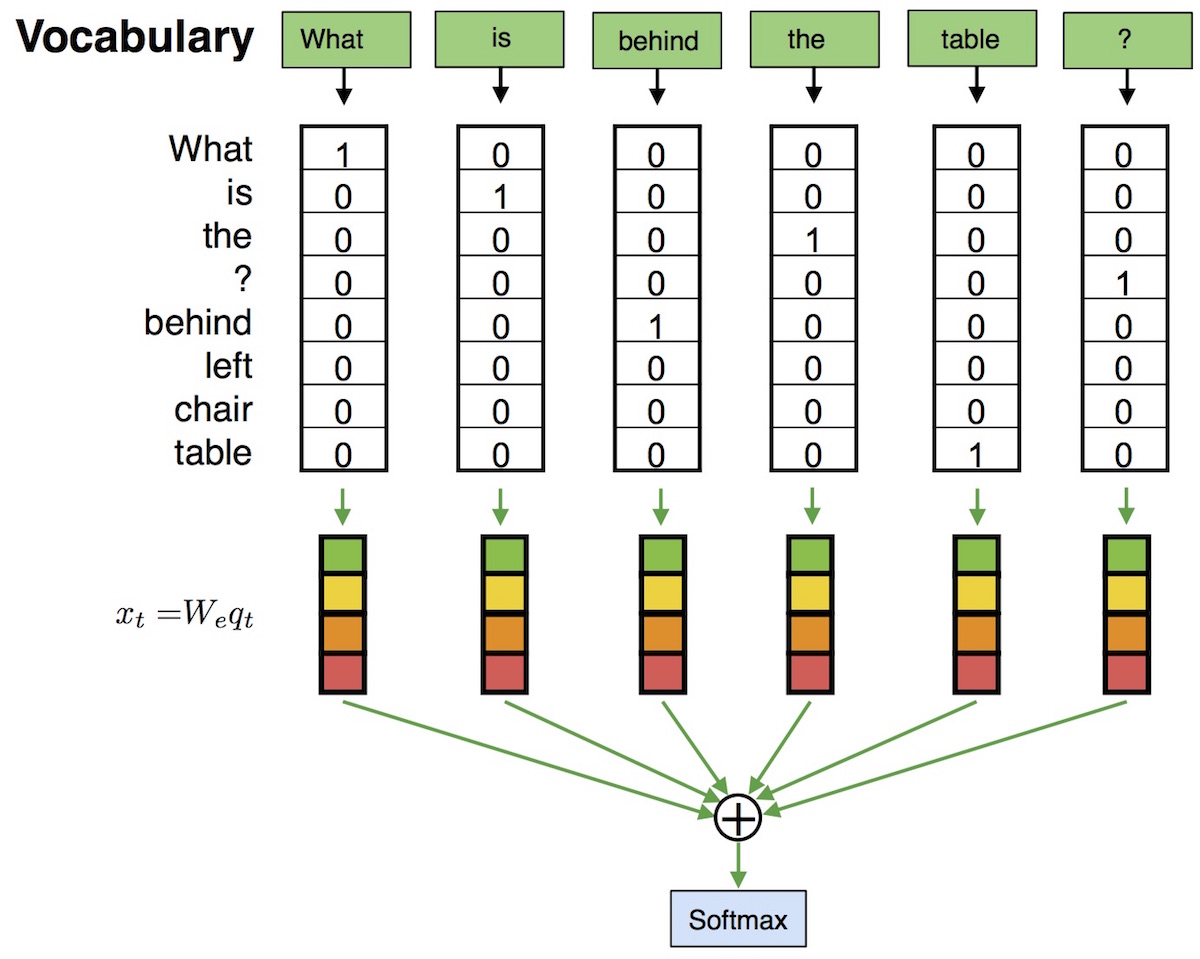}
\caption{Bag-Of-Words (BOW) representation of the input that is next follow by `Softmax'.}
\label{tutorial:bow}
\end{figure}

   \figref{tutorial:bow} illustrates the BOW (Bag Of Words) method. As we have
already seen before, we first
encode the input sentence into one-hot vector representations. Such a
(very) sparse representation is next embedded into a denser space by a
matrix $W_e$. Next, the denser representations are summed up and
classified via `Softmax'. 
Notice that, if $W_e$ were an identity matrix, we would obtain a histogram of the word's occurrences.

\begin{verbatim}
What is your biggest complain about such a BOW representation? 
What happens if instead of 'What is behind the table' we would have 
'is What the behind table'? How does the BOW representation change? 
\end{verbatim}

\noindent Let us now define a BOW model using our tools.
    \begin{Verbatim}[commandchars=\\\{\}]
{\color{incolor}In [{\color{incolor} }]:} \PY{c+c1}{\PYZsh{}== Model definition}
        
        \PY{c+c1}{\PYZsh{} First we define a model using keras/kraino}
        \PY{k+kn}{from} \PY{n+nn}{keras.layers.core} \PY{k+kn}{import} \PY{n}{Activation}
        \PY{k+kn}{from} \PY{n+nn}{keras.layers.core} \PY{k+kn}{import} \PY{n}{Dense}
        \PY{k+kn}{from} \PY{n+nn}{keras.layers.core} \PY{k+kn}{import} \PY{n}{Dropout}
        \PY{k+kn}{from} \PY{n+nn}{keras.layers.core} \PY{k+kn}{import} \PY{n}{TimeDistributedMerge}
        \PY{k+kn}{from} \PY{n+nn}{keras.layers.embeddings} \PY{k+kn}{import} \PY{n}{Embedding}
        
        \PY{k+kn}{from} \PY{n+nn}{kraino.core.model\PYZus{}zoo} \PY{k+kn}{import} \PY{n}{AbstractSequentialModel}
        \PY{k+kn}{from} \PY{n+nn}{kraino.core.model\PYZus{}zoo} \PY{k+kn}{import} \PY{n}{AbstractSingleAnswer}
        \PY{k+kn}{from} \PY{n+nn}{kraino.core.model\PYZus{}zoo} \PY{k+kn}{import} \PY{n}{AbstractSequentialMultiplewordAnswer}
        \PY{k+kn}{from} \PY{n+nn}{kraino.core.model\PYZus{}zoo} \PY{k+kn}{import} \PY{n}{Config}
        \PY{k+kn}{from} \PY{n+nn}{kraino.core.keras\PYZus{}extensions} \PY{k+kn}{import} \PY{n}{DropMask}
        \PY{k+kn}{from} \PY{n+nn}{kraino.core.keras\PYZus{}extensions} \PY{k+kn}{import} \PY{n}{LambdaWithMask}
        \PY{k+kn}{from} \PY{n+nn}{kraino.core.keras\PYZus{}extensions} \PY{k+kn}{import} \PY{n}{time\PYZus{}distributed\PYZus{}masked\PYZus{}ave}
        
        \PY{c+c1}{\PYZsh{} This model inherits from AbstractSingleAnswer, }
        \PY{c+c1}{\PYZsh{} and so it produces single answer words}
        \PY{c+c1}{\PYZsh{} To use multiple answer words, }
        \PY{c+c1}{\PYZsh{} you need to inherit from AbstractSequentialMultiplewordAnswer}
        \PY{k}{class} \PY{n+nc}{BlindBOW}\PY{p}{(}\PY{n}{AbstractSequentialModel}\PY{p}{,} \PY{n}{AbstractSingleAnswer}\PY{p}{)}\PY{p}{:}
            \PY{l+s+sd}{\PYZdq{}\PYZdq{}\PYZdq{}}
        \PY{l+s+sd}{    BOW Language only model that produces single word answers.}
        \PY{l+s+sd}{    \PYZdq{}\PYZdq{}\PYZdq{}}
            \PY{k}{def} \PY{n+nf}{create}\PY{p}{(}\PY{n+nb+bp}{self}\PY{p}{)}\PY{p}{:}
                \PY{n+nb+bp}{self}\PY{o}{.}\PY{n}{add}\PY{p}{(}\PY{n}{Embedding}\PY{p}{(}
                        \PY{n+nb+bp}{self}\PY{o}{.}\PY{n}{\PYZus{}config}\PY{o}{.}\PY{n}{input\PYZus{}dim}\PY{p}{,} 
                        \PY{n+nb+bp}{self}\PY{o}{.}\PY{n}{\PYZus{}config}\PY{o}{.}\PY{n}{textual\PYZus{}embedding\PYZus{}dim}\PY{p}{,} 
                        \PY{n}{mask\PYZus{}zero}\PY{o}{=}\PY{n+nb+bp}{True}\PY{p}{)}\PY{p}{)}
                \PY{n+nb+bp}{self}\PY{o}{.}\PY{n}{add}\PY{p}{(}\PY{n}{LambdaWithMask}
                        \PY{p}{(}\PY{n}{time\PYZus{}distributed\PYZus{}masked\PYZus{}ave}\PY{p}{,} \PY{n}{output\PYZus{}shape}\PY{o}{=}\PY{p}{[}\PY{n+nb+bp}{self}\PY{o}{.}\PY{n}{output\PYZus{}shape}\PY{p}{[}\PY{l+m+mi}{2}\PY{p}{]}\PY{p}{]}\PY{p}{)}\PY{p}{)}
                \PY{n+nb+bp}{self}\PY{o}{.}\PY{n}{add}\PY{p}{(}\PY{n}{DropMask}\PY{p}{(}\PY{p}{)}\PY{p}{)}
                \PY{n+nb+bp}{self}\PY{o}{.}\PY{n}{add}\PY{p}{(}\PY{n}{Dropout}\PY{p}{(}\PY{l+m+mf}{0.5}\PY{p}{)}\PY{p}{)}
                \PY{n+nb+bp}{self}\PY{o}{.}\PY{n}{add}\PY{p}{(}\PY{n}{Dense}\PY{p}{(}\PY{n+nb+bp}{self}\PY{o}{.}\PY{n}{\PYZus{}config}\PY{o}{.}\PY{n}{output\PYZus{}dim}\PY{p}{)}\PY{p}{)}
                \PY{n+nb+bp}{self}\PY{o}{.}\PY{n}{add}\PY{p}{(}\PY{n}{Activation}\PY{p}{(}\PY{l+s+s1}{\PYZsq{}}\PY{l+s+s1}{softmax}\PY{l+s+s1}{\PYZsq{}}\PY{p}{)}\PY{p}{)}
                
\end{Verbatim}

    \begin{Verbatim}[commandchars=\\\{\}]
{\color{incolor}In [{\color{incolor} }]:} \PY{n}{model\PYZus{}config} \PY{o}{=} \PY{n}{Config}\PY{p}{(}
            \PY{n}{textual\PYZus{}embedding\PYZus{}dim}\PY{o}{=}\PY{l+m+mi}{500}\PY{p}{,}
            \PY{n}{input\PYZus{}dim}\PY{o}{=}\PY{n+nb}{len}\PY{p}{(}\PY{n}{word2index\PYZus{}x}\PY{o}{.}\PY{n}{keys}\PY{p}{(}\PY{p}{)}\PY{p}{)}\PY{p}{,}
            \PY{n}{output\PYZus{}dim}\PY{o}{=}\PY{n+nb}{len}\PY{p}{(}\PY{n}{word2index\PYZus{}y}\PY{o}{.}\PY{n}{keys}\PY{p}{(}\PY{p}{)}\PY{p}{)}\PY{p}{)}
        \PY{n}{model} \PY{o}{=} \PY{n}{BlindBOW}\PY{p}{(}\PY{n}{model\PYZus{}config}\PY{p}{)}
        \PY{n}{model}\PY{o}{.}\PY{n}{create}\PY{p}{(}\PY{p}{)}
        
        \PY{n}{model}\PY{o}{.}\PY{n}{compile}\PY{p}{(}
            \PY{n}{loss}\PY{o}{=}\PY{l+s+s1}{\PYZsq{}}\PY{l+s+s1}{categorical\PYZus{}crossentropy}\PY{l+s+s1}{\PYZsq{}}\PY{p}{,} 
            \PY{n}{optimizer}\PY{o}{=}\PY{l+s+s1}{\PYZsq{}}\PY{l+s+s1}{adam}\PY{l+s+s1}{\PYZsq{}}\PY{p}{)}
        \PY{n}{text\PYZus{}bow\PYZus{}model} \PY{o}{=} \PY{n}{model}
\end{Verbatim}

    \begin{Verbatim}[commandchars=\\\{\}]
{\color{incolor}In [{\color{incolor} }]:} \PY{c+c1}{\PYZsh{}== Model training}
        \PY{n}{text\PYZus{}bow\PYZus{}model}\PY{o}{.}\PY{n}{fit}\PY{p}{(}
            \PY{n}{train\PYZus{}x}\PY{p}{,} 
            \PY{n}{train\PYZus{}y}\PY{p}{,}
            \PY{n}{batch\PYZus{}size}\PY{o}{=}\PY{l+m+mi}{512}\PY{p}{,}
            \PY{n}{nb\PYZus{}epoch}\PY{o}{=}\PY{l+m+mi}{40}\PY{p}{,}
            \PY{n}{validation\PYZus{}split}\PY{o}{=}\PY{l+m+mf}{0.1}\PY{p}{,}
            \PY{n}{show\PYZus{}accuracy}\PY{o}{=}\PY{n+nb+bp}{True}\PY{p}{)}
\end{Verbatim}

    \subsubsection{Recurrent Neural Network}\label{recurrent-neural-network}

\begin{figure}[htbp]
\centering
\includegraphics[width=0.75\linewidth]{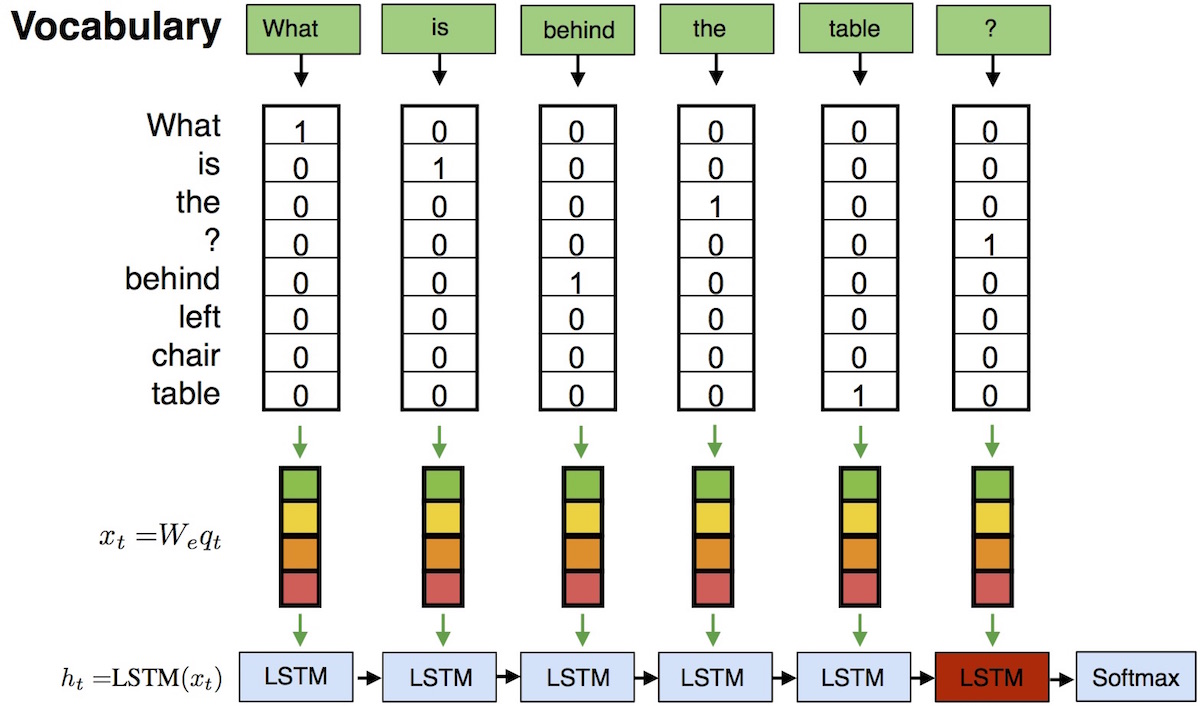}
\caption{Recurrent Neural Network.}
\label{tutorial:rnn}
\end{figure}

    Although BOW is working pretty well, there is still something very
disturbing about this approach. Consider the following question: `what is on the right side of the black telephone and on the left side of the red chair ?'
    If we swap `chair' with `telephone' in the question, we would get a
different meaning. Recurrent Neural Networks (RNNs) have
been developed to mitigate this issue by directly processing  time
series. As \figref{tutorial:rnn} illustrates, the (temporarily) first word
embedding is given to an RNN unit. The RNN unit next processes such an
embedding and outputs to the second RNN unit. This unit takes both the
output of the first RNN unit and the 2nd word embedding as inputs, and
outputs some algebraic combination of both inputs. And so on. The last
recurrent unit builds the representation of the whole sequence. Its
output is next given to Softmax for the classification. One among the
challenges that such approaches have to deal with is maintaining long-term
dependencies. Roughly speaking, as new inputs are coming in the following steps  it is getting
easier to `forget' information from the beginning (the first temporal step).
LSTM \citep{hochreiter97nc} and GRU \citep{cho2014learning} are two
particularly popular Recurrent Neural Networks that can preserve such
longer dependencies to some extent\footnote{\url{http://karpathy.github.io/2015/05/21/rnn-effectiveness/}}.

\noindent Let us create a Recurrent Neural Network in the following.

    \begin{Verbatim}[commandchars=\\\{\}]
{\color{incolor}In [{\color{incolor} }]:} \PY{c+c1}{\PYZsh{}== Model definition}
        
        \PY{c+c1}{\PYZsh{} First we define a model using keras/kraino}
        \PY{k+kn}{from} \PY{n+nn}{keras.layers.core} \PY{k+kn}{import} \PY{n}{Activation}
        \PY{k+kn}{from} \PY{n+nn}{keras.layers.core} \PY{k+kn}{import} \PY{n}{Dense}
        \PY{k+kn}{from} \PY{n+nn}{keras.layers.core} \PY{k+kn}{import} \PY{n}{Dropout}
        \PY{k+kn}{from} \PY{n+nn}{keras.layers.embeddings} \PY{k+kn}{import} \PY{n}{Embedding}
        \PY{k+kn}{from} \PY{n+nn}{keras.layers.recurrent} \PY{k+kn}{import} \PY{n}{GRU}
        \PY{k+kn}{from} \PY{n+nn}{keras.layers.recurrent} \PY{k+kn}{import} \PY{n}{LSTM}
        
        \PY{k+kn}{from} \PY{n+nn}{kraino.core.model\PYZus{}zoo} \PY{k+kn}{import} \PY{n}{AbstractSequentialModel}
        \PY{k+kn}{from} \PY{n+nn}{kraino.core.model\PYZus{}zoo} \PY{k+kn}{import} \PY{n}{AbstractSingleAnswer}
        \PY{k+kn}{from} \PY{n+nn}{kraino.core.model\PYZus{}zoo} \PY{k+kn}{import} \PY{n}{AbstractSequentialMultiplewordAnswer}
        \PY{k+kn}{from} \PY{n+nn}{kraino.core.model\PYZus{}zoo} \PY{k+kn}{import} \PY{n}{Config}
        \PY{k+kn}{from} \PY{n+nn}{kraino.core.keras\PYZus{}extensions} \PY{k+kn}{import} \PY{n}{DropMask}
        \PY{k+kn}{from} \PY{n+nn}{kraino.core.keras\PYZus{}extensions} \PY{k+kn}{import} \PY{n}{LambdaWithMask}
        \PY{k+kn}{from} \PY{n+nn}{kraino.core.keras\PYZus{}extensions} \PY{k+kn}{import} \PY{n}{time\PYZus{}distributed\PYZus{}masked\PYZus{}ave}
        
        \PY{c+c1}{\PYZsh{} This model inherits from AbstractSingleAnswer, }
        \PY{c+c1}{\PYZsh{} and so it produces single answer words}
        \PY{c+c1}{\PYZsh{} To use multiple answer words, }
        \PY{c+c1}{\PYZsh{} you need to inherit from AbstractSequentialMultiplewordAnswer}
        \PY{k}{class} \PY{n+nc}{BlindRNN}\PY{p}{(}\PY{n}{AbstractSequentialModel}\PY{p}{,} \PY{n}{AbstractSingleAnswer}\PY{p}{)}\PY{p}{:}
            \PY{l+s+sd}{\PYZdq{}\PYZdq{}\PYZdq{}}
        \PY{l+s+sd}{    RNN Language only model that produces single word answers.}
        \PY{l+s+sd}{    \PYZdq{}\PYZdq{}\PYZdq{}}
            \PY{k}{def} \PY{n+nf}{create}\PY{p}{(}\PY{n+nb+bp}{self}\PY{p}{)}\PY{p}{:}
                \PY{n+nb+bp}{self}\PY{o}{.}\PY{n}{add}\PY{p}{(}\PY{n}{Embedding}\PY{p}{(}
                        \PY{n+nb+bp}{self}\PY{o}{.}\PY{n}{\PYZus{}config}\PY{o}{.}\PY{n}{input\PYZus{}dim}\PY{p}{,} 
                        \PY{n+nb+bp}{self}\PY{o}{.}\PY{n}{\PYZus{}config}\PY{o}{.}\PY{n}{textual\PYZus{}embedding\PYZus{}dim}\PY{p}{,} 
                        \PY{n}{mask\PYZus{}zero}\PY{o}{=}\PY{n+nb+bp}{True}\PY{p}{)}\PY{p}{)}
                \PY{c+c1}{\PYZsh{}TODO: Replace averaging with RNN (you can choose between LSTM and GRU)}
                \PY{n+nb+bp}{self}\PY{o}{.}\PY{n}{add}\PY{p}{(}\PY{n}{GRU}\PY{p}{(}\PY{n+nb+bp}{self}\PY{o}{.}\PY{n}{\PYZus{}config}\PY{o}{.}\PY{n}{hidden\PYZus{}state\PYZus{}dim}\PY{p}{,} 
                              \PY{n}{return\PYZus{}sequences}\PY{o}{=}\PY{n+nb+bp}{False}\PY{p}{)}\PY{p}{)}
                \PY{n+nb+bp}{self}\PY{o}{.}\PY{n}{add}\PY{p}{(}\PY{n}{Dropout}\PY{p}{(}\PY{l+m+mf}{0.5}\PY{p}{)}\PY{p}{)}
                \PY{n+nb+bp}{self}\PY{o}{.}\PY{n}{add}\PY{p}{(}\PY{n}{Dense}\PY{p}{(}\PY{n+nb+bp}{self}\PY{o}{.}\PY{n}{\PYZus{}config}\PY{o}{.}\PY{n}{output\PYZus{}dim}\PY{p}{)}\PY{p}{)}
                \PY{n+nb+bp}{self}\PY{o}{.}\PY{n}{add}\PY{p}{(}\PY{n}{Activation}\PY{p}{(}\PY{l+s+s1}{\PYZsq{}}\PY{l+s+s1}{softmax}\PY{l+s+s1}{\PYZsq{}}\PY{p}{)}\PY{p}{)}
                
\end{Verbatim}

    \begin{Verbatim}[commandchars=\\\{\}]
{\color{incolor}In [{\color{incolor} }]:} \PY{n}{model\PYZus{}config} \PY{o}{=} \PY{n}{Config}\PY{p}{(}
            \PY{n}{textual\PYZus{}embedding\PYZus{}dim}\PY{o}{=}\PY{l+m+mi}{500}\PY{p}{,}
            \PY{n}{hidden\PYZus{}state\PYZus{}dim}\PY{o}{=}\PY{l+m+mi}{500}\PY{p}{,}
            \PY{n}{input\PYZus{}dim}\PY{o}{=}\PY{n+nb}{len}\PY{p}{(}\PY{n}{word2index\PYZus{}x}\PY{o}{.}\PY{n}{keys}\PY{p}{(}\PY{p}{)}\PY{p}{)}\PY{p}{,}
            \PY{n}{output\PYZus{}dim}\PY{o}{=}\PY{n+nb}{len}\PY{p}{(}\PY{n}{word2index\PYZus{}y}\PY{o}{.}\PY{n}{keys}\PY{p}{(}\PY{p}{)}\PY{p}{)}\PY{p}{)}
        \PY{n}{model} \PY{o}{=} \PY{n}{BlindRNN}\PY{p}{(}\PY{n}{model\PYZus{}config}\PY{p}{)}
        \PY{n}{model}\PY{o}{.}\PY{n}{create}\PY{p}{(}\PY{p}{)}
        \PY{n}{model}\PY{o}{.}\PY{n}{compile}\PY{p}{(}
            \PY{n}{loss}\PY{o}{=}\PY{l+s+s1}{\PYZsq{}}\PY{l+s+s1}{categorical\PYZus{}crossentropy}\PY{l+s+s1}{\PYZsq{}}\PY{p}{,} 
            \PY{n}{optimizer}\PY{o}{=}\PY{l+s+s1}{\PYZsq{}}\PY{l+s+s1}{adam}\PY{l+s+s1}{\PYZsq{}}\PY{p}{)}
        \PY{n}{text\PYZus{}rnn\PYZus{}model} \PY{o}{=} \PY{n}{model}
\end{Verbatim}

    \begin{Verbatim}[commandchars=\\\{\}]
{\color{incolor}In [{\color{incolor} }]:} \PY{c+c1}{\PYZsh{}== Model training}
        \PY{n}{text\PYZus{}rnn\PYZus{}model}\PY{o}{.}\PY{n}{fit}\PY{p}{(}
            \PY{n}{train\PYZus{}x}\PY{p}{,} 
            \PY{n}{train\PYZus{}y}\PY{p}{,}
            \PY{n}{batch\PYZus{}size}\PY{o}{=}\PY{l+m+mi}{512}\PY{p}{,}
            \PY{n}{nb\PYZus{}epoch}\PY{o}{=}\PY{l+m+mi}{40}\PY{p}{,}
            \PY{n}{validation\PYZus{}split}\PY{o}{=}\PY{l+m+mf}{0.1}\PY{p}{,}
            \PY{n}{show\PYZus{}accuracy}\PY{o}{=}\PY{n+nb+bp}{True}\PY{p}{)}
\end{Verbatim}

The curious reader is encouraged to experiment with the language-only models.
For instance, to see the influence of particular modules to the overall performance,
the reader can do the following exercise.

\begin{verbatim}
	Change the number of hidden states.
	Change the number of epochs used to train a model.
	Modify models by using more RNN layers, or deeper classifiers.
\end{verbatim}

\paragraph{Summary}
RNN models, as opposite to BOW, consider order of the words in the
question. Moreover, a substantial number of questions can be
answered without any access to images. This can be explained as models
learn some specific dataset statistics, some of them can be interpreted as
common sense knowledge.

\section{Evaluation Measures}\label{evaluation-measures}

    To be able to monitor a progress on a task, we need to find ways to
evaluate architectures on the task. Otherwise, we would not know how to judge architectures, or even worse, we would not even know what the goal is.
Moreover, we should also aim at automatic evaluation measures, otherwise
reproducibility is questionable, and the evaluation costs are high.

\subsubsection{Ambiguities}\label{ambiguities}
Although an early work on the Visual Turing Test argues for keeping the answer words from a fixed vocabulary
in order to keep an evaluation simpler \citep{malinowski14nips, malinowski14visualturing, malinowski2015hard}, it is still difficult to automatically evaluate architectures due to ambiguities that occur in the answers.
We have ambiguities in naming objects, sometimes due
to synonyms, but sometimes due to fuzziness. For instance, is `chair' ==
`armchair' or `chair' != `armchair' or something in between? Such
semantic boundaries become even more fuzzy when we increase the number
of categories. We could easily find a mutually exclusive set of 10
different categories, but what if there are 1000 categories, or 10000
categories? Arguably, we cannot think in terms of an equivalence class
anymore, but rather in terms of similarities. That is `chair' is
semantically more similar to `armchair', than to `horse'. This simple
example shows the main drawback of a traditional, binary evaluation
measure which is Accuracy. This metric scores 1 if the names are the same and 0
otherwise. So that Acc(`chair', `armchair') == Acc(`chair', `horse'). 
We call these ambiguities, word-level ambiguities, but there are other
ambiguities that are arguably more difficult to handle. For instance,
the same question can be phrased in multiple other ways. The language of
spatial relations is also ambiguous. Language tends to
be also rather vague - we sometimes skip details and resort to common
sense. Some ambiguities are rooted in a culture. 
To address world-level ambiguities, \citet{malinowski14nips} propose WUPS.
To address ambiguities caused by various interpretations of an image or a question, \citet{malinowski2015ask} propose Consensus measures.
For the sake of simplicity, in this tutorial, we only use WUPS.
On the other hand, arguably, it is easier to evaluate architectures on
DAQUAR than on  Image Captioning datasets. The former restricts the
output space to $N$ categories, while it still requires a holistic  comprehension.
Let us remind that \figref{tutorial:challenges} shows a few ambiguities that exists in DAQUAR.

    \subsubsection{Wu-Palmer Similarity}\label{wu-palmer-similarity}

\begin{figure}[htbp]
\centering
\includegraphics[width=0.35\linewidth]{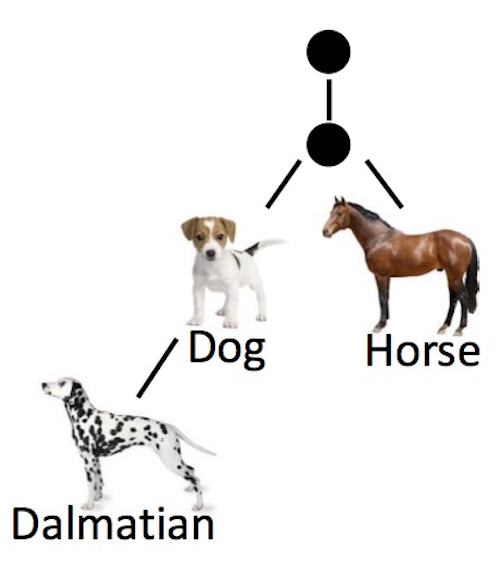}
\caption{A toy-sized ontology.}
\label{tutorial:small_ontology}
\end{figure}

    Given an ontology a Wu-Palmer Similarity between two words (or broader
concepts) is a soft measure defined as
\[WuP(a,b) := \frac{lca(a,b)}{depth(a) + depth(b)}\] where $lca(a,b)$ is
the least common ancestor of $a$ and $b$, and $depth(a)$ is depth of $a$
in the ontology. \figref{tutorial:small_ontology} shows a toy-sized ontology.
The curious reader can, based on \figref{tutorial:small_ontology}, address the following questions.

\begin{verbatim}
	What is WuP(Dog, Horse) and WuP(Dog, Dalmatian) according to the toy-sized ontology? 
	Can you  calculate Acc(Dog, Horse) and Acc(Dog, Dalmatian)?
\end{verbatim}

\subsubsection{WUPS}\label{wups}
    Wu-Palmer Similarity depends on the choice of ontology. One popular, large ontology
is WordNet \citep{miller1995wordnet,fellbaum1999wordnet}. Although Wu-Palmer
Similarity may work on shallow ontologies, we are rather interested in
ontologies with hundreds or even thousands of categories. In indoor
scenarios, it turns out that many indoor `things' share similar levels in
the ontology, and hence Wu-Palmer Similarities are very small between two entities.
The following code exemplifies the issue.
    \begin{Verbatim}[commandchars=\\\{\}]
{\color{incolor}In [{\color{incolor} }]:} \PY{k+kn}{from} \PY{n+nn}{nltk.corpus} \PY{k+kn}{import} \PY{n}{wordnet} \PY{k}{as} \PY{n}{wn}
        \PY{n}{armchair\PYZus{}synset} \PY{o}{=} \PY{n}{wn}\PY{o}{.}\PY{n}{synset}\PY{p}{(}\PY{l+s+s1}{\PYZsq{}}\PY{l+s+s1}{armchair.n.01}\PY{l+s+s1}{\PYZsq{}}\PY{p}{)}
        \PY{n}{chair\PYZus{}synset} \PY{o}{=} \PY{n}{wn}\PY{o}{.}\PY{n}{synset}\PY{p}{(}\PY{l+s+s1}{\PYZsq{}}\PY{l+s+s1}{chair.n.01}\PY{l+s+s1}{\PYZsq{}}\PY{p}{)}
        \PY{n}{wardrobe\PYZus{}synset} \PY{o}{=} \PY{n}{wn}\PY{o}{.}\PY{n}{synset}\PY{p}{(}\PY{l+s+s1}{\PYZsq{}}\PY{l+s+s1}{wardrobe.n.01}\PY{l+s+s1}{\PYZsq{}}\PY{p}{)}
        
        \PY{k}{print}\PY{p}{(}\PY{n}{armchair\PYZus{}synset}\PY{o}{.}\PY{n}{wup\PYZus{}similarity}\PY{p}{(}\PY{n}{armchair\PYZus{}synset}\PY{p}{)}\PY{p}{)}
        \PY{k}{print}\PY{p}{(}\PY{n}{armchair\PYZus{}synset}\PY{o}{.}\PY{n}{wup\PYZus{}similarity}\PY{p}{(}\PY{n}{chair\PYZus{}synset}\PY{p}{)}\PY{p}{)}
        \PY{k}{print}\PY{p}{(}\PY{n}{armchair\PYZus{}synset}\PY{o}{.}\PY{n}{wup\PYZus{}similarity}\PY{p}{(}\PY{n}{wardrobe\PYZus{}synset}\PY{p}{)}\PY{p}{)}
        \PY{n}{wn}\PY{o}{.}\PY{n}{synset}\PY{p}{(}\PY{l+s+s1}{\PYZsq{}}\PY{l+s+s1}{chair.n.01}\PY{l+s+s1}{\PYZsq{}}\PY{p}{)}\PY{o}{.}\PY{n}{wup\PYZus{}similarity}\PY{p}{(}\PY{n}{wn}\PY{o}{.}\PY{n}{synset}\PY{p}{(}\PY{l+s+s1}{\PYZsq{}}\PY{l+s+s1}{person.n.01}\PY{l+s+s1}{\PYZsq{}}\PY{p}{)}\PY{p}{)}
\end{Verbatim}

\noindent As we can see that `armchair' and `wardrobe' are surprisingly
close to each other. It is because, for large ontologies, all the indoor `things' are
semantically `indoor things'.
    This issue has motivated us to define thresholded Wu-Palmer Similarity
Score, defined as follows \[
\begin{array}{rl}
WuP(a,b) & \text{if}\; WuP(a,b) \ge \tau \\
0.1 \cdot WuP(a,b) & \text{otherwise}
\end{array}
\] where $\tau$ is a hand-chosen threshold. Empirically, we found that
$\tau=0.9$ works fine on DAQUAR \citep{malinowski14nips}.
Moreover, since DAQUAR has answers as sets of answer words, so that
`knife,fork' == `fork,knife', we have extended the above measure to work
with the sets. We call it Wu-Palmer Set score, or shortly WUPS.

    A detailed exposition of WUPS is beyond this tutorial, but a curious
reader is encouraged to read the `Performance Measure' paragraph in \citet{malinowski14nips}.
Note that the
measure in \citet{malinowski14nips} is
defined broader, and essentially it abstracts  away from any particular
similarities such as Wu-Palmer Similarity, or an ontology. WUPS at 0.9 is WUPS with
threshold $\tau=0.9$.
It is worth noting, that a practical implementation of WUPS  needs to deal with synsets. Thus it is
recommended to download the script from \url{http://datasets.d2.mpi-inf.mpg.de/mateusz14visual-turing/calculate_wups.py}
or re-implement it with caution. 

    \subsubsection{Consensus}\label{consensus}
The consensus measure handles ambiguities that are caused by various interpretations of a question or an image.
    In this tutorial, we do not cover this measure. A curious
reader is encouraged to read the `Human Consensus' in \citet{malinowski2015ask}.

\subsubsection{A few caveats}\label{a-few-caveats}
We present a few caveats when using WUPS. These can be especially useful if one wants to adapt WUPS to other datasets.

\paragraph{Lack of coverage} Since WUPS is based on an ontology, not always
it recognizes words. For instance `garbage bin' is missing, but `garbage
can' is perfectly fine. You can check it by yourself, either with the
source code provided above, or by using an online script\footnote{\url{http://wordnetweb.princeton.edu/perl/webwn}}.

\paragraph{Synsets} The execution of the following code

\begin{Shaded}
\begin{Highlighting}[]
\NormalTok{wn.synsets(}\StringTok{'chair'}\NormalTok{)}
\end{Highlighting}
\end{Shaded}

\noindent produces a list with many elements. These elements are semantically equivalent\footnote{\url{https://en.wikipedia.org/wiki/Synonym_ring}}.

For instance the following definition of `chair'
\begin{Shaded}
\begin{Highlighting}[]
\NormalTok{wn.synset(}\StringTok{'chair.n.03'}\NormalTok{).definition()}
\end{Highlighting}
\end{Shaded}

\noindent indicates a person (e.g. a chairman). Indeed, the following gives quite high value

\begin{Shaded}
\begin{Highlighting}[]
\NormalTok{wn.synset(}\StringTok{'chair.n.03'}\NormalTok{).wup_similarity(wn.synset(}\StringTok{'person.n.01'}\NormalTok{))}
\end{Highlighting}
\end{Shaded}

\noindent however the following one has a more preferred, much lower value

\begin{Shaded}
\begin{Highlighting}[]
\NormalTok{wn.synset(}\StringTok{'chair.n.01'}\NormalTok{).wup_similarity(wn.synset(}\StringTok{'person.n.01'}\NormalTok{))}
\end{Highlighting}
\end{Shaded}

\noindent How to deal with such a problem? In DAQUAR we take an optimistic perspective and
always consider the highest similarity score. This works with WUPS 0.9
and a restricted indoor domain with a vocabulary based only on the training
set. To sum up, this issue should be taken with a caution whenever WUPS is adapted to other domains.

\paragraph{Ontology} Since WUPS is based on an ontology, specifically on
WordNet, it may give different scores on different ontologies, or even
on different versions of the same ontology.

\paragraph{Threshold} A good threshold $\tau$ is dataset dependent. In our
case $\tau = 0.9$ seems to work well, while $\tau = 0.0$ is too
forgivable and is rather reported due to the `historical' reasons.
However, following our papers, you should still consider to report plain
set-based accuracy scores (so that Acc(`knife,'fork',`fork,knife')==1;
it can be computed by our script\footnote{\url{http://datasets.d2.mpi-inf.mpg.de/mateusz14visual-turing/calculate_wups.py}} using the argument -1 to WUPS.

    \subsubsection{Summary}\label{summary}

    WUPS is an evaluation measure that works with sets and word-level
ambiguities. Arguably, WUPS at 0.9 is the most practical measure.

\section{New Predictions}\label{new-predictions}
Once the training of our models is over, we can evaluate their performance on a previously unknown test set.
In the following, we show how to make predictions using  the already discussed blind models.
    \subsubsection{Predictions - BOW}\label{predictions---bow}

\noindent We start from encoding textual input into one-hot vector representations.
    \begin{Verbatim}[commandchars=\\\{\}]
{\color{incolor}In [{\color{incolor} }]:} \PY{n}{test\PYZus{}text\PYZus{}representation} \PY{o}{=} \PY{n}{dp}\PY{p}{[}\PY{l+s+s1}{\PYZsq{}}\PY{l+s+s1}{text}\PY{l+s+s1}{\PYZsq{}}\PY{p}{]}\PY{p}{(}\PY{n}{train\PYZus{}or\PYZus{}test}\PY{o}{=}\PY{l+s+s1}{\PYZsq{}}\PY{l+s+s1}{test}\PY{l+s+s1}{\PYZsq{}}\PY{p}{)}
        \PY{n}{test\PYZus{}raw\PYZus{}x} \PY{o}{=} \PY{n}{test\PYZus{}text\PYZus{}representation}\PY{p}{[}\PY{l+s+s1}{\PYZsq{}}\PY{l+s+s1}{x}\PY{l+s+s1}{\PYZsq{}}\PY{p}{]}
        \PY{n}{test\PYZus{}one\PYZus{}hot\PYZus{}x} \PY{o}{=} \PY{n}{encode\PYZus{}questions\PYZus{}index}\PY{p}{(}\PY{n}{test\PYZus{}raw\PYZus{}x}\PY{p}{,} \PY{n}{word2index\PYZus{}x}\PY{p}{)}
        \PY{n}{test\PYZus{}x} \PY{o}{=} \PY{n}{sequence}\PY{o}{.}\PY{n}{pad\PYZus{}sequences}\PY{p}{(}\PY{n}{test\PYZus{}one\PYZus{}hot\PYZus{}x}\PY{p}{,} \PY{n}{maxlen}\PY{o}{=}\PY{n}{MAXLEN}\PY{p}{)}
\end{Verbatim}

\noindent   Given encoded test questions, we use the maximum likelihood principle to
withdraw answers.

    \begin{Verbatim}[commandchars=\\\{\}]
{\color{incolor}In [{\color{incolor} }]:} \PY{k+kn}{from} \PY{n+nn}{numpy} \PY{k+kn}{import} \PY{n}{argmax}
        \PY{c+c1}{\PYZsh{} predict the probabilities for every word}
        \PY{n}{predictions\PYZus{}scores} \PY{o}{=} \PY{n}{text\PYZus{}bow\PYZus{}model}\PY{o}{.}\PY{n}{predict}\PY{p}{(}\PY{p}{[}\PY{n}{test\PYZus{}x}\PY{p}{]}\PY{p}{)}
        \PY{k}{print}\PY{p}{(}\PY{n}{predictions\PYZus{}scores}\PY{o}{.}\PY{n}{shape}\PY{p}{)}
        \PY{c+c1}{\PYZsh{} follow the maximum likelihood principle, and get the best indices to vocabulary}
        \PY{n}{predictions\PYZus{}best} \PY{o}{=} \PY{n}{argmax}\PY{p}{(}\PY{n}{predictions\PYZus{}scores}\PY{p}{,} \PY{n}{axis}\PY{o}{=}\PY{o}{\PYZhy{}}\PY{l+m+mi}{1}\PY{p}{)}
        \PY{k}{print}\PY{p}{(}\PY{n}{predictions\PYZus{}best}\PY{o}{.}\PY{n}{shape}\PY{p}{)}
        \PY{c+c1}{\PYZsh{} decode the predicted indices into word answers}
        \PY{n}{predictions\PYZus{}answers} \PY{o}{=} \PY{p}{[}\PY{n}{index2word\PYZus{}y}\PY{p}{[}\PY{n}{x}\PY{p}{]} \PY{k}{for} \PY{n}{x} \PY{o+ow}{in} \PY{n}{predictions\PYZus{}best}\PY{p}{]}
        \PY{k}{print}\PY{p}{(}\PY{n+nb}{len}\PY{p}{(}\PY{n}{predictions\PYZus{}answers}\PY{p}{)}\PY{p}{)}
\end{Verbatim}

\noindent Now, we evaluate the answers using WUPS. 

    \begin{Verbatim}[commandchars=\\\{\}]
{\color{incolor}In [{\color{incolor} }]:} \PY{k+kn}{from} \PY{n+nn}{kraino.utils} \PY{k+kn}{import} \PY{n}{print\PYZus{}metrics}
        \PY{n}{test\PYZus{}raw\PYZus{}y} \PY{o}{=} \PY{n}{test\PYZus{}text\PYZus{}representation}\PY{p}{[}\PY{l+s+s1}{\PYZsq{}}\PY{l+s+s1}{y}\PY{l+s+s1}{\PYZsq{}}\PY{p}{]}
        \PY{n}{\PYZus{}} \PY{o}{=} \PY{n}{print\PYZus{}metrics}\PY{o}{.}\PY{n}{select}\PY{p}{[}\PY{l+s+s1}{\PYZsq{}}\PY{l+s+s1}{wups}\PY{l+s+s1}{\PYZsq{}}\PY{p}{]}\PY{p}{(}
                \PY{n}{gt\PYZus{}list}\PY{o}{=}\PY{n}{test\PYZus{}raw\PYZus{}y}\PY{p}{,}
                \PY{n}{pred\PYZus{}list}\PY{o}{=}\PY{n}{predictions\PYZus{}answers}\PY{p}{,}
                \PY{n}{verbose}\PY{o}{=}\PY{l+m+mi}{1}\PY{p}{,}
                \PY{n}{extra\PYZus{}vars}\PY{o}{=}\PY{n+nb+bp}{None}\PY{p}{)}
\end{Verbatim}

\noindent Let us see the predictions.

    \begin{Verbatim}[commandchars=\\\{\}]
{\color{incolor}In [{\color{incolor} }]:} \PY{k+kn}{from} \PY{n+nn}{numpy} \PY{k+kn}{import} \PY{n}{random}
        \PY{n}{test\PYZus{}image\PYZus{}name\PYZus{}list} \PY{o}{=} \PY{n}{test\PYZus{}text\PYZus{}representation}\PY{p}{[}\PY{l+s+s1}{\PYZsq{}}\PY{l+s+s1}{img\PYZus{}name}\PY{l+s+s1}{\PYZsq{}}\PY{p}{]}
        \PY{n}{indices\PYZus{}to\PYZus{}see} \PY{o}{=} \PY{n}{random}\PY{o}{.}\PY{n}{randint}\PY{p}{(}\PY{n}{low}\PY{o}{=}\PY{l+m+mi}{0}\PY{p}{,} \PY{n}{high}\PY{o}{=}\PY{n+nb}{len}\PY{p}{(}\PY{n}{test\PYZus{}image\PYZus{}name\PYZus{}list}\PY{p}{)}\PY{p}{,} \PY{n}{size}\PY{o}{=}\PY{l+m+mi}{5}\PY{p}{)}
        \PY{k}{for} \PY{n}{index\PYZus{}now} \PY{o+ow}{in} \PY{n}{indices\PYZus{}to\PYZus{}see}\PY{p}{:}
            \PY{k}{print}\PY{p}{(}\PY{n}{test\PYZus{}raw\PYZus{}x}\PY{p}{[}\PY{n}{index\PYZus{}now}\PY{p}{]}\PY{p}{,} \PY{n}{predictions\PYZus{}answers}\PY{p}{[}\PY{n}{index\PYZus{}now}\PY{p}{]}\PY{p}{)}
\end{Verbatim}
Without looking at images, a curious reader may attempt to answer the following questions.
    \begin{verbatim}
Do you agree with the answers given above? What are your guesses?
Of course, neither you nor the model have seen any images so far.
\end{verbatim}

\noindent However, what happens if the reader can actually see the images?

\begin{verbatim}
Execute the code below.
Do your answers change after seeing the images?
\end{verbatim}

    \begin{Verbatim}[commandchars=\\\{\}]
{\color{incolor}In [{\color{incolor}1}]:} \PY{k+kn}{from} \PY{n+nn}{matplotlib.pyplot} \PY{k+kn}{import} \PY{n}{axis}
        \PY{k+kn}{from} \PY{n+nn}{matplotlib.pyplot} \PY{k+kn}{import} \PY{n}{figure}
        \PY{k+kn}{from} \PY{n+nn}{matplotlib.pyplot} \PY{k+kn}{import} \PY{n}{imshow}
        
        \PY{k+kn}{import} \PY{n+nn}{numpy} \PY{k+kn}{as} \PY{n+nn}{np}
        \PY{k+kn}{from} \PY{n+nn}{PIL} \PY{k+kn}{import} \PY{n}{Image}
        
        \PY{o}{\PYZpc{}}\PY{k}{matplotlib} inline
        \PY{k}{for} \PY{n}{index\PYZus{}now} \PY{o+ow}{in} \PY{n}{indices\PYZus{}to\PYZus{}see}\PY{p}{:}
            \PY{n}{image\PYZus{}name\PYZus{}now} \PY{o}{=} \PY{n}{test\PYZus{}image\PYZus{}name\PYZus{}list}\PY{p}{[}\PY{n}{index\PYZus{}now}\PY{p}{]}
            \PY{n}{pil\PYZus{}im} \PY{o}{=} \PY{n}{Image}\PY{o}{.}\PY{n}{open}\PY{p}{(}\PY{l+s+s1}{\PYZsq{}}\PY{l+s+s1}{data/daquar/images/\PYZob{}0\PYZcb{}.png}\PY{l+s+s1}{\PYZsq{}}\PY{o}{.}\PY{n}{format}\PY{p}{(}\PY{n}{image\PYZus{}name\PYZus{}now}\PY{p}{)}\PY{p}{,} \PY{l+s+s1}{\PYZsq{}}\PY{l+s+s1}{r}\PY{l+s+s1}{\PYZsq{}}\PY{p}{)}
            \PY{n}{fig} \PY{o}{=} \PY{n}{figure}\PY{p}{(}\PY{p}{)}
            \PY{n}{fig}\PY{o}{.}\PY{n}{text}\PY{p}{(}\PY{o}{.}\PY{l+m+mi}{2}\PY{p}{,}\PY{o}{.}\PY{l+m+mo}{05}\PY{p}{,}\PY{n}{test\PYZus{}raw\PYZus{}x}\PY{p}{[}\PY{n}{index\PYZus{}now}\PY{p}{]}\PY{p}{,} \PY{n}{fontsize}\PY{o}{=}\PY{l+m+mi}{14}\PY{p}{)}
            \PY{n}{axis}\PY{p}{(}\PY{l+s+s1}{\PYZsq{}}\PY{l+s+s1}{off}\PY{l+s+s1}{\PYZsq{}}\PY{p}{)}
            \PY{n}{imshow}\PY{p}{(}\PY{n}{np}\PY{o}{.}\PY{n}{asarray}\PY{p}{(}\PY{n}{pil\PYZus{}im}\PY{p}{)}\PY{p}{)}
\end{Verbatim}

\noindent  Finally, let us also see the ground truth answers by executing the following code.

    \begin{Verbatim}[commandchars=\\\{\}]
{\color{incolor}In [{\color{incolor} }]:} \PY{k}{print}\PY{p}{(}\PY{l+s+s1}{\PYZsq{}}\PY{l+s+s1}{question, prediction, ground truth answer}\PY{l+s+s1}{\PYZsq{}}\PY{p}{)}
        \PY{k}{for} \PY{n}{index\PYZus{}now} \PY{o+ow}{in} \PY{n}{indices\PYZus{}to\PYZus{}see}\PY{p}{:}
            \PY{k}{print}\PY{p}{(}\PY{n}{test\PYZus{}raw\PYZus{}x}\PY{p}{[}\PY{n}{index\PYZus{}now}\PY{p}{]}\PY{p}{,}
                 \PY{n}{predictions\PYZus{}answers}\PY{p}{[}\PY{n}{index\PYZus{}now}\PY{p}{]}\PY{p}{,} \PY{n}{test\PYZus{}raw\PYZus{}y}\PY{p}{[}\PY{n}{index\PYZus{}now}\PY{p}{]}\PY{p}{)}
\end{Verbatim}

\noindent In the code above, we have randomly taken questions, and hence  different
executions of the code may lead to different answers.

\subsubsection{Predictions - RNN}\label{predictions---rnn}

Let us do similar predictions with a Recurrent Neural Network.
This time, we use Kraino, to make the code shorter.

    \begin{Verbatim}[commandchars=\\\{\}]
{\color{incolor}In [{\color{incolor} }]:} \PY{k+kn}{from} \PY{n+nn}{kraino.core.model\PYZus{}zoo} \PY{k+kn}{import} \PY{n}{word\PYZus{}generator}
        \PY{c+c1}{\PYZsh{} we first need to add word\PYZus{}generator to \PYZus{}config }
        \PY{c+c1}{\PYZsh{} (we could have done this before, in the Config constructor)}
        \PY{c+c1}{\PYZsh{} we use maximum likelihood as a word generator}
        \PY{n}{text\PYZus{}rnn\PYZus{}model}\PY{o}{.}\PY{n}{\PYZus{}config}\PY{o}{.}\PY{n}{word\PYZus{}generator} \PY{o}{=} \PY{n}{word\PYZus{}generator}\PY{p}{[}\PY{l+s+s1}{\PYZsq{}}\PY{l+s+s1}{max\PYZus{}likelihood}\PY{l+s+s1}{\PYZsq{}}\PY{p}{]}
        \PY{n}{predictions\PYZus{}answers} \PY{o}{=} \PY{n}{text\PYZus{}rnn\PYZus{}model}\PY{o}{.}\PY{n}{decode\PYZus{}predictions}\PY{p}{(}
            \PY{n}{X}\PY{o}{=}\PY{n}{test\PYZus{}x}\PY{p}{,}
            \PY{n}{temperature}\PY{o}{=}\PY{n+nb+bp}{None}\PY{p}{,}
            \PY{n}{index2word}\PY{o}{=}\PY{n}{index2word\PYZus{}y}\PY{p}{,}
            \PY{n}{verbose}\PY{o}{=}\PY{l+m+mi}{0}\PY{p}{)}
\end{Verbatim}

    \begin{Verbatim}[commandchars=\\\{\}]
{\color{incolor}In [{\color{incolor} }]:} \PY{n}{\PYZus{}} \PY{o}{=} \PY{n}{print\PYZus{}metrics}\PY{o}{.}\PY{n}{select}\PY{p}{[}\PY{l+s+s1}{\PYZsq{}}\PY{l+s+s1}{wups}\PY{l+s+s1}{\PYZsq{}}\PY{p}{]}\PY{p}{(}
                \PY{n}{gt\PYZus{}list}\PY{o}{=}\PY{n}{test\PYZus{}raw\PYZus{}y}\PY{p}{,}
                \PY{n}{pred\PYZus{}list}\PY{o}{=}\PY{n}{predictions\PYZus{}answers}\PY{p}{,}
                \PY{n}{verbose}\PY{o}{=}\PY{l+m+mi}{1}\PY{p}{,}
                \PY{n}{extra\PYZus{}vars}\PY{o}{=}\PY{n+nb+bp}{None}\PY{p}{)}
\end{Verbatim}

\noindent A curious reader is encouraged to try the following exercise.
    \begin{verbatim}
Visualise question, predicted answers, ground truth answers as before.
Check also images.
\end{verbatim}

\section{Visual Features}\label{visual-features}
\begin{figure}[htbp]
\centering
\includegraphics[width=0.75\linewidth]{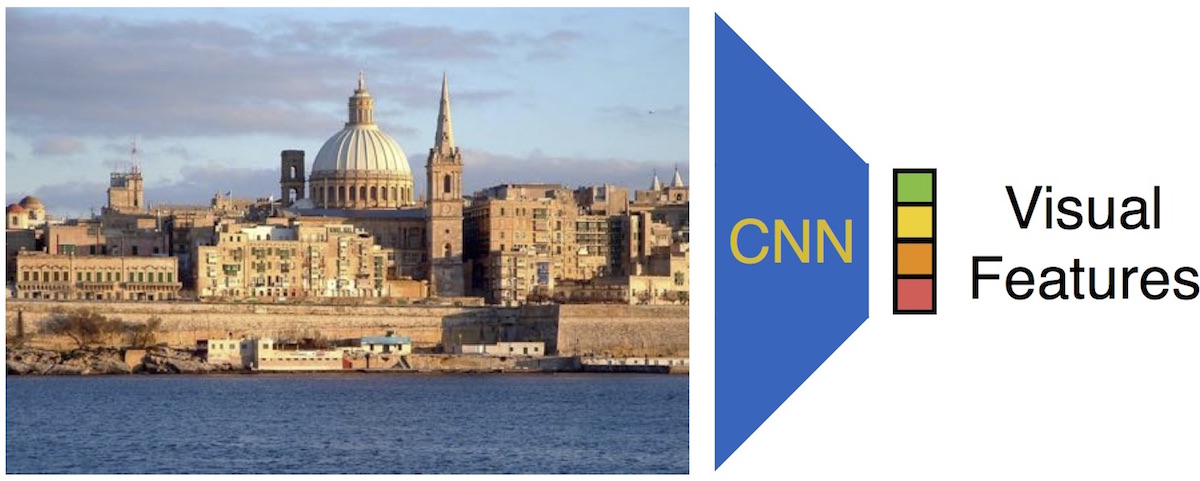}
\caption{Features extractor. Neural responses of some layer to the visual input are considered as features.}
\label{tutorial:features_extractor}
\end{figure}

All the considered so far architectures predict answers  based only on questions, even though the questions concern images.
Therefore, in this section, we also build visual features.
As shown in \figref{tutorial:features_extractor}, a quite common practice is to:
\begin{enumerate}
\item Take an already pre-trained CNN; often pre-training is done in some large-scale classification task such as ImageNet \citep{ILSVRCarxiv14}.
\item `Chop off' a CNN representation after some layer. We  use responses of
that layer as visual features.
\end{enumerate}

\noindent In this tutorial, we use features extracted from the second last
$4096$ dimensional layer of the VGG Net \citep{simonyan2014very}. We have already
extracted features in advance using Caffe \citep{jia2014caffe} - another excellent
framework for deep learning, particularly good for CNNs.

\noindent The following code gives visual features aligned with textual features.

    \begin{Verbatim}[commandchars=\\\{\}]
{\color{incolor}In [{\color{incolor} }]:} \PY{c+c1}{\PYZsh{} this contains a list of the image names of our interest; }
        \PY{c+c1}{\PYZsh{} it also makes sure that visual and textual features are aligned correspondingly}
        \PY{n}{train\PYZus{}image\PYZus{}names} \PY{o}{=} \PY{n}{train\PYZus{}text\PYZus{}representation}\PY{p}{[}\PY{l+s+s1}{\PYZsq{}}\PY{l+s+s1}{img\PYZus{}name}\PY{l+s+s1}{\PYZsq{}}\PY{p}{]}
        \PY{c+c1}{\PYZsh{} the name for visual features that we use}
        \PY{c+c1}{\PYZsh{} CNN\PYZus{}NAME=\PYZsq{}vgg\PYZus{}net\PYZsq{}}
        \PY{c+c1}{\PYZsh{} CNN\PYZus{}NAME=\PYZsq{}googlenet\PYZsq{}}
        \PY{n}{CNN\PYZus{}NAME}\PY{o}{=}\PY{l+s+s1}{\PYZsq{}}\PY{l+s+s1}{fb\PYZus{}resnet}\PY{l+s+s1}{\PYZsq{}}
        \PY{c+c1}{\PYZsh{} the layer in CNN that is used to extract features}
        \PY{c+c1}{\PYZsh{} PERCEPTION\PYZus{}LAYER=\PYZsq{}fc7\PYZsq{}}
        \PY{c+c1}{\PYZsh{} PERCEPTION\PYZus{}LAYER=\PYZsq{}pool5\PYZhy{}7x7\PYZus{}s1\PYZsq{}}
        \PY{c+c1}{\PYZsh{} PERCEPTION\PYZus{}LAYER=\PYZsq{}res5c\PYZhy{}152\PYZsq{}}
        \PY{c+c1}{\PYZsh{} l2 prefix since there are l2\PYZhy{}normalized visual features}
        \PY{n}{PERCEPTION\PYZus{}LAYER}\PY{o}{=}\PY{l+s+s1}{\PYZsq{}}\PY{l+s+s1}{l2\PYZus{}res5c\PYZhy{}152}\PY{l+s+s1}{\PYZsq{}} 
        
        \PY{n}{train\PYZus{}visual\PYZus{}features} \PY{o}{=} \PY{n}{dp}\PY{p}{[}\PY{l+s+s1}{\PYZsq{}}\PY{l+s+s1}{perception}\PY{l+s+s1}{\PYZsq{}}\PY{p}{]}\PY{p}{(}
            \PY{n}{train\PYZus{}or\PYZus{}test}\PY{o}{=}\PY{l+s+s1}{\PYZsq{}}\PY{l+s+s1}{train}\PY{l+s+s1}{\PYZsq{}}\PY{p}{,}
            \PY{n}{names\PYZus{}list}\PY{o}{=}\PY{n}{train\PYZus{}image\PYZus{}names}\PY{p}{,}
            \PY{n}{parts\PYZus{}extractor}\PY{o}{=}\PY{n+nb+bp}{None}\PY{p}{,}
            \PY{n}{max\PYZus{}parts}\PY{o}{=}\PY{n+nb+bp}{None}\PY{p}{,}
            \PY{n}{perception}\PY{o}{=}\PY{n}{CNN\PYZus{}NAME}\PY{p}{,}
            \PY{n}{layer}\PY{o}{=}\PY{n}{PERCEPTION\PYZus{}LAYER}\PY{p}{,}
            \PY{n}{second\PYZus{}layer}\PY{o}{=}\PY{n+nb+bp}{None}
            \PY{p}{)}
        \PY{n}{train\PYZus{}visual\PYZus{}features}\PY{o}{.}\PY{n}{shape}
\end{Verbatim}

\section{Vision+Language}\label{visionlanguage}

Given visual features, we can now build a full model that answer questions about images.
As we can see in \figref{tutorial:challenges}, it is hard to answer correctly on questions without seeing images.

\noindent Let us create an input as a pair of textual and visual features using the following code.

    \begin{Verbatim}[commandchars=\\\{\}]
{\color{incolor}In [{\color{incolor} }]:} \PY{n}{train\PYZus{}input} \PY{o}{=} \PY{p}{[}\PY{n}{train\PYZus{}x}\PY{p}{,} \PY{n}{train\PYZus{}visual\PYZus{}features}\PY{p}{]}
\end{Verbatim}

\noindent In the following, we investigate two approaches to question answering: an orderless BOW, and an RNN.
\subsubsection{BOW + Vision}\label{bow-vision}
\begin{figure}[htbp]
\centering
\includegraphics[width=0.75\linewidth]{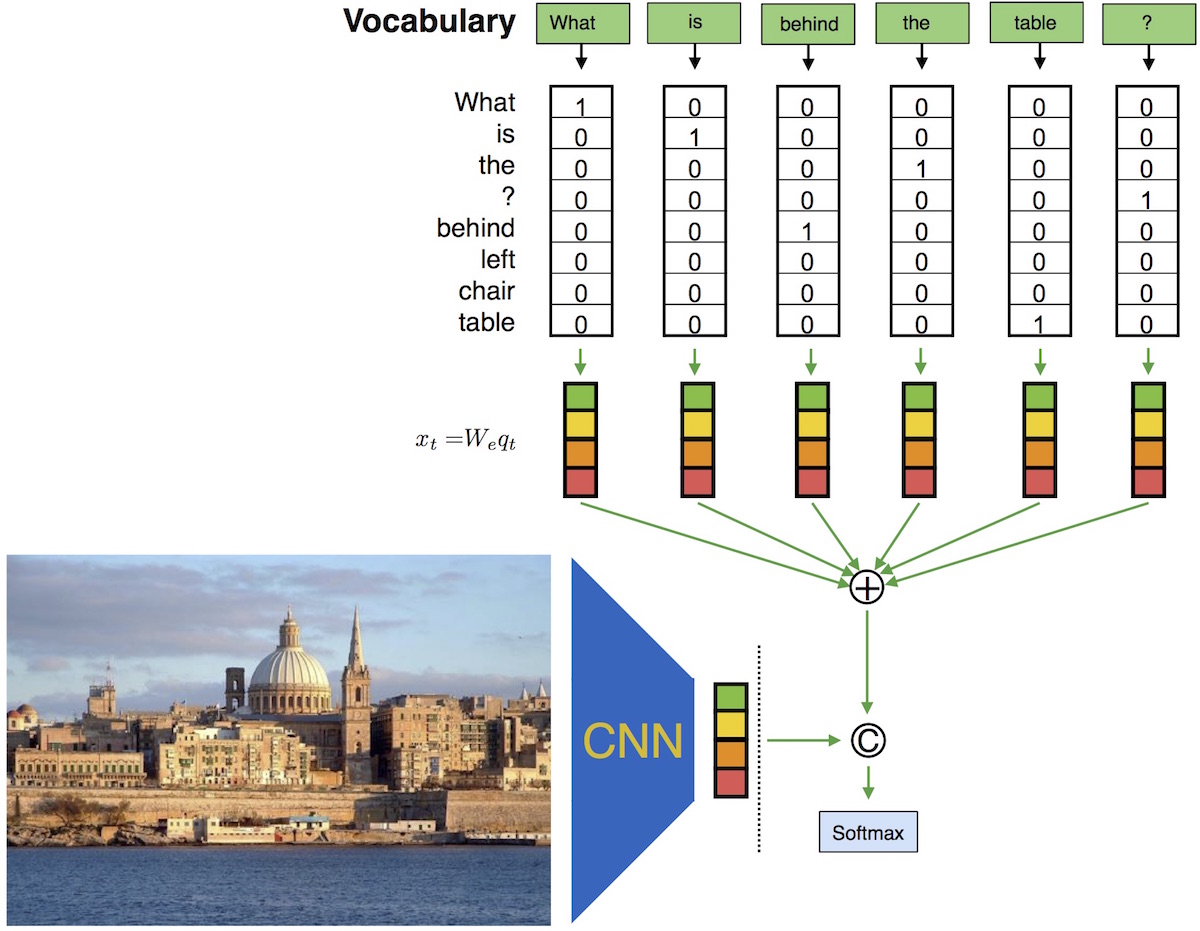}
\caption{BOW with visual features.}
\label{tutorial:bow_vision}
\end{figure}

Similarly to our blind model, we start with a BOW encoding of a question.
Here, we  explore two ways of combining both modalities (circle with
`C' in \figref{tutorial:bow_vision}): concatenation, and piece-wise multiplication.
For the sake of simplicity, we do not fine-tune the visual representation (dotted line symbolizes 
the barrier that blocks back-propagation in \figref{tutorial:bow_vision}).

    \begin{Verbatim}[commandchars=\\\{\}]
{\color{incolor}In [{\color{incolor} }]:} \PY{c+c1}{\PYZsh{}== Model definition}
        
        \PY{c+c1}{\PYZsh{} First we define a model using keras/kraino}
        \PY{k+kn}{from} \PY{n+nn}{keras.models} \PY{k+kn}{import} \PY{n}{Sequential}
        \PY{k+kn}{from} \PY{n+nn}{keras.layers.core} \PY{k+kn}{import} \PY{n}{Activation}
        \PY{k+kn}{from} \PY{n+nn}{keras.layers.core} \PY{k+kn}{import} \PY{n}{Dense}
        \PY{k+kn}{from} \PY{n+nn}{keras.layers.core} \PY{k+kn}{import} \PY{n}{Dropout}
        \PY{k+kn}{from} \PY{n+nn}{keras.layers.core} \PY{k+kn}{import} \PY{n}{Layer}
        \PY{k+kn}{from} \PY{n+nn}{keras.layers.core} \PY{k+kn}{import} \PY{n}{Merge}
        \PY{k+kn}{from} \PY{n+nn}{keras.layers.core} \PY{k+kn}{import} \PY{n}{TimeDistributedMerge}
        \PY{k+kn}{from} \PY{n+nn}{keras.layers.embeddings} \PY{k+kn}{import} \PY{n}{Embedding}
        
        \PY{k+kn}{from} \PY{n+nn}{kraino.core.model\PYZus{}zoo} \PY{k+kn}{import} \PY{n}{AbstractSequentialModel}
        \PY{k+kn}{from} \PY{n+nn}{kraino.core.model\PYZus{}zoo} \PY{k+kn}{import} \PY{n}{AbstractSingleAnswer}
        \PY{k+kn}{from} \PY{n+nn}{kraino.core.model\PYZus{}zoo} \PY{k+kn}{import} \PY{n}{AbstractSequentialMultiplewordAnswer}
        \PY{k+kn}{from} \PY{n+nn}{kraino.core.model\PYZus{}zoo} \PY{k+kn}{import} \PY{n}{Config}
        \PY{k+kn}{from} \PY{n+nn}{kraino.core.keras\PYZus{}extensions} \PY{k+kn}{import} \PY{n}{DropMask}
        \PY{k+kn}{from} \PY{n+nn}{kraino.core.keras\PYZus{}extensions} \PY{k+kn}{import} \PY{n}{LambdaWithMask}
        \PY{k+kn}{from} \PY{n+nn}{kraino.core.keras\PYZus{}extensions} \PY{k+kn}{import} \PY{n}{time\PYZus{}distributed\PYZus{}masked\PYZus{}ave}
        
        \PY{c+c1}{\PYZsh{} This model inherits from AbstractSingleAnswer, }
        \PY{c+c1}{\PYZsh{} and so it produces single answer words}
        \PY{c+c1}{\PYZsh{} To use multiple answer words, }
        \PY{c+c1}{\PYZsh{} you need to inherit from AbstractSequentialMultiplewordAnswer}
        \PY{k}{class} \PY{n+nc}{VisionLanguageBOW}\PY{p}{(}\PY{n}{AbstractSequentialModel}\PY{p}{,} \PY{n}{AbstractSingleAnswer}\PY{p}{)}\PY{p}{:}
            \PY{l+s+sd}{\PYZdq{}\PYZdq{}\PYZdq{}}
        \PY{l+s+sd}{    BOW Language only model that produces single word answers.}
        \PY{l+s+sd}{    \PYZdq{}\PYZdq{}\PYZdq{}}
            \PY{k}{def} \PY{n+nf}{create}\PY{p}{(}\PY{n+nb+bp}{self}\PY{p}{)}\PY{p}{:}
                \PY{n}{language\PYZus{}model} \PY{o}{=} \PY{n}{Sequential}\PY{p}{(}\PY{p}{)}
                \PY{n}{language\PYZus{}model}\PY{o}{.}\PY{n}{add}\PY{p}{(}\PY{n}{Embedding}\PY{p}{(}
                        \PY{n+nb+bp}{self}\PY{o}{.}\PY{n}{\PYZus{}config}\PY{o}{.}\PY{n}{input\PYZus{}dim}\PY{p}{,} 
                        \PY{n+nb+bp}{self}\PY{o}{.}\PY{n}{\PYZus{}config}\PY{o}{.}\PY{n}{textual\PYZus{}embedding\PYZus{}dim}\PY{p}{,} 
                        \PY{n}{mask\PYZus{}zero}\PY{o}{=}\PY{n+nb+bp}{True}\PY{p}{)}\PY{p}{)}
                \PY{n}{language\PYZus{}model}\PY{o}{.}\PY{n}{add}\PY{p}{(}\PY{n}{LambdaWithMask}\PY{p}{(}
                        \PY{n}{time\PYZus{}distributed\PYZus{}masked\PYZus{}ave}\PY{p}{,} 
                        \PY{n}{output\PYZus{}shape}\PY{o}{=}\PY{p}{[}\PY{n}{language\PYZus{}model}\PY{o}{.}\PY{n}{output\PYZus{}shape}\PY{p}{[}\PY{l+m+mi}{2}\PY{p}{]}\PY{p}{]}\PY{p}{)}\PY{p}{)}
                \PY{n}{language\PYZus{}model}\PY{o}{.}\PY{n}{add}\PY{p}{(}\PY{n}{DropMask}\PY{p}{(}\PY{p}{)}\PY{p}{)}
                \PY{n}{visual\PYZus{}model} \PY{o}{=} \PY{n}{Sequential}\PY{p}{(}\PY{p}{)}
                \PY{k}{if} \PY{n+nb+bp}{self}\PY{o}{.}\PY{n}{\PYZus{}config}\PY{o}{.}\PY{n}{visual\PYZus{}embedding\PYZus{}dim} \PY{o}{\PYZgt{}} \PY{l+m+mi}{0}\PY{p}{:}
                    \PY{n}{visual\PYZus{}model}\PY{o}{.}\PY{n}{add}\PY{p}{(}\PY{n}{Dense}\PY{p}{(}
                            \PY{n+nb+bp}{self}\PY{o}{.}\PY{n}{\PYZus{}config}\PY{o}{.}\PY{n}{visual\PYZus{}embedding\PYZus{}dim}\PY{p}{,}
                            \PY{n}{input\PYZus{}shape}\PY{o}{=}\PY{p}{(}\PY{n+nb+bp}{self}\PY{o}{.}\PY{n}{\PYZus{}config}\PY{o}{.}\PY{n}{visual\PYZus{}dim}\PY{p}{,}\PY{p}{)}\PY{p}{)}\PY{p}{)}
                \PY{k}{else}\PY{p}{:}
                    \PY{n}{visual\PYZus{}model}\PY{o}{.}\PY{n}{add}\PY{p}{(}\PY{n}{Layer}\PY{p}{(}\PY{n}{input\PYZus{}shape}\PY{o}{=}\PY{p}{(}\PY{n+nb+bp}{self}\PY{o}{.}\PY{n}{\PYZus{}config}\PY{o}{.}\PY{n}{visual\PYZus{}dim}\PY{p}{,}\PY{p}{)}\PY{p}{)}\PY{p}{)}
                \PY{n+nb+bp}{self}\PY{o}{.}\PY{n}{add}\PY{p}{(}\PY{n}{Merge}\PY{p}{(}\PY{p}{[}\PY{n}{language\PYZus{}model}\PY{p}{,} 
                        \PY{n}{visual\PYZus{}model}\PY{p}{]}\PY{p}{,} \PY{n}{mode}\PY{o}{=}\PY{n+nb+bp}{self}\PY{o}{.}\PY{n}{\PYZus{}config}\PY{o}{.}\PY{n}{multimodal\PYZus{}merge\PYZus{}mode}\PY{p}{)}\PY{p}{)}
                \PY{n+nb+bp}{self}\PY{o}{.}\PY{n}{add}\PY{p}{(}\PY{n}{Dropout}\PY{p}{(}\PY{l+m+mf}{0.5}\PY{p}{)}\PY{p}{)}
                \PY{n+nb+bp}{self}\PY{o}{.}\PY{n}{add}\PY{p}{(}\PY{n}{Dense}\PY{p}{(}\PY{n+nb+bp}{self}\PY{o}{.}\PY{n}{\PYZus{}config}\PY{o}{.}\PY{n}{output\PYZus{}dim}\PY{p}{)}\PY{p}{)}
                \PY{n+nb+bp}{self}\PY{o}{.}\PY{n}{add}\PY{p}{(}\PY{n}{Activation}\PY{p}{(}\PY{l+s+s1}{\PYZsq{}}\PY{l+s+s1}{softmax}\PY{l+s+s1}{\PYZsq{}}\PY{p}{)}\PY{p}{)}
                
\end{Verbatim}

    \begin{Verbatim}[commandchars=\\\{\}]
{\color{incolor}In [{\color{incolor} }]:} \PY{c+c1}{\PYZsh{} dimensionality of embeddings}
        \PY{n}{EMBEDDING\PYZus{}DIM} \PY{o}{=} \PY{l+m+mi}{500}
        \PY{c+c1}{\PYZsh{} kind of multimodal fusion (ave, concat, mul, sum)}
        \PY{n}{MULTIMODAL\PYZus{}MERGE\PYZus{}MODE} \PY{o}{=} \PY{l+s+s1}{\PYZsq{}}\PY{l+s+s1}{concat}\PY{l+s+s1}{\PYZsq{}}
        
        \PY{n}{model\PYZus{}config} \PY{o}{=} \PY{n}{Config}\PY{p}{(}
            \PY{n}{textual\PYZus{}embedding\PYZus{}dim}\PY{o}{=}\PY{n}{EMBEDDING\PYZus{}DIM}\PY{p}{,}
            \PY{n}{visual\PYZus{}embedding\PYZus{}dim}\PY{o}{=}\PY{l+m+mi}{0}\PY{p}{,}
            \PY{n}{multimodal\PYZus{}merge\PYZus{}mode}\PY{o}{=}\PY{n}{MULTIMODAL\PYZus{}MERGE\PYZus{}MODE}\PY{p}{,}
            \PY{n}{input\PYZus{}dim}\PY{o}{=}\PY{n+nb}{len}\PY{p}{(}\PY{n}{word2index\PYZus{}x}\PY{o}{.}\PY{n}{keys}\PY{p}{(}\PY{p}{)}\PY{p}{)}\PY{p}{,}
            \PY{n}{output\PYZus{}dim}\PY{o}{=}\PY{n+nb}{len}\PY{p}{(}\PY{n}{word2index\PYZus{}y}\PY{o}{.}\PY{n}{keys}\PY{p}{(}\PY{p}{)}\PY{p}{)}\PY{p}{,}
            \PY{n}{visual\PYZus{}dim}\PY{o}{=}\PY{n}{train\PYZus{}visual\PYZus{}features}\PY{o}{.}\PY{n}{shape}\PY{p}{[}\PY{l+m+mi}{1}\PY{p}{]}\PY{p}{)}
        \PY{n}{model} \PY{o}{=} \PY{n}{VisionLanguageBOW}\PY{p}{(}\PY{n}{model\PYZus{}config}\PY{p}{)}
        \PY{n}{model}\PY{o}{.}\PY{n}{create}\PY{p}{(}\PY{p}{)}
        \PY{n}{model}\PY{o}{.}\PY{n}{compile}\PY{p}{(}
            \PY{n}{loss}\PY{o}{=}\PY{l+s+s1}{\PYZsq{}}\PY{l+s+s1}{categorical\PYZus{}crossentropy}\PY{l+s+s1}{\PYZsq{}}\PY{p}{,} 
            \PY{n}{optimizer}\PY{o}{=}\PY{l+s+s1}{\PYZsq{}}\PY{l+s+s1}{adam}\PY{l+s+s1}{\PYZsq{}}\PY{p}{)}
\end{Verbatim}

    \begin{Verbatim}[commandchars=\\\{\}]
{\color{incolor}In [{\color{incolor} }]:} \PY{c+c1}{\PYZsh{}== Model training}
        \PY{n}{model}\PY{o}{.}\PY{n}{fit}\PY{p}{(}
            \PY{n}{train\PYZus{}input}\PY{p}{,} 
            \PY{n}{train\PYZus{}y}\PY{p}{,}
            \PY{n}{batch\PYZus{}size}\PY{o}{=}\PY{l+m+mi}{512}\PY{p}{,}
            \PY{n}{nb\PYZus{}epoch}\PY{o}{=}\PY{l+m+mi}{40}\PY{p}{,}
            \PY{n}{validation\PYZus{}split}\PY{o}{=}\PY{l+m+mf}{0.1}\PY{p}{,}
            \PY{n}{show\PYZus{}accuracy}\PY{o}{=}\PY{n+nb+bp}{True}\PY{p}{)}
\end{Verbatim}

\noindent To achieve better results, we can use another operator that combines both modalities together.
For instance, we can use a piece-wise multiplication.

    \begin{Verbatim}[commandchars=\\\{\}]
{\color{incolor}In [{\color{incolor} }]:} \PY{c+c1}{\PYZsh{}== Model definition}
        
        \PY{c+c1}{\PYZsh{} First we define a model using keras/kraino}
        \PY{k+kn}{from} \PY{n+nn}{keras.models} \PY{k+kn}{import} \PY{n}{Sequential}
        \PY{k+kn}{from} \PY{n+nn}{keras.layers.core} \PY{k+kn}{import} \PY{n}{Activation}
        \PY{k+kn}{from} \PY{n+nn}{keras.layers.core} \PY{k+kn}{import} \PY{n}{Dense}
        \PY{k+kn}{from} \PY{n+nn}{keras.layers.core} \PY{k+kn}{import} \PY{n}{Dropout}
        \PY{k+kn}{from} \PY{n+nn}{keras.layers.core} \PY{k+kn}{import} \PY{n}{Layer}
        \PY{k+kn}{from} \PY{n+nn}{keras.layers.core} \PY{k+kn}{import} \PY{n}{Merge}
        \PY{k+kn}{from} \PY{n+nn}{keras.layers.core} \PY{k+kn}{import} \PY{n}{TimeDistributedMerge}
        \PY{k+kn}{from} \PY{n+nn}{keras.layers.embeddings} \PY{k+kn}{import} \PY{n}{Embedding}
        
        \PY{k+kn}{from} \PY{n+nn}{kraino.core.model\PYZus{}zoo} \PY{k+kn}{import} \PY{n}{AbstractSequentialModel}
        \PY{k+kn}{from} \PY{n+nn}{kraino.core.model\PYZus{}zoo} \PY{k+kn}{import} \PY{n}{AbstractSingleAnswer}
        \PY{k+kn}{from} \PY{n+nn}{kraino.core.model\PYZus{}zoo} \PY{k+kn}{import} \PY{n}{AbstractSequentialMultiplewordAnswer}
        \PY{k+kn}{from} \PY{n+nn}{kraino.core.model\PYZus{}zoo} \PY{k+kn}{import} \PY{n}{Config}
        \PY{k+kn}{from} \PY{n+nn}{kraino.core.keras\PYZus{}extensions} \PY{k+kn}{import} \PY{n}{DropMask}
        \PY{k+kn}{from} \PY{n+nn}{kraino.core.keras\PYZus{}extensions} \PY{k+kn}{import} \PY{n}{LambdaWithMask}
        \PY{k+kn}{from} \PY{n+nn}{kraino.core.keras\PYZus{}extensions} \PY{k+kn}{import} \PY{n}{time\PYZus{}distributed\PYZus{}masked\PYZus{}ave}
        
        \PY{c+c1}{\PYZsh{} This model inherits from AbstractSingleAnswer, }
        \PY{c+c1}{\PYZsh{} and so it produces single answer words}
        \PY{c+c1}{\PYZsh{} To use multiple answer words, }
        \PY{c+c1}{\PYZsh{} you need to inherit from AbstractSequentialMultiplewordAnswer}
        \PY{k}{class} \PY{n+nc}{VisionLanguageBOW}\PY{p}{(}\PY{n}{AbstractSequentialModel}\PY{p}{,} \PY{n}{AbstractSingleAnswer}\PY{p}{)}\PY{p}{:}
            \PY{l+s+sd}{\PYZdq{}\PYZdq{}\PYZdq{}}
        \PY{l+s+sd}{    BOW Language only model that produces single word answers.}
        \PY{l+s+sd}{    \PYZdq{}\PYZdq{}\PYZdq{}}
            \PY{k}{def} \PY{n+nf}{create}\PY{p}{(}\PY{n+nb+bp}{self}\PY{p}{)}\PY{p}{:}
                \PY{n}{language\PYZus{}model} \PY{o}{=} \PY{n}{Sequential}\PY{p}{(}\PY{p}{)}
                \PY{n}{language\PYZus{}model}\PY{o}{.}\PY{n}{add}\PY{p}{(}\PY{n}{Embedding}\PY{p}{(}
                        \PY{n+nb+bp}{self}\PY{o}{.}\PY{n}{\PYZus{}config}\PY{o}{.}\PY{n}{input\PYZus{}dim}\PY{p}{,} 
                        \PY{n+nb+bp}{self}\PY{o}{.}\PY{n}{\PYZus{}config}\PY{o}{.}\PY{n}{textual\PYZus{}embedding\PYZus{}dim}\PY{p}{,} 
                        \PY{n}{mask\PYZus{}zero}\PY{o}{=}\PY{n+nb+bp}{True}\PY{p}{)}\PY{p}{)}
                \PY{n}{language\PYZus{}model}\PY{o}{.}\PY{n}{add}\PY{p}{(}\PY{n}{LambdaWithMask}\PY{p}{(}
                        \PY{n}{time\PYZus{}distributed\PYZus{}masked\PYZus{}ave}\PY{p}{,} 
                        \PY{n}{output\PYZus{}shape}\PY{o}{=}\PY{p}{[}\PY{n}{language\PYZus{}model}\PY{o}{.}\PY{n}{output\PYZus{}shape}\PY{p}{[}\PY{l+m+mi}{2}\PY{p}{]}\PY{p}{]}\PY{p}{)}\PY{p}{)}
                \PY{n}{language\PYZus{}model}\PY{o}{.}\PY{n}{add}\PY{p}{(}\PY{n}{DropMask}\PY{p}{(}\PY{p}{)}\PY{p}{)}
                \PY{n}{visual\PYZus{}model} \PY{o}{=} \PY{n}{Sequential}\PY{p}{(}\PY{p}{)}
                \PY{k}{if} \PY{n+nb+bp}{self}\PY{o}{.}\PY{n}{\PYZus{}config}\PY{o}{.}\PY{n}{visual\PYZus{}embedding\PYZus{}dim} \PY{o}{\PYZgt{}} \PY{l+m+mi}{0}\PY{p}{:}
                    \PY{n}{visual\PYZus{}model}\PY{o}{.}\PY{n}{add}\PY{p}{(}\PY{n}{Dense}\PY{p}{(}
                            \PY{n+nb+bp}{self}\PY{o}{.}\PY{n}{\PYZus{}config}\PY{o}{.}\PY{n}{visual\PYZus{}embedding\PYZus{}dim}\PY{p}{,}
                            \PY{n}{input\PYZus{}shape}\PY{o}{=}\PY{p}{(}\PY{n+nb+bp}{self}\PY{o}{.}\PY{n}{\PYZus{}config}\PY{o}{.}\PY{n}{visual\PYZus{}dim}\PY{p}{,}\PY{p}{)}\PY{p}{)}\PY{p}{)}
                \PY{k}{else}\PY{p}{:}
                    \PY{n}{visual\PYZus{}model}\PY{o}{.}\PY{n}{add}\PY{p}{(}\PY{n}{Layer}\PY{p}{(}\PY{n}{input\PYZus{}shape}\PY{o}{=}\PY{p}{(}\PY{n+nb+bp}{self}\PY{o}{.}\PY{n}{\PYZus{}config}\PY{o}{.}\PY{n}{visual\PYZus{}dim}\PY{p}{,}\PY{p}{)}\PY{p}{)}\PY{p}{)}
                \PY{n+nb+bp}{self}\PY{o}{.}\PY{n}{add}\PY{p}{(}\PY{n}{Merge}\PY{p}{(}\PY{p}{[}\PY{n}{language\PYZus{}model}\PY{p}{,} 
                            \PY{n}{visual\PYZus{}model}\PY{p}{]}\PY{p}{,} \PY{n}{mode}\PY{o}{=}\PY{n+nb+bp}{self}\PY{o}{.}\PY{n}{\PYZus{}config}\PY{o}{.}\PY{n}{multimodal\PYZus{}merge\PYZus{}mode}\PY{p}{)}\PY{p}{)}
                \PY{n+nb+bp}{self}\PY{o}{.}\PY{n}{add}\PY{p}{(}\PY{n}{Dropout}\PY{p}{(}\PY{l+m+mf}{0.5}\PY{p}{)}\PY{p}{)}
                \PY{n+nb+bp}{self}\PY{o}{.}\PY{n}{add}\PY{p}{(}\PY{n}{Dense}\PY{p}{(}\PY{n+nb+bp}{self}\PY{o}{.}\PY{n}{\PYZus{}config}\PY{o}{.}\PY{n}{output\PYZus{}dim}\PY{p}{)}\PY{p}{)}
                \PY{n+nb+bp}{self}\PY{o}{.}\PY{n}{add}\PY{p}{(}\PY{n}{Activation}\PY{p}{(}\PY{l+s+s1}{\PYZsq{}}\PY{l+s+s1}{softmax}\PY{l+s+s1}{\PYZsq{}}\PY{p}{)}\PY{p}{)}
                
\end{Verbatim}

    \begin{Verbatim}[commandchars=\\\{\}]
{\color{incolor}In [{\color{incolor} }]:} \PY{c+c1}{\PYZsh{} dimensionality of embeddings}
        \PY{n}{EMBEDDING\PYZus{}DIM} \PY{o}{=} \PY{l+m+mi}{500}
        \PY{c+c1}{\PYZsh{} kind of multimodal fusion (ave, concat, mul, sum)}
        \PY{n}{MULTIMODAL\PYZus{}MERGE\PYZus{}MODE} \PY{o}{=} \PY{l+s+s1}{\PYZsq{}}\PY{l+s+s1}{mul}\PY{l+s+s1}{\PYZsq{}}
        
        \PY{n}{model\PYZus{}config} \PY{o}{=} \PY{n}{Config}\PY{p}{(}
            \PY{n}{textual\PYZus{}embedding\PYZus{}dim}\PY{o}{=}\PY{n}{EMBEDDING\PYZus{}DIM}\PY{p}{,}
            \PY{n}{visual\PYZus{}embedding\PYZus{}dim}\PY{o}{=}\PY{n}{EMBEDDING\PYZus{}DIM}\PY{p}{,}
            \PY{n}{multimodal\PYZus{}merge\PYZus{}mode}\PY{o}{=}\PY{n}{MULTIMODAL\PYZus{}MERGE\PYZus{}MODE}\PY{p}{,}
            \PY{n}{input\PYZus{}dim}\PY{o}{=}\PY{n+nb}{len}\PY{p}{(}\PY{n}{word2index\PYZus{}x}\PY{o}{.}\PY{n}{keys}\PY{p}{(}\PY{p}{)}\PY{p}{)}\PY{p}{,}
            \PY{n}{output\PYZus{}dim}\PY{o}{=}\PY{n+nb}{len}\PY{p}{(}\PY{n}{word2index\PYZus{}y}\PY{o}{.}\PY{n}{keys}\PY{p}{(}\PY{p}{)}\PY{p}{)}\PY{p}{,}
            \PY{n}{visual\PYZus{}dim}\PY{o}{=}\PY{n}{train\PYZus{}visual\PYZus{}features}\PY{o}{.}\PY{n}{shape}\PY{p}{[}\PY{l+m+mi}{1}\PY{p}{]}\PY{p}{)}
        \PY{n}{model} \PY{o}{=} \PY{n}{VisionLanguageBOW}\PY{p}{(}\PY{n}{model\PYZus{}config}\PY{p}{)}
        \PY{n}{model}\PY{o}{.}\PY{n}{create}\PY{p}{(}\PY{p}{)}
        \PY{n}{model}\PY{o}{.}\PY{n}{compile}\PY{p}{(}
            \PY{n}{loss}\PY{o}{=}\PY{l+s+s1}{\PYZsq{}}\PY{l+s+s1}{categorical\PYZus{}crossentropy}\PY{l+s+s1}{\PYZsq{}}\PY{p}{,} 
            \PY{n}{optimizer}\PY{o}{=}\PY{l+s+s1}{\PYZsq{}}\PY{l+s+s1}{adam}\PY{l+s+s1}{\PYZsq{}}\PY{p}{)}
        \PY{n}{text\PYZus{}image\PYZus{}bow\PYZus{}model} \PY{o}{=} \PY{n}{model}
\end{Verbatim}

    \begin{Verbatim}[commandchars=\\\{\}]
{\color{incolor}In [{\color{incolor} }]:} \PY{c+c1}{\PYZsh{}== Model training}
        \PY{n}{text\PYZus{}image\PYZus{}bow\PYZus{}model}\PY{o}{.}\PY{n}{fit}\PY{p}{(}
            \PY{n}{train\PYZus{}input}\PY{p}{,} 
            \PY{n}{train\PYZus{}y}\PY{p}{,}
            \PY{n}{batch\PYZus{}size}\PY{o}{=}\PY{l+m+mi}{512}\PY{p}{,}
            \PY{n}{nb\PYZus{}epoch}\PY{o}{=}\PY{l+m+mi}{40}\PY{p}{,}
            \PY{n}{validation\PYZus{}split}\PY{o}{=}\PY{l+m+mf}{0.1}\PY{p}{,}
            \PY{n}{show\PYZus{}accuracy}\PY{o}{=}\PY{n+nb+bp}{True}\PY{p}{)}
\end{Verbatim}

\noindent At the end of this section, a curious reader can try to answer the following questions.
    \begin{verbatim}
If we merge language and visual features with 'mul', 
do we need to set both embeddings to have the same number  of dimensions? 
That is, do we require to have textual_embedding_dim == visual_embedding_dim?
\end{verbatim}

\subsubsection{RNN + Vision}\label{rnn-vision}
\begin{figure}[htbp]
\centering
\includegraphics[width=0.66\linewidth]{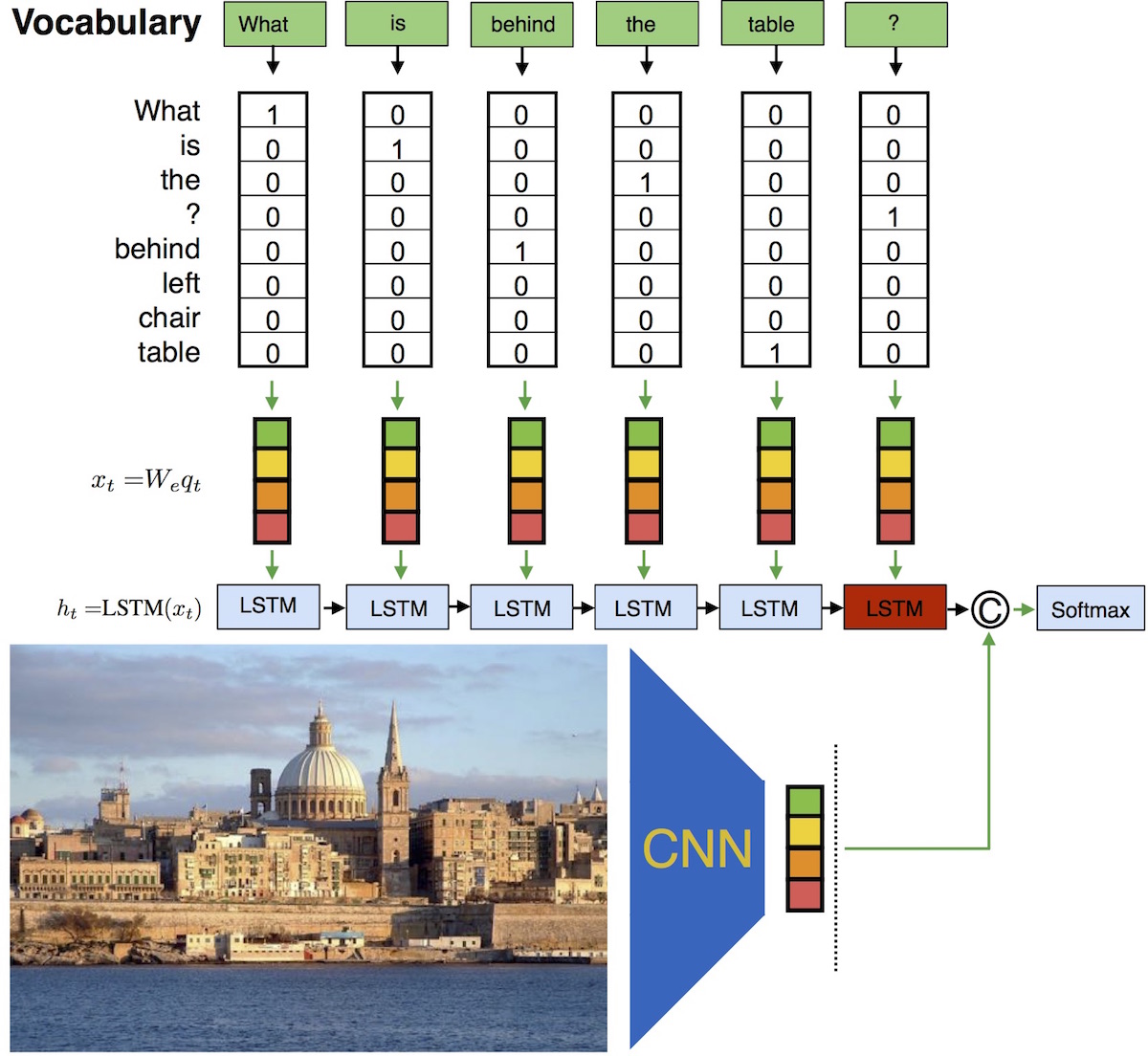}
\caption{RNN with visual features.}
\label{tutorial:RNN_vision}
\end{figure}

Now, we  repeat the BOW experiments but with RNN.
\figref{tutorial:RNN_vision} depicts the architecture.

    \begin{Verbatim}[commandchars=\\\{\}]
{\color{incolor}In [{\color{incolor} }]:} \PY{c+c1}{\PYZsh{}== Model definition}
        
        \PY{c+c1}{\PYZsh{} First we define a model using keras/kraino}
        \PY{k+kn}{from} \PY{n+nn}{keras.models} \PY{k+kn}{import} \PY{n}{Sequential}
        \PY{k+kn}{from} \PY{n+nn}{keras.layers.core} \PY{k+kn}{import} \PY{n}{Activation}
        \PY{k+kn}{from} \PY{n+nn}{keras.layers.core} \PY{k+kn}{import} \PY{n}{Dense}
        \PY{k+kn}{from} \PY{n+nn}{keras.layers.core} \PY{k+kn}{import} \PY{n}{Dropout}
        \PY{k+kn}{from} \PY{n+nn}{keras.layers.core} \PY{k+kn}{import} \PY{n}{Layer}
        \PY{k+kn}{from} \PY{n+nn}{keras.layers.core} \PY{k+kn}{import} \PY{n}{Merge}
        \PY{k+kn}{from} \PY{n+nn}{keras.layers.core} \PY{k+kn}{import} \PY{n}{TimeDistributedMerge}
        \PY{k+kn}{from} \PY{n+nn}{keras.layers.embeddings} \PY{k+kn}{import} \PY{n}{Embedding}
        \PY{k+kn}{from} \PY{n+nn}{keras.layers.recurrent} \PY{k+kn}{import} \PY{n}{GRU}
        \PY{k+kn}{from} \PY{n+nn}{keras.layers.recurrent} \PY{k+kn}{import} \PY{n}{LSTM}
        
        \PY{k+kn}{from} \PY{n+nn}{kraino.core.model\PYZus{}zoo} \PY{k+kn}{import} \PY{n}{AbstractSequentialModel}
        \PY{k+kn}{from} \PY{n+nn}{kraino.core.model\PYZus{}zoo} \PY{k+kn}{import} \PY{n}{AbstractSingleAnswer}
        \PY{k+kn}{from} \PY{n+nn}{kraino.core.model\PYZus{}zoo} \PY{k+kn}{import} \PY{n}{AbstractSequentialMultiplewordAnswer}
        \PY{k+kn}{from} \PY{n+nn}{kraino.core.model\PYZus{}zoo} \PY{k+kn}{import} \PY{n}{Config}
        \PY{k+kn}{from} \PY{n+nn}{kraino.core.keras\PYZus{}extensions} \PY{k+kn}{import} \PY{n}{DropMask}
        \PY{k+kn}{from} \PY{n+nn}{kraino.core.keras\PYZus{}extensions} \PY{k+kn}{import} \PY{n}{LambdaWithMask}
        \PY{k+kn}{from} \PY{n+nn}{kraino.core.keras\PYZus{}extensions} \PY{k+kn}{import} \PY{n}{time\PYZus{}distributed\PYZus{}masked\PYZus{}ave}
        
        \PY{c+c1}{\PYZsh{} This model inherits from AbstractSingleAnswer, }
        \PY{c+c1}{\PYZsh{} and so it produces single answer words}
        \PY{c+c1}{\PYZsh{} To use multiple answer words, }
        \PY{c+c1}{\PYZsh{} you need to inherit from AbstractSequentialMultiplewordAnswer}
        \PY{k}{class} \PY{n+nc}{VisionLanguageLSTM}\PY{p}{(}\PY{n}{AbstractSequentialModel}\PY{p}{,} \PY{n}{AbstractSingleAnswer}\PY{p}{)}\PY{p}{:}
            \PY{l+s+sd}{\PYZdq{}\PYZdq{}\PYZdq{}}
        \PY{l+s+sd}{    BOW Language only model that produces single word answers.}
        \PY{l+s+sd}{    \PYZdq{}\PYZdq{}\PYZdq{}}
            \PY{k}{def} \PY{n+nf}{create}\PY{p}{(}\PY{n+nb+bp}{self}\PY{p}{)}\PY{p}{:}
                \PY{n}{language\PYZus{}model} \PY{o}{=} \PY{n}{Sequential}\PY{p}{(}\PY{p}{)}
                \PY{n}{language\PYZus{}model}\PY{o}{.}\PY{n}{add}\PY{p}{(}\PY{n}{Embedding}\PY{p}{(}
                        \PY{n+nb+bp}{self}\PY{o}{.}\PY{n}{\PYZus{}config}\PY{o}{.}\PY{n}{input\PYZus{}dim}\PY{p}{,} 
                        \PY{n+nb+bp}{self}\PY{o}{.}\PY{n}{\PYZus{}config}\PY{o}{.}\PY{n}{textual\PYZus{}embedding\PYZus{}dim}\PY{p}{,} 
                        \PY{n}{mask\PYZus{}zero}\PY{o}{=}\PY{n+nb+bp}{True}\PY{p}{)}\PY{p}{)}
  
                \PY{n}{language\PYZus{}model}\PY{o}{.}\PY{n}{add}\PY{p}{(}\PY{n}{LSTM}\PY{p}{(}\PY{n+nb+bp}{self}\PY{o}{.}\PY{n}{\PYZus{}config}\PY{o}{.}\PY{n}{hidden\PYZus{}state\PYZus{}dim}\PY{p}{,} 
                              \PY{n}{return\PYZus{}sequences}\PY{o}{=}\PY{n+nb+bp}{False}\PY{p}{)}\PY{p}{)}
        
                \PY{n}{visual\PYZus{}model} \PY{o}{=} \PY{n}{Sequential}\PY{p}{(}\PY{p}{)}
                \PY{k}{if} \PY{n+nb+bp}{self}\PY{o}{.}\PY{n}{\PYZus{}config}\PY{o}{.}\PY{n}{visual\PYZus{}embedding\PYZus{}dim} \PY{o}{\PYZgt{}} \PY{l+m+mi}{0}\PY{p}{:}
                    \PY{n}{visual\PYZus{}model}\PY{o}{.}\PY{n}{add}\PY{p}{(}\PY{n}{Dense}\PY{p}{(}
                            \PY{n+nb+bp}{self}\PY{o}{.}\PY{n}{\PYZus{}config}\PY{o}{.}\PY{n}{visual\PYZus{}embedding\PYZus{}dim}\PY{p}{,}
                            \PY{n}{input\PYZus{}shape}\PY{o}{=}\PY{p}{(}\PY{n+nb+bp}{self}\PY{o}{.}\PY{n}{\PYZus{}config}\PY{o}{.}\PY{n}{visual\PYZus{}dim}\PY{p}{,}\PY{p}{)}\PY{p}{)}\PY{p}{)}
                \PY{k}{else}\PY{p}{:}
                    \PY{n}{visual\PYZus{}model}\PY{o}{.}\PY{n}{add}\PY{p}{(}\PY{n}{Layer}\PY{p}{(}\PY{n}{input\PYZus{}shape}\PY{o}{=}\PY{p}{(}\PY{n+nb+bp}{self}\PY{o}{.}\PY{n}{\PYZus{}config}\PY{o}{.}\PY{n}{visual\PYZus{}dim}\PY{p}{,}\PY{p}{)}\PY{p}{)}\PY{p}{)}
                \PY{n+nb+bp}{self}\PY{o}{.}\PY{n}{add}\PY{p}{(}\PY{n}{Merge}\PY{p}{(}\PY{p}{[}\PY{n}{language\PYZus{}model}\PY{p}{,} 
                            \PY{n}{visual\PYZus{}model}\PY{p}{]}\PY{p}{,} \PY{n}{mode}\PY{o}{=}\PY{n+nb+bp}{self}\PY{o}{.}\PY{n}{\PYZus{}config}\PY{o}{.}\PY{n}{multimodal\PYZus{}merge\PYZus{}mode}\PY{p}{)}\PY{p}{)}
                \PY{n+nb+bp}{self}\PY{o}{.}\PY{n}{add}\PY{p}{(}\PY{n}{Dropout}\PY{p}{(}\PY{l+m+mf}{0.5}\PY{p}{)}\PY{p}{)}
                \PY{n+nb+bp}{self}\PY{o}{.}\PY{n}{add}\PY{p}{(}\PY{n}{Dense}\PY{p}{(}\PY{n+nb+bp}{self}\PY{o}{.}\PY{n}{\PYZus{}config}\PY{o}{.}\PY{n}{output\PYZus{}dim}\PY{p}{)}\PY{p}{)}
                \PY{n+nb+bp}{self}\PY{o}{.}\PY{n}{add}\PY{p}{(}\PY{n}{Activation}\PY{p}{(}\PY{l+s+s1}{\PYZsq{}}\PY{l+s+s1}{softmax}\PY{l+s+s1}{\PYZsq{}}\PY{p}{)}\PY{p}{)}
                
                
        \PY{c+c1}{\PYZsh{} dimensionality of embeddings}
        \PY{n}{EMBEDDING\PYZus{}DIM} \PY{o}{=} \PY{l+m+mi}{500}
        \PY{c+c1}{\PYZsh{} kind of multimodal fusion (ave, concat, mul, sum)}
        \PY{n}{MULTIMODAL\PYZus{}MERGE\PYZus{}MODE} \PY{o}{=} \PY{l+s+s1}{\PYZsq{}}\PY{l+s+s1}{sum}\PY{l+s+s1}{\PYZsq{}}
        
        \PY{n}{model\PYZus{}config} \PY{o}{=} \PY{n}{Config}\PY{p}{(}
            \PY{n}{textual\PYZus{}embedding\PYZus{}dim}\PY{o}{=}\PY{n}{EMBEDDING\PYZus{}DIM}\PY{p}{,}
            \PY{n}{visual\PYZus{}embedding\PYZus{}dim}\PY{o}{=}\PY{n}{EMBEDDING\PYZus{}DIM}\PY{p}{,}
            \PY{n}{hidden\PYZus{}state\PYZus{}dim}\PY{o}{=}\PY{n}{EMBEDDING\PYZus{}DIM}\PY{p}{,}
            \PY{n}{multimodal\PYZus{}merge\PYZus{}mode}\PY{o}{=}\PY{n}{MULTIMODAL\PYZus{}MERGE\PYZus{}MODE}\PY{p}{,}
            \PY{n}{input\PYZus{}dim}\PY{o}{=}\PY{n+nb}{len}\PY{p}{(}\PY{n}{word2index\PYZus{}x}\PY{o}{.}\PY{n}{keys}\PY{p}{(}\PY{p}{)}\PY{p}{)}\PY{p}{,}
            \PY{n}{output\PYZus{}dim}\PY{o}{=}\PY{n+nb}{len}\PY{p}{(}\PY{n}{word2index\PYZus{}y}\PY{o}{.}\PY{n}{keys}\PY{p}{(}\PY{p}{)}\PY{p}{)}\PY{p}{,}
            \PY{n}{visual\PYZus{}dim}\PY{o}{=}\PY{n}{train\PYZus{}visual\PYZus{}features}\PY{o}{.}\PY{n}{shape}\PY{p}{[}\PY{l+m+mi}{1}\PY{p}{]}\PY{p}{)}
        \PY{n}{model} \PY{o}{=} \PY{n}{VisionLanguageLSTM}\PY{p}{(}\PY{n}{model\PYZus{}config}\PY{p}{)}
        \PY{n}{model}\PY{o}{.}\PY{n}{create}\PY{p}{(}\PY{p}{)}
        \PY{n}{model}\PY{o}{.}\PY{n}{compile}\PY{p}{(}
            \PY{n}{loss}\PY{o}{=}\PY{l+s+s1}{\PYZsq{}}\PY{l+s+s1}{categorical\PYZus{}crossentropy}\PY{l+s+s1}{\PYZsq{}}\PY{p}{,} 
            \PY{n}{optimizer}\PY{o}{=}\PY{l+s+s1}{\PYZsq{}}\PY{l+s+s1}{adam}\PY{l+s+s1}{\PYZsq{}}\PY{p}{)}
        \PY{n}{text\PYZus{}image\PYZus{}rnn\PYZus{}model} \PY{o}{=} \PY{n}{model}
\end{Verbatim}

\subsubsection{Batch Size}\label{batch-size}
We can do training with batch size set to $512$. If an error occurs due to a memory consumption, lowering the batch size should help.
    \begin{Verbatim}[commandchars=\\\{\}]
{\color{incolor}In [{\color{incolor} }]:} \PY{c+c1}{\PYZsh{}== Model training}
        \PY{n}{text\PYZus{}image\PYZus{}rnn\PYZus{}model}\PY{o}{.}\PY{n}{fit}\PY{p}{(}
            \PY{n}{train\PYZus{}input}\PY{p}{,}
            \PY{n}{train\PYZus{}y}\PY{p}{,}
            \PY{n}{batch\PYZus{}size}\PY{o}{=}\PY{l+m+mi}{512}\PY{p}{,}
            \PY{n}{nb\PYZus{}epoch}\PY{o}{=}\PY{l+m+mi}{40}\PY{p}{,}
            \PY{n}{validation\PYZus{}split}\PY{o}{=}\PY{l+m+mf}{0.1}\PY{p}{,}
            \PY{n}{show\PYZus{}accuracy}\PY{o}{=}\PY{n+nb+bp}{True}\PY{p}{)}
\end{Verbatim}

\noindent A curious reader may experiment with a few different batch sizes, and answer the following questions.
    \begin{verbatim}
Can you experiment with batch-size=1, and next with batch-size=5000?
Can you explain both issues regarding the batch size? 
When do you get the best performance, with multiplication, concatenation, or summation?
\end{verbatim}

\paragraph{Summary}
As previously, using RNN makes the sequence processing order-aware. This
time, however, we combine two modalities so that the whole model `sees'
 images. Finally, it is also important how both modalities are combined.

    \section{New Predictions with
Vision+Language}\label{new-predictions-with-visionlanguage}

    \subsubsection{Predictions (Features)}\label{predictions-features}

    \begin{Verbatim}[commandchars=\\\{\}]
{\color{incolor}In [{\color{incolor} }]:} \PY{n}{test\PYZus{}image\PYZus{}names} \PY{o}{=} \PY{n}{test\PYZus{}text\PYZus{}representation}\PY{p}{[}\PY{l+s+s1}{\PYZsq{}}\PY{l+s+s1}{img\PYZus{}name}\PY{l+s+s1}{\PYZsq{}}\PY{p}{]}
        \PY{n}{test\PYZus{}visual\PYZus{}features} \PY{o}{=} \PY{n}{dp}\PY{p}{[}\PY{l+s+s1}{\PYZsq{}}\PY{l+s+s1}{perception}\PY{l+s+s1}{\PYZsq{}}\PY{p}{]}\PY{p}{(}
            \PY{n}{train\PYZus{}or\PYZus{}test}\PY{o}{=}\PY{l+s+s1}{\PYZsq{}}\PY{l+s+s1}{test}\PY{l+s+s1}{\PYZsq{}}\PY{p}{,}
            \PY{n}{names\PYZus{}list}\PY{o}{=}\PY{n}{test\PYZus{}image\PYZus{}names}\PY{p}{,}
            \PY{n}{parts\PYZus{}extractor}\PY{o}{=}\PY{n+nb+bp}{None}\PY{p}{,}
            \PY{n}{max\PYZus{}parts}\PY{o}{=}\PY{n+nb+bp}{None}\PY{p}{,}
            \PY{n}{perception}\PY{o}{=}\PY{n}{CNN\PYZus{}NAME}\PY{p}{,}
            \PY{n}{layer}\PY{o}{=}\PY{n}{PERCEPTION\PYZus{}LAYER}\PY{p}{,}
            \PY{n}{second\PYZus{}layer}\PY{o}{=}\PY{n+nb+bp}{None}
            \PY{p}{)}
        \PY{n}{test\PYZus{}visual\PYZus{}features}\PY{o}{.}\PY{n}{shape}
\end{Verbatim}

    \begin{Verbatim}[commandchars=\\\{\}]
{\color{incolor}In [{\color{incolor} }]:} \PY{n}{test\PYZus{}input} \PY{o}{=} \PY{p}{[}\PY{n}{test\PYZus{}x}\PY{p}{,} \PY{n}{test\PYZus{}visual\PYZus{}features}\PY{p}{]}
\end{Verbatim}

    \subsubsection{Predictions (Bow with
Vision)}\label{predictions-bow-with-vision}

    \begin{Verbatim}[commandchars=\\\{\}]
{\color{incolor}In [{\color{incolor} }]:} \PY{k+kn}{from} \PY{n+nn}{kraino.core.model\PYZus{}zoo} \PY{k+kn}{import} \PY{n}{word\PYZus{}generator}
        \PY{c+c1}{\PYZsh{} we first need to add word\PYZus{}generator to \PYZus{}config }
        \PY{c+c1}{\PYZsh{} (we could have done this before, in the Config constructor)}
        \PY{c+c1}{\PYZsh{} we use maximum likelihood as a word generator}
        \PY{n}{text\PYZus{}image\PYZus{}bow\PYZus{}model}\PY{o}{.}\PY{n}{\PYZus{}config}\PY{o}{.}\PY{n}{word\PYZus{}generator} \PY{o}{=} \PY{n}{word\PYZus{}generator}\PY{p}{[}\PY{l+s+s1}{\PYZsq{}}\PY{l+s+s1}{max\PYZus{}likelihood}\PY{l+s+s1}{\PYZsq{}}\PY{p}{]}
        \PY{n}{predictions\PYZus{}answers} \PY{o}{=} \PY{n}{text\PYZus{}image\PYZus{}bow\PYZus{}model}\PY{o}{.}\PY{n}{decode\PYZus{}predictions}\PY{p}{(}
            \PY{n}{X}\PY{o}{=}\PY{n}{test\PYZus{}input}\PY{p}{,}
            \PY{n}{temperature}\PY{o}{=}\PY{n+nb+bp}{None}\PY{p}{,}
            \PY{n}{index2word}\PY{o}{=}\PY{n}{index2word\PYZus{}y}\PY{p}{,}
            \PY{n}{verbose}\PY{o}{=}\PY{l+m+mi}{0}\PY{p}{)}
\end{Verbatim}

    \begin{Verbatim}[commandchars=\\\{\}]
{\color{incolor}In [{\color{incolor} }]:} \PY{n}{\PYZus{}} \PY{o}{=} \PY{n}{print\PYZus{}metrics}\PY{o}{.}\PY{n}{select}\PY{p}{[}\PY{l+s+s1}{\PYZsq{}}\PY{l+s+s1}{wups}\PY{l+s+s1}{\PYZsq{}}\PY{p}{]}\PY{p}{(}
                \PY{n}{gt\PYZus{}list}\PY{o}{=}\PY{n}{test\PYZus{}raw\PYZus{}y}\PY{p}{,}
                \PY{n}{pred\PYZus{}list}\PY{o}{=}\PY{n}{predictions\PYZus{}answers}\PY{p}{,}
                \PY{n}{verbose}\PY{o}{=}\PY{l+m+mi}{1}\PY{p}{,}
                \PY{n}{extra\PYZus{}vars}\PY{o}{=}\PY{n+nb+bp}{None}\PY{p}{)}
\end{Verbatim}

    \subsubsection{Predictions (RNN with
Vision)}\label{predictions-rnn-with-vision}

    \begin{Verbatim}[commandchars=\\\{\}]
{\color{incolor}In [{\color{incolor} }]:} \PY{k+kn}{from} \PY{n+nn}{kraino.core.model\PYZus{}zoo} \PY{k+kn}{import} \PY{n}{word\PYZus{}generator}
        \PY{c+c1}{\PYZsh{} we first need to add word\PYZus{}generator to \PYZus{}config }
        \PY{c+c1}{\PYZsh{} (we could have done this before, in the Config constructor)}
        \PY{c+c1}{\PYZsh{} we use maximum likelihood as a word generator}
        \PY{n}{text\PYZus{}image\PYZus{}rnn\PYZus{}model}\PY{o}{.}\PY{n}{\PYZus{}config}\PY{o}{.}\PY{n}{word\PYZus{}generator} \PY{o}{=} \PY{n}{word\PYZus{}generator}\PY{p}{[}\PY{l+s+s1}{\PYZsq{}}\PY{l+s+s1}{max\PYZus{}likelihood}\PY{l+s+s1}{\PYZsq{}}\PY{p}{]}
        \PY{n}{predictions\PYZus{}answers} \PY{o}{=} \PY{n}{text\PYZus{}image\PYZus{}rnn\PYZus{}model}\PY{o}{.}\PY{n}{decode\PYZus{}predictions}\PY{p}{(}
            \PY{n}{X}\PY{o}{=}\PY{n}{test\PYZus{}input}\PY{p}{,}
            \PY{n}{temperature}\PY{o}{=}\PY{n+nb+bp}{None}\PY{p}{,}
            \PY{n}{index2word}\PY{o}{=}\PY{n}{index2word\PYZus{}y}\PY{p}{,}
            \PY{n}{verbose}\PY{o}{=}\PY{l+m+mi}{0}\PY{p}{)}
\end{Verbatim}

    \begin{Verbatim}[commandchars=\\\{\}]
{\color{incolor}In [{\color{incolor} }]:} \PY{n}{\PYZus{}} \PY{o}{=} \PY{n}{print\PYZus{}metrics}\PY{o}{.}\PY{n}{select}\PY{p}{[}\PY{l+s+s1}{\PYZsq{}}\PY{l+s+s1}{wups}\PY{l+s+s1}{\PYZsq{}}\PY{p}{]}\PY{p}{(}
                \PY{n}{gt\PYZus{}list}\PY{o}{=}\PY{n}{test\PYZus{}raw\PYZus{}y}\PY{p}{,}
                \PY{n}{pred\PYZus{}list}\PY{o}{=}\PY{n}{predictions\PYZus{}answers}\PY{p}{,}
                \PY{n}{verbose}\PY{o}{=}\PY{l+m+mi}{1}\PY{p}{,}
                \PY{n}{extra\PYZus{}vars}\PY{o}{=}\PY{n+nb+bp}{None}\PY{p}{)}
\end{Verbatim}

\section{VQA}\label{vqa}
    The models that we have built so far can be transferred to other
dataset. Let us consider a recently introduced large-scale dataset, which is named VQA \citep{antol2015vqa}.
In this section, we train and
evaluate VQA models.
Since the reader should already be familiar with all the pieces, we just quickly jump into coding.
For the sake of simplicity, we 
use only BOW architectures.
Since VQA hides the test data for the purpose of challenge, we  use
the publicly available validation set to evaluate the architectures.

    \subsubsection{VQA Language Features}\label{vqa-language-features}

    \begin{Verbatim}[commandchars=\\\{\}]
{\color{incolor}In [{\color{incolor} }]:} \PY{c+c1}{\PYZsh{}TODO: Execute the following procedure (Shift+Enter)}
        \PY{k+kn}{from} \PY{n+nn}{kraino.utils} \PY{k+kn}{import} \PY{n}{data\PYZus{}provider}
        
        \PY{n}{vqa\PYZus{}dp} \PY{o}{=} \PY{n}{data\PYZus{}provider}\PY{o}{.}\PY{n}{select}\PY{p}{[}\PY{l+s+s1}{\PYZsq{}}\PY{l+s+s1}{vqa\PYZhy{}real\PYZus{}images\PYZhy{}open\PYZus{}ended}\PY{l+s+s1}{\PYZsq{}}\PY{p}{]}
        \PY{c+c1}{\PYZsh{} VQA has a few answers associated with one question. }
        \PY{c+c1}{\PYZsh{} We take the most frequently occuring answers (single\PYZus{}frequent).}
        \PY{c+c1}{\PYZsh{} Formal argument \PYZsq{}keep\PYZus{}top\PYZus{}qa\PYZus{}pairs\PYZsq{} allows to filter out }
        \PY{c+c1}{\PYZsh{} rare answers with the associated questions.}
        \PY{c+c1}{\PYZsh{} We use 0 as we want to keep all question answer pairs, }
        \PY{c+c1}{\PYZsh{} but you can change into 1000 and see how the results differ}
        \PY{n}{vqa\PYZus{}train\PYZus{}text\PYZus{}representation} \PY{o}{=} \PY{n}{vqa\PYZus{}dp}\PY{p}{[}\PY{l+s+s1}{\PYZsq{}}\PY{l+s+s1}{text}\PY{l+s+s1}{\PYZsq{}}\PY{p}{]}\PY{p}{(}
            \PY{n}{train\PYZus{}or\PYZus{}test}\PY{o}{=}\PY{l+s+s1}{\PYZsq{}}\PY{l+s+s1}{train}\PY{l+s+s1}{\PYZsq{}}\PY{p}{,}
            \PY{n}{answer\PYZus{}mode}\PY{o}{=}\PY{l+s+s1}{\PYZsq{}}\PY{l+s+s1}{single\PYZus{}frequent}\PY{l+s+s1}{\PYZsq{}}\PY{p}{,}
            \PY{n}{keep\PYZus{}top\PYZus{}qa\PYZus{}pairs}\PY{o}{=}\PY{l+m+mi}{1000}\PY{p}{)}
        \PY{n}{vqa\PYZus{}val\PYZus{}text\PYZus{}representation} \PY{o}{=} \PY{n}{vqa\PYZus{}dp}\PY{p}{[}\PY{l+s+s1}{\PYZsq{}}\PY{l+s+s1}{text}\PY{l+s+s1}{\PYZsq{}}\PY{p}{]}\PY{p}{(}
            \PY{n}{train\PYZus{}or\PYZus{}test}\PY{o}{=}\PY{l+s+s1}{\PYZsq{}}\PY{l+s+s1}{val}\PY{l+s+s1}{\PYZsq{}}\PY{p}{,}
            \PY{n}{answer\PYZus{}mode}\PY{o}{=}\PY{l+s+s1}{\PYZsq{}}\PY{l+s+s1}{single\PYZus{}frequent}\PY{l+s+s1}{\PYZsq{}}\PY{p}{)}
\end{Verbatim}

    \begin{Verbatim}[commandchars=\\\{\}]
{\color{incolor}In [{\color{incolor} }]:} \PY{k+kn}{from} \PY{n+nn}{toolz} \PY{k+kn}{import} \PY{n}{frequencies}
        \PY{n}{vqa\PYZus{}train\PYZus{}raw\PYZus{}x} \PY{o}{=} \PY{n}{vqa\PYZus{}train\PYZus{}text\PYZus{}representation}\PY{p}{[}\PY{l+s+s1}{\PYZsq{}}\PY{l+s+s1}{x}\PY{l+s+s1}{\PYZsq{}}\PY{p}{]}
        \PY{n}{vqa\PYZus{}train\PYZus{}raw\PYZus{}y} \PY{o}{=} \PY{n}{vqa\PYZus{}train\PYZus{}text\PYZus{}representation}\PY{p}{[}\PY{l+s+s1}{\PYZsq{}}\PY{l+s+s1}{y}\PY{l+s+s1}{\PYZsq{}}\PY{p}{]}
        \PY{n}{vqa\PYZus{}val\PYZus{}raw\PYZus{}x} \PY{o}{=} \PY{n}{vqa\PYZus{}val\PYZus{}text\PYZus{}representation}\PY{p}{[}\PY{l+s+s1}{\PYZsq{}}\PY{l+s+s1}{x}\PY{l+s+s1}{\PYZsq{}}\PY{p}{]}
        \PY{n}{vqa\PYZus{}val\PYZus{}raw\PYZus{}y} \PY{o}{=} \PY{n}{vqa\PYZus{}val\PYZus{}text\PYZus{}representation}\PY{p}{[}\PY{l+s+s1}{\PYZsq{}}\PY{l+s+s1}{y}\PY{l+s+s1}{\PYZsq{}}\PY{p}{]}
        \PY{c+c1}{\PYZsh{} we start from building the frequencies table}
        \PY{n}{vqa\PYZus{}wordcount\PYZus{}x} \PY{o}{=} \PY{n}{frequencies}\PY{p}{(}\PY{l+s+s1}{\PYZsq{}}\PY{l+s+s1}{ }\PY{l+s+s1}{\PYZsq{}}\PY{o}{.}\PY{n}{join}\PY{p}{(}\PY{n}{vqa\PYZus{}train\PYZus{}raw\PYZus{}x}\PY{p}{)}\PY{o}{.}\PY{n}{split}\PY{p}{(}\PY{l+s+s1}{\PYZsq{}}\PY{l+s+s1}{ }\PY{l+s+s1}{\PYZsq{}}\PY{p}{)}\PY{p}{)}
        \PY{c+c1}{\PYZsh{} we can keep all answer words in the answer as a class}
        \PY{c+c1}{\PYZsh{} therefore we use an artificial split symbol \PYZsq{}\PYZob{}\PYZsq{}}
        \PY{c+c1}{\PYZsh{} to not split the answer into words}
        \PY{c+c1}{\PYZsh{} you can see the difference if you replace \PYZsq{}\PYZob{}\PYZsq{} }
        \PY{c+c1}{\PYZsh{} with \PYZsq{} \PYZsq{} and print vqa\PYZus{}wordcount\PYZus{}y}
        \PY{n}{vqa\PYZus{}wordcount\PYZus{}y} \PY{o}{=} \PY{n}{frequencies}\PY{p}{(}\PY{l+s+s1}{\PYZsq{}}\PY{l+s+s1}{\PYZob{}}\PY{l+s+s1}{\PYZsq{}}\PY{o}{.}\PY{n}{join}\PY{p}{(}\PY{n}{vqa\PYZus{}train\PYZus{}raw\PYZus{}y}\PY{p}{)}\PY{o}{.}\PY{n}{split}\PY{p}{(}\PY{l+s+s1}{\PYZsq{}}\PY{l+s+s1}{\PYZob{}}\PY{l+s+s1}{\PYZsq{}}\PY{p}{)}\PY{p}{)}
        \PY{n}{vqa\PYZus{}wordcount\PYZus{}y}
\end{Verbatim}

    \subsubsection{Language-Only}\label{language-only}

    \begin{Verbatim}[commandchars=\\\{\}]
{\color{incolor}In [{\color{incolor} }]:} \PY{k+kn}{from} \PY{n+nn}{keras.preprocessing} \PY{k+kn}{import} \PY{n}{sequence}
        \PY{k+kn}{from} \PY{n+nn}{kraino.utils.input\PYZus{}output\PYZus{}space} \PY{k+kn}{import} \PY{n}{build\PYZus{}vocabulary}
        \PY{k+kn}{from} \PY{n+nn}{kraino.utils.input\PYZus{}output\PYZus{}space} \PY{k+kn}{import} \PY{n}{encode\PYZus{}questions\PYZus{}index}
        \PY{k+kn}{from} \PY{n+nn}{kraino.utils.input\PYZus{}output\PYZus{}space} \PY{k+kn}{import} \PY{n}{encode\PYZus{}answers\PYZus{}one\PYZus{}hot}
        \PY{n}{MAXLEN}\PY{o}{=}\PY{l+m+mi}{30}
        \PY{n}{vqa\PYZus{}word2index\PYZus{}x}\PY{p}{,} \PY{n}{vqa\PYZus{}index2word\PYZus{}x} \PY{o}{=} \PY{n}{build\PYZus{}vocabulary}
            \PY{p}{(}\PY{n}{this\PYZus{}wordcount} \PY{o}{=} \PY{n}{vqa\PYZus{}wordcount\PYZus{}x}\PY{p}{)}
        \PY{n}{vqa\PYZus{}word2index\PYZus{}y}\PY{p}{,} \PY{n}{vqa\PYZus{}index2word\PYZus{}y} \PY{o}{=} \PY{n}{build\PYZus{}vocabulary}
            \PY{p}{(}\PY{n}{this\PYZus{}wordcount} \PY{o}{=} \PY{n}{vqa\PYZus{}wordcount\PYZus{}y}\PY{p}{)}
        \PY{n}{vqa\PYZus{}train\PYZus{}x} \PY{o}{=} \PY{n}{sequence}\PY{o}{.}\PY{n}{pad\PYZus{}sequences}\PY{p}{(}\PY{n}{encode\PYZus{}questions\PYZus{}index}
            \PY{p}{(}\PY{n}{vqa\PYZus{}train\PYZus{}raw\PYZus{}x}\PY{p}{,} \PY{n}{vqa\PYZus{}word2index\PYZus{}x}\PY{p}{)}\PY{p}{,} \PY{n}{maxlen}\PY{o}{=}\PY{n}{MAXLEN}\PY{p}{)}
        \PY{n}{vqa\PYZus{}val\PYZus{}x} \PY{o}{=} \PY{n}{sequence}\PY{o}{.}\PY{n}{pad\PYZus{}sequences}\PY{p}{(}\PY{n}{encode\PYZus{}questions\PYZus{}index}
            \PY{p}{(}\PY{n}{vqa\PYZus{}val\PYZus{}raw\PYZus{}x}\PY{p}{,} \PY{n}{vqa\PYZus{}word2index\PYZus{}x}\PY{p}{)}\PY{p}{,} \PY{n}{maxlen}\PY{o}{=}\PY{n}{MAXLEN}\PY{p}{)}
        \PY{n}{vqa\PYZus{}train\PYZus{}y}\PY{p}{,} \PY{n}{\PYZus{}} \PY{o}{=} \PY{n}{encode\PYZus{}answers\PYZus{}one\PYZus{}hot}\PY{p}{(}
            \PY{n}{vqa\PYZus{}train\PYZus{}raw\PYZus{}y}\PY{p}{,} 
            \PY{n}{vqa\PYZus{}word2index\PYZus{}y}\PY{p}{,} 
            \PY{n}{answer\PYZus{}words\PYZus{}delimiter}\PY{o}{=}\PY{n}{vqa\PYZus{}train\PYZus{}text\PYZus{}representation}\PY{p}{[}\PY{l+s+s1}{\PYZsq{}}\PY{l+s+s1}{answer\PYZus{}words\PYZus{}delimiter}\PY{l+s+s1}{\PYZsq{}}\PY{p}{]}\PY{p}{}
            \PY{n}{is\PYZus{}only\PYZus{}first\PYZus{}answer\PYZus{}word}\PY{o}{=}\PY{n+nb+bp}{True}\PY{p}{,}
            \PY{n}{max\PYZus{}answer\PYZus{}time\PYZus{}steps}\PY{o}{=}\PY{l+m+mi}{1}\PY{p}{)}
        \PY{n}{vqa\PYZus{}val\PYZus{}y}\PY{p}{,} \PY{n}{\PYZus{}} \PY{o}{=} \PY{n}{encode\PYZus{}answers\PYZus{}one\PYZus{}hot}\PY{p}{(}
            \PY{n}{vqa\PYZus{}val\PYZus{}raw\PYZus{}y}\PY{p}{,} 
            \PY{n}{vqa\PYZus{}word2index\PYZus{}y}\PY{p}{,} 
            \PY{n}{answer\PYZus{}words\PYZus{}delimiter}\PY{o}{=}\PY{n}{vqa\PYZus{}train\PYZus{}text\PYZus{}representation}\PY{p}{[}\PY{l+s+s1}{\PYZsq{}}\PY{l+s+s1}{answer\PYZus{}words\PYZus{}delimiter}\PY{l+s+s1}{\PYZsq{}}\PY{p}{]}\PY{p}{}
            \PY{n}{is\PYZus{}only\PYZus{}first\PYZus{}answer\PYZus{}word}\PY{o}{=}\PY{n+nb+bp}{True}\PY{p}{,}
            \PY{n}{max\PYZus{}answer\PYZus{}time\PYZus{}steps}\PY{o}{=}\PY{l+m+mi}{1}\PY{p}{)}
\end{Verbatim}

    \begin{Verbatim}[commandchars=\\\{\}]
{\color{incolor}In [{\color{incolor} }]:} \PY{k+kn}{from} \PY{n+nn}{kraino.core.model\PYZus{}zoo} \PY{k+kn}{import} \PY{n}{Config}
        \PY{k+kn}{from} \PY{n+nn}{kraino.core.model\PYZus{}zoo} \PY{k+kn}{import} \PY{n}{word\PYZus{}generator}
        \PY{c+c1}{\PYZsh{} We are re\PYZhy{}using the BlindBOW mode}
        \PY{c+c1}{\PYZsh{} Please make sure you have run the cell with the class definition}
        \PY{c+c1}{\PYZsh{} VQA is larger, so we can increase the dimensionality of the embedding}
        \PY{n}{vqa\PYZus{}model\PYZus{}config} \PY{o}{=} \PY{n}{Config}\PY{p}{(}
            \PY{n}{textual\PYZus{}embedding\PYZus{}dim}\PY{o}{=}\PY{l+m+mi}{1000}\PY{p}{,}
            \PY{n}{input\PYZus{}dim}\PY{o}{=}\PY{n+nb}{len}\PY{p}{(}\PY{n}{vqa\PYZus{}word2index\PYZus{}x}\PY{o}{.}\PY{n}{keys}\PY{p}{(}\PY{p}{)}\PY{p}{)}\PY{p}{,}
            \PY{n}{output\PYZus{}dim}\PY{o}{=}\PY{n+nb}{len}\PY{p}{(}\PY{n}{vqa\PYZus{}word2index\PYZus{}y}\PY{o}{.}\PY{n}{keys}\PY{p}{(}\PY{p}{)}\PY{p}{)}\PY{p}{,}
            \PY{n}{word\PYZus{}generator} \PY{o}{=} \PY{n}{word\PYZus{}generator}\PY{p}{[}\PY{l+s+s1}{\PYZsq{}}\PY{l+s+s1}{max\PYZus{}likelihood}\PY{l+s+s1}{\PYZsq{}}\PY{p}{]}\PY{p}{)}
        \PY{n}{vqa\PYZus{}text\PYZus{}bow\PYZus{}model} \PY{o}{=} \PY{n}{BlindBOW}\PY{p}{(}\PY{n}{vqa\PYZus{}model\PYZus{}config}\PY{p}{)}
        \PY{n}{vqa\PYZus{}text\PYZus{}bow\PYZus{}model}\PY{o}{.}\PY{n}{create}\PY{p}{(}\PY{p}{)}
        \PY{n}{vqa\PYZus{}text\PYZus{}bow\PYZus{}model}\PY{o}{.}\PY{n}{compile}\PY{p}{(}
            \PY{n}{loss}\PY{o}{=}\PY{l+s+s1}{\PYZsq{}}\PY{l+s+s1}{categorical\PYZus{}crossentropy}\PY{l+s+s1}{\PYZsq{}}\PY{p}{,} 
            \PY{n}{optimizer}\PY{o}{=}\PY{l+s+s1}{\PYZsq{}}\PY{l+s+s1}{adam}\PY{l+s+s1}{\PYZsq{}}\PY{p}{)}
\end{Verbatim}

    \begin{Verbatim}[commandchars=\\\{\}]
{\color{incolor}In [{\color{incolor} }]:} \PY{n}{vqa\PYZus{}text\PYZus{}bow\PYZus{}model}\PY{o}{.}\PY{n}{fit}\PY{p}{(}
            \PY{n}{vqa\PYZus{}train\PYZus{}x}\PY{p}{,} 
            \PY{n}{vqa\PYZus{}train\PYZus{}y}\PY{p}{,}
            \PY{n}{batch\PYZus{}size}\PY{o}{=}\PY{l+m+mi}{512}\PY{p}{,}
            \PY{n}{nb\PYZus{}epoch}\PY{o}{=}\PY{l+m+mi}{10}\PY{p}{,}
            \PY{n}{validation\PYZus{}split}\PY{o}{=}\PY{l+m+mf}{0.1}\PY{p}{,}
            \PY{n}{show\PYZus{}accuracy}\PY{o}{=}\PY{n+nb+bp}{True}\PY{p}{)}
\end{Verbatim}

    \begin{Verbatim}[commandchars=\\\{\}]
        \PY{n}{vqa\PYZus{}predictions\PYZus{}answers} \PY{o}{=} \PY{n}{vqa\PYZus{}text\PYZus{}bow\PYZus{}model}\PY{o}{.}\PY{n}{decode\PYZus{}predictions}\PY{p}{(}
            \PY{n}{X}\PY{o}{=}\PY{n}{vqa\PYZus{}val\PYZus{}x}\PY{p}{,}
            \PY{n}{temperature}\PY{o}{=}\PY{n+nb+bp}{None}\PY{p}{,}
            \PY{n}{index2word}\PY{o}{=}\PY{n}{vqa\PYZus{}index2word\PYZus{}y}\PY{p}{,}
            \PY{n}{verbose}\PY{o}{=}\PY{l+m+mi}{0}\PY{p}{)}
\end{Verbatim}

    \begin{Verbatim}[commandchars=\\\{\}]
        \PY{n}{vqa\PYZus{}vars} \PY{o}{=} \PY{p}{\PYZob{}}
            \PY{l+s+s1}{\PYZsq{}}\PY{l+s+s1}{question\PYZus{}id}\PY{l+s+s1}{\PYZsq{}}\PY{p}{:}\PY{n}{vqa\PYZus{}val\PYZus{}text\PYZus{}representation}\PY{p}{[}\PY{l+s+s1}{\PYZsq{}}\PY{l+s+s1}{question\PYZus{}id}\PY{l+s+s1}{\PYZsq{}}\PY{p}{]}\PY{p}{,}
            \PY{l+s+s1}{\PYZsq{}}\PY{l+s+s1}{vqa\PYZus{}object}\PY{l+s+s1}{\PYZsq{}}\PY{p}{:}\PY{n}{vqa\PYZus{}val\PYZus{}text\PYZus{}representation}\PY{p}{[}\PY{l+s+s1}{\PYZsq{}}\PY{l+s+s1}{vqa\PYZus{}object}\PY{l+s+s1}{\PYZsq{}}\PY{p}{]}\PY{p}{,}
            \PY{l+s+s1}{\PYZsq{}}\PY{l+s+s1}{resfun}\PY{l+s+s1}{\PYZsq{}}\PY{p}{:} 
                \PY{k}{lambda} \PY{n}{x}\PY{p}{:} \PYZbs{}
                    \PY{n}{vqa\PYZus{}val\PYZus{}text\PYZus{}representation}\PY{p}{[}\PY{l+s+s1}{\PYZsq{}}\PY{l+s+s1}{vqa\PYZus{}object}\PY{l+s+s1}{\PYZsq{}}\PY{p}{]}\PY{o}{.}\PY{n}{loadRes}\PY{p}{(}
	                  \PY{n}{x}\PY{p}{,} \PY{n}{vqa\PYZus{}val\PYZus{}text\PYZus{}representation}\PY{p}{[}\PY{l+s+s1}{\PYZsq{}}\PY{l+s+s1}{questions\PYZus{}path}\PY{l+s+s1}{\PYZsq{}}\PY{p}{]}\PY{p}{)}
        \PY{p}{\PYZcb{}}
\end{Verbatim}

    \begin{Verbatim}[commandchars=\\\{\}]
{\color{incolor}In [{\color{incolor} }]:} \PY{k+kn}{from} \PY{n+nn}{kraino.utils} \PY{k+kn}{import} \PY{n}{print\PYZus{}metrics}
        
        
        \PY{n}{\PYZus{}} \PY{o}{=} \PY{n}{print\PYZus{}metrics}\PY{o}{.}\PY{n}{select}\PY{p}{[}\PY{l+s+s1}{\PYZsq{}}\PY{l+s+s1}{vqa}\PY{l+s+s1}{\PYZsq{}}\PY{p}{]}\PY{p}{(}
                \PY{n}{gt\PYZus{}list}\PY{o}{=}\PY{n}{vqa\PYZus{}val\PYZus{}raw\PYZus{}y}\PY{p}{,}
                \PY{n}{pred\PYZus{}list}\PY{o}{=}\PY{n}{vqa\PYZus{}predictions\PYZus{}answers}\PY{p}{,}
                \PY{n}{verbose}\PY{o}{=}\PY{l+m+mi}{1}\PY{p}{,}
                \PY{n}{extra\PYZus{}vars}\PY{o}{=}\PY{n}{vqa\PYZus{}vars}\PY{p}{)}
\end{Verbatim}

    \subsubsection{VQA Language+Vision}\label{vqa-languagevision}

    \begin{Verbatim}[commandchars=\\\{\}]
{\color{incolor}In [{\color{incolor} }]:} \PY{c+c1}{\PYZsh{} the name for visual features that we use}
        \PY{n}{VQA\PYZus{}CNN\PYZus{}NAME}\PY{o}{=}\PY{l+s+s1}{\PYZsq{}}\PY{l+s+s1}{vgg\PYZus{}net}\PY{l+s+s1}{\PYZsq{}}
        \PY{c+c1}{\PYZsh{} VQA\PYZus{}CNN\PYZus{}NAME=\PYZsq{}googlenet\PYZsq{}}
        \PY{c+c1}{\PYZsh{} the layer in CNN that is used to extract features}
        \PY{n}{VQA\PYZus{}PERCEPTION\PYZus{}LAYER}\PY{o}{=}\PY{l+s+s1}{\PYZsq{}}\PY{l+s+s1}{fc7}\PY{l+s+s1}{\PYZsq{}}
        \PY{c+c1}{\PYZsh{} PERCEPTION\PYZus{}LAYER=\PYZsq{}pool5\PYZhy{}7x7\PYZus{}s1\PYZsq{}}
        
        \PY{n}{vqa\PYZus{}train\PYZus{}visual\PYZus{}features} \PY{o}{=} \PY{n}{vqa\PYZus{}dp}\PY{p}{[}\PY{l+s+s1}{\PYZsq{}}\PY{l+s+s1}{perception}\PY{l+s+s1}{\PYZsq{}}\PY{p}{]}\PY{p}{(}
            \PY{n}{train\PYZus{}or\PYZus{}test}\PY{o}{=}\PY{l+s+s1}{\PYZsq{}}\PY{l+s+s1}{train}\PY{l+s+s1}{\PYZsq{}}\PY{p}{,}
            \PY{n}{names\PYZus{}list}\PY{o}{=}\PY{n}{vqa\PYZus{}train\PYZus{}text\PYZus{}representation}\PY{p}{[}\PY{l+s+s1}{\PYZsq{}}\PY{l+s+s1}{img\PYZus{}name}\PY{l+s+s1}{\PYZsq{}}\PY{p}{]}\PY{p}{,}
            \PY{n}{parts\PYZus{}extractor}\PY{o}{=}\PY{n+nb+bp}{None}\PY{p}{,}
            \PY{n}{max\PYZus{}parts}\PY{o}{=}\PY{n+nb+bp}{None}\PY{p}{,}
            \PY{n}{perception}\PY{o}{=}\PY{n}{VQA\PYZus{}CNN\PYZus{}NAME}\PY{p}{,}
            \PY{n}{layer}\PY{o}{=}\PY{n}{VQA\PYZus{}PERCEPTION\PYZus{}LAYER}\PY{p}{,}
            \PY{n}{second\PYZus{}layer}\PY{o}{=}\PY{n+nb+bp}{None}
            \PY{p}{)}
        \PY{n}{vqa\PYZus{}train\PYZus{}visual\PYZus{}features}\PY{o}{.}\PY{n}{shape}
\end{Verbatim}

    \begin{Verbatim}[commandchars=\\\{\}]
{\color{incolor}In [{\color{incolor} }]:} \PY{n}{vqa\PYZus{}val\PYZus{}visual\PYZus{}features} \PY{o}{=} \PY{n}{vqa\PYZus{}dp}\PY{p}{[}\PY{l+s+s1}{\PYZsq{}}\PY{l+s+s1}{perception}\PY{l+s+s1}{\PYZsq{}}\PY{p}{]}\PY{p}{(}
            \PY{n}{train\PYZus{}or\PYZus{}test}\PY{o}{=}\PY{l+s+s1}{\PYZsq{}}\PY{l+s+s1}{val}\PY{l+s+s1}{\PYZsq{}}\PY{p}{,}
            \PY{n}{names\PYZus{}list}\PY{o}{=}\PY{n}{vqa\PYZus{}val\PYZus{}text\PYZus{}representation}\PY{p}{[}\PY{l+s+s1}{\PYZsq{}}\PY{l+s+s1}{img\PYZus{}name}\PY{l+s+s1}{\PYZsq{}}\PY{p}{]}\PY{p}{,}
            \PY{n}{parts\PYZus{}extractor}\PY{o}{=}\PY{n+nb+bp}{None}\PY{p}{,}
            \PY{n}{max\PYZus{}parts}\PY{o}{=}\PY{n+nb+bp}{None}\PY{p}{,}
            \PY{n}{perception}\PY{o}{=}\PY{n}{VQA\PYZus{}CNN\PYZus{}NAME}\PY{p}{,}
            \PY{n}{layer}\PY{o}{=}\PY{n}{VQA\PYZus{}PERCEPTION\PYZus{}LAYER}\PY{p}{,}
            \PY{n}{second\PYZus{}layer}\PY{o}{=}\PY{n+nb+bp}{None}
            \PY{p}{)}
        \PY{n}{vqa\PYZus{}val\PYZus{}visual\PYZus{}features}\PY{o}{.}\PY{n}{shape}
\end{Verbatim}

    \begin{Verbatim}[commandchars=\\\{\}]
{\color{incolor}In [{\color{incolor} }]:} \PY{k+kn}{from} \PY{n+nn}{kraino.core.model\PYZus{}zoo} \PY{k+kn}{import} \PY{n}{Config}
        \PY{k+kn}{from} \PY{n+nn}{kraino.core.model\PYZus{}zoo} \PY{k+kn}{import} \PY{n}{word\PYZus{}generator}
        
        \PY{c+c1}{\PYZsh{} dimensionality of embeddings}
        \PY{n}{VQA\PYZus{}EMBEDDING\PYZus{}DIM} \PY{o}{=} \PY{l+m+mi}{1000}
        \PY{c+c1}{\PYZsh{} kind of multimodal fusion (ave, concat, mul, sum)}
        \PY{n}{VQA\PYZus{}MULTIMODAL\PYZus{}MERGE\PYZus{}MODE} \PY{o}{=} \PY{l+s+s1}{\PYZsq{}}\PY{l+s+s1}{mul}\PY{l+s+s1}{\PYZsq{}}
        
        \PY{n}{vqa\PYZus{}model\PYZus{}config} \PY{o}{=} \PY{n}{Config}\PY{p}{(}
            \PY{n}{textual\PYZus{}embedding\PYZus{}dim}\PY{o}{=}\PY{n}{VQA\PYZus{}EMBEDDING\PYZus{}DIM}\PY{p}{,}
            \PY{n}{visual\PYZus{}embedding\PYZus{}dim}\PY{o}{=}\PY{n}{VQA\PYZus{}EMBEDDING\PYZus{}DIM}\PY{p}{,}
            \PY{n}{multimodal\PYZus{}merge\PYZus{}mode}\PY{o}{=}\PY{n}{VQA\PYZus{}MULTIMODAL\PYZus{}MERGE\PYZus{}MODE}\PY{p}{,}
            \PY{n}{input\PYZus{}dim}\PY{o}{=}\PY{n+nb}{len}\PY{p}{(}\PY{n}{vqa\PYZus{}word2index\PYZus{}x}\PY{o}{.}\PY{n}{keys}\PY{p}{(}\PY{p}{)}\PY{p}{)}\PY{p}{,}
            \PY{n}{output\PYZus{}dim}\PY{o}{=}\PY{n+nb}{len}\PY{p}{(}\PY{n}{vqa\PYZus{}word2index\PYZus{}y}\PY{o}{.}\PY{n}{keys}\PY{p}{(}\PY{p}{)}\PY{p}{)}\PY{p}{,}
            \PY{n}{visual\PYZus{}dim}\PY{o}{=}\PY{n}{vqa\PYZus{}train\PYZus{}visual\PYZus{}features}\PY{o}{.}\PY{n}{shape}\PY{p}{[}\PY{l+m+mi}{1}\PY{p}{]}\PY{p}{,}
            \PY{n}{word\PYZus{}generator}\PY{o}{=}\PY{n}{word\PYZus{}generator}\PY{p}{[}\PY{l+s+s1}{\PYZsq{}}\PY{l+s+s1}{max\PYZus{}likelihood}\PY{l+s+s1}{\PYZsq{}}\PY{p}{]}\PY{p}{)}
        \PY{n}{vqa\PYZus{}text\PYZus{}image\PYZus{}bow\PYZus{}model} \PY{o}{=} \PY{n}{VisionLanguageBOW}\PY{p}{(}\PY{n}{vqa\PYZus{}model\PYZus{}config}\PY{p}{)}
        \PY{n}{vqa\PYZus{}text\PYZus{}image\PYZus{}bow\PYZus{}model}\PY{o}{.}\PY{n}{create}\PY{p}{(}\PY{p}{)}
        \PY{n}{vqa\PYZus{}text\PYZus{}image\PYZus{}bow\PYZus{}model}\PY{o}{.}\PY{n}{compile}\PY{p}{(}
            \PY{n}{loss}\PY{o}{=}\PY{l+s+s1}{\PYZsq{}}\PY{l+s+s1}{categorical\PYZus{}crossentropy}\PY{l+s+s1}{\PYZsq{}}\PY{p}{,} 
            \PY{n}{optimizer}\PY{o}{=}\PY{l+s+s1}{\PYZsq{}}\PY{l+s+s1}{adam}\PY{l+s+s1}{\PYZsq{}}\PY{p}{)}
\end{Verbatim}

    \begin{Verbatim}[commandchars=\\\{\}]
{\color{incolor}In [{\color{incolor} }]:} \PY{n}{vqa\PYZus{}train\PYZus{}input} \PY{o}{=} \PY{p}{[}\PY{n}{vqa\PYZus{}train\PYZus{}x}\PY{p}{,} \PY{n}{vqa\PYZus{}train\PYZus{}visual\PYZus{}features}\PY{p}{]}
        \PY{n}{vqa\PYZus{}val\PYZus{}input} \PY{o}{=} \PY{p}{[}\PY{n}{vqa\PYZus{}val\PYZus{}x}\PY{p}{,} \PY{n}{vqa\PYZus{}val\PYZus{}visual\PYZus{}features}\PY{p}{]}
\end{Verbatim}

    \begin{Verbatim}[commandchars=\\\{\}]
{\color{incolor}In [{\color{incolor} }]:} \PY{c+c1}{\PYZsh{}== Model training}
        \PY{n}{vqa\PYZus{}text\PYZus{}image\PYZus{}bow\PYZus{}model}\PY{o}{.}\PY{n}{fit}\PY{p}{(}
            \PY{n}{vqa\PYZus{}train\PYZus{}input}\PY{p}{,} 
            \PY{n}{vqa\PYZus{}train\PYZus{}y}\PY{p}{,}
            \PY{n}{batch\PYZus{}size}\PY{o}{=}\PY{l+m+mi}{512}\PY{p}{,}
            \PY{n}{nb\PYZus{}epoch}\PY{o}{=}\PY{l+m+mi}{10}\PY{p}{,}
            \PY{n}{validation\PYZus{}split}\PY{o}{=}\PY{l+m+mf}{0.1}\PY{p}{,}
            \PY{n}{show\PYZus{}accuracy}\PY{o}{=}\PY{n+nb+bp}{True}\PY{p}{)}
\end{Verbatim}

    \begin{Verbatim}[commandchars=\\\{\}]
        \PY{c+c1}{\PYZsh{} we use maximum likelihood as a word generator}
        \PY{n}{vqa\PYZus{}predictions\PYZus{}answers} \PY{o}{=} \PY{n}{vqa\PYZus{}text\PYZus{}image\PYZus{}bow\PYZus{}model}\PY{o}{.}\PY{n}{decode\PYZus{}predictions}\PY{p}{(}
            \PY{n}{X}\PY{o}{=}\PY{n}{vqa\PYZus{}val\PYZus{}input}\PY{p}{,}
            \PY{n}{temperature}\PY{o}{=}\PY{n+nb+bp}{None}\PY{p}{,}
            \PY{n}{index2word}\PY{o}{=}\PY{n}{vqa\PYZus{}index2word\PYZus{}y}\PY{p}{,}
            \PY{n}{verbose}\PY{o}{=}\PY{l+m+mi}{0}\PY{p}{)}
\end{Verbatim}

    \begin{Verbatim}[commandchars=\\\{\}]
        \PY{n}{vqa\PYZus{}vars} \PY{o}{=} \PY{p}{\PYZob{}}
            \PY{l+s+s1}{\PYZsq{}}\PY{l+s+s1}{question\PYZus{}id}\PY{l+s+s1}{\PYZsq{}}\PY{p}{:}\PY{n}{vqa\PYZus{}val\PYZus{}text\PYZus{}representation}\PY{p}{[}\PY{l+s+s1}{\PYZsq{}}\PY{l+s+s1}{question\PYZus{}id}\PY{l+s+s1}{\PYZsq{}}\PY{p}{]}\PY{p}{,}
            \PY{l+s+s1}{\PYZsq{}}\PY{l+s+s1}{vqa\PYZus{}object}\PY{l+s+s1}{\PYZsq{}}\PY{p}{:}\PY{n}{vqa\PYZus{}val\PYZus{}text\PYZus{}representation}\PY{p}{[}\PY{l+s+s1}{\PYZsq{}}\PY{l+s+s1}{vqa\PYZus{}object}\PY{l+s+s1}{\PYZsq{}}\PY{p}{]}\PY{p}{,}
            \PY{l+s+s1}{\PYZsq{}}\PY{l+s+s1}{resfun}\PY{l+s+s1}{\PYZsq{}}\PY{p}{:} 
                \PY{k}{lambda} \PY{n}{x}\PY{p}{:} \PYZbs{}
                    \PY{n}{vqa\PYZus{}val\PYZus{}text\PYZus{}representation}\PY{p}{[}\PY{l+s+s1}{\PYZsq{}}\PY{l+s+s1}{vqa\PYZus{}object}\PY{l+s+s1}{\PYZsq{}}\PY{p}{]}\PY{o}{.}\PY{n}{loadRes}\PY{p}{(}
					\PY{n}{x}\PY{p}{,} \PY{n}{vqa\PYZus{}val\PYZus{}text\PYZus{}representation}\PY{p}{[}\PY{l+s+s1}{\PYZsq{}}\PY{l+s+s1}{questions\PYZus{}path}\PY{l+s+s1}{\PYZsq{}}\PY{p}{]}\PY{p}{)}
        \PY{p}{\PYZcb{}}
\end{Verbatim}

    \begin{Verbatim}[commandchars=\\\{\}]
{\color{incolor}In [{\color{incolor} }]:} \PY{k+kn}{from} \PY{n+nn}{kraino.utils} \PY{k+kn}{import} \PY{n}{print\PYZus{}metrics}
        
        
        \PY{n}{\PYZus{}} \PY{o}{=} \PY{n}{print\PYZus{}metrics}\PY{o}{.}\PY{n}{select}\PY{p}{[}\PY{l+s+s1}{\PYZsq{}}\PY{l+s+s1}{vqa}\PY{l+s+s1}{\PYZsq{}}\PY{p}{]}\PY{p}{(}
                \PY{n}{gt\PYZus{}list}\PY{o}{=}\PY{n}{vqa\PYZus{}val\PYZus{}raw\PYZus{}y}\PY{p}{,}
                \PY{n}{pred\PYZus{}list}\PY{o}{=}\PY{n}{vqa\PYZus{}predictions\PYZus{}answers}\PY{p}{,}
                \PY{n}{verbose}\PY{o}{=}\PY{l+m+mi}{1}\PY{p}{,}
                \PY{n}{extra\PYZus{}vars}\PY{o}{=}\PY{n}{vqa\PYZus{}vars}\PY{p}{)}
\end{Verbatim}

\section{New Research Opportunities}\label{new-research-opportunities}
The task that tests machines via questions about the content of images is
a quite new research direction that recently has gained popularity.
Therefore, many opportunities are available. We end the tutorial by enlisting a few possible directions.
    \begin{itemize}
\itemsep1pt\parskip0pt\parsep0pt
\item
  \textbf{Global Representation} In this tutorial, we use 
  a global, full-frame representation of the images. Such a representation may destroy
  too much information. Therefore, it seems a fine-grained
  alternatives should be valid options. Maybe we should use detections, or object proposals (e.g. \citet{ilievski2016fda} use a question dependent detections, and \citet{malinowski2016mean} use object proposals to enrich a visual representation). 
  We could also use attention models, which become quite successful in answering questions about images \citep{lu2016hierarchical}.
  However, there is still a hope for global representations if
  they are trained end-to-end for the task, and question dependent. In the end, our global
  representation is extracted from CNNs trained on a different dataset
  (ImageNet), and for different task (object classification).
\item
  \textbf{3D Scene Representation} Most of current approaches, and all
  neural-based approaches, are trained on 2D images. However, 
  some spatial relations such as `behind' may need a 3d representation of
  the scene (in fact \citet{malinowski14nips} design spatial rules using a 3d coordinate system). 
  DAQUAR is built on top of \citet{silbermanECCV12} that provides both modes (2D images, and 3D depth), however, such a richer visual information is currently not fully exploited.
\item
  \textbf{Recurrent Neural Networks} There is disturbingly small gap
  between BOW and RNN models. As we have seen in the tutorial, some questions
  clearly require an order, but such questions at the same time become
  longer, semantically more difficult, and require better a visual
  understanding of the world. To handle them we may need other RNNs
  architectures, or better ways of fusing two modalities, or better
  \textbf{Global Representation}.
\item
  \textbf{Logical Reasoning} There are few questions that require a bit
  more sophisticated logical reasoning such as negation. Can Recurrent
  Neural Networks learn such logical operators? What about
  compositionality of the language? Perhaps, we should aim at mixed approaches, similar to the work of \citet{andreas2015deep}.
\item
  \textbf{Language + Vision} There is still a small gap between Language
  Only and Vision + Language models. But clearly, we need pictures to
  answer questions about images. So what is missing here? Is it due to
  \textbf{Global Representation}, \textbf{3D Scene Representation} or
  there is something missing in fusing two modalities? The latter is studied, with encouraging results, in \citet{fukui2016multimodal}.
\item
  \textbf{Learning from Few Examples} In the Visual Turing Test, many
  questions are quite unique. But then how the models can generalize to
  new questions? What if a question is completely new, but its parts
  have been already observed (compositionality)? Can models guess the
  meaning of a new word from its context? 
\item
  \textbf{Ambiguities} How to deal with ambiguities? They are all
  inherent in the task, so cannot be just ignored, and should be
  incorporated into question answering methods as well as evaluation
  metrics.
\item
  \textbf{Evaluation Measures} Although we have WUPS and Consensus, both
  are far from being perfect. Consensus has higher annotation cost for
  ambiguous tasks, and is unclear how to formally define good consensus
  measure. WUPS is an ontology dependent, which may be quite costly to build for all interesting domains?
  Finally, the current evaluation metrics ignore the tail of the answer distribution encouraging models to focus only on a few most frequent answers.
\end{itemize}

\bibliographystyle{plainnat}
\bibliography{bibliomateusz_ask_journal}
\end{document}